\PassOptionsToPackage{unicode}{hyperref}
\PassOptionsToPackage{hyphens}{url}
\documentclass[
]{article}
\usepackage{xcolor}
\usepackage{amsmath,amssymb}
\usepackage{iftex}
\ifPDFTeX
  \usepackage[T1]{fontenc}
  \usepackage[utf8]{inputenc}
  \usepackage{textcomp} 
\else 
  \usepackage{unicode-math} 
  \defaultfontfeatures{Scale=MatchLowercase}
  \defaultfontfeatures[\rmfamily]{Ligatures=TeX,Scale=1}
\fi
\usepackage{lmodern}
\ifPDFTeX\else
\fi
\IfFileExists{upquote.sty}{\usepackage{upquote}}{}
\IfFileExists{microtype.sty}{
  \usepackage[]{microtype}
  \UseMicrotypeSet[protrusion]{basicmath} 
}{}
\makeatletter
\@ifundefined{KOMAClassName}{
  \IfFileExists{parskip.sty}{%
    \usepackage{parskip}
  }{
    \setlength{\parindent}{0pt}
    \setlength{\parskip}{6pt plus 2pt minus 1pt}}
}{
  \KOMAoptions{parskip=half}}
\makeatother
\usepackage{color}
\usepackage{fancyvrb}

\DefineVerbatimEnvironment{Highlighting}{Verbatim}{commandchars=\\\{\}}
\newenvironment{Shaded}{}{}

\newcommand{\DataTypeTok}[1]{\textcolor[rgb]{0.56,0.13,0.00}{#1}}
\newcommand{\DecValTok}[1]{\textcolor[rgb]{0.25,0.63,0.44}{#1}}

\newcommand{\FunctionTok}[1]{\textcolor[rgb]{0.02,0.16,0.49}{#1}}

\newcommand{\KeywordTok}[1]{\textcolor[rgb]{0.00,0.44,0.13}{\textbf{#1}}}

\newcommand{\OtherTok}[1]{\textcolor[rgb]{0.00,0.44,0.13}{#1}}

\newcommand{\StringTok}[1]{\textcolor[rgb]{0.25,0.44,0.63}{#1}}

\usepackage{longtable,booktabs,array}
\usepackage{calc} 
\usepackage{etoolbox}
\makeatletter
\patchcmd\longtable{\par}{\if@noskipsec\mbox{}\fi\par}{}{}
\makeatother
\IfFileExists{footnotehyper.sty}{\usepackage{footnotehyper}}{\usepackage{footnote}}
\makesavenoteenv{longtable}
\usepackage{graphicx}
\usepackage[margin=1in]{geometry}
\usepackage{caption}
\usepackage{pdflscape}

\usepackage{newunicodechar}
\newunicodechar{✓}{\ensuremath{\checkmark}}
\newunicodechar{✗}{\ensuremath{\times}}
\newunicodechar{τ}{\ensuremath{\tau}}
\newunicodechar{κ}{\ensuremath{\kappa}}
\newunicodechar{§}{\S}
\newunicodechar{↔}{\ensuremath{\leftrightarrow}}
\newunicodechar{≈}{\ensuremath{\approx}}
\newunicodechar{Δ}{\ensuremath{\Delta}}
\newunicodechar{→}{\ensuremath{\rightarrow}}
\newunicodechar{−}{\ensuremath{-}}
\newunicodechar{≥}{\ensuremath{\geq}}
\newunicodechar{≤}{\ensuremath{\leq}}
\newunicodechar{≡}{\ensuremath{\equiv}}
\newunicodechar{ρ}{\ensuremath{\rho}}
\newunicodechar{∈}{\ensuremath{\in}}
\newunicodechar{≫}{\ensuremath{\gg}}
\newunicodechar{≪}{\ensuremath{\ll}}

\makeatletter
\newsavebox\pandoc@box
\newcommand*\pandocbounded[1]{
  \sbox\pandoc@box{#1}%
  \Gscale@div\@tempa{\textheight}{\dimexpr\ht\pandoc@box+\dp\pandoc@box\relax}%
  \Gscale@div\@tempb{\linewidth}{\wd\pandoc@box}%
  \ifdim\@tempb\p@<\@tempa\p@\let\@tempa\@tempb\fi
  \ifdim\@tempa\p@<\p@\scalebox{\@tempa}{\usebox\pandoc@box}%
  \else\usebox{\pandoc@box}%
  \fi%
}
\def\fps@figure{htbp}
\makeatother
\providecommand{\tightlist}{%
  \setlength{\itemsep}{0pt}\setlength{\parskip}{0pt}}
\usepackage{bookmark}
\IfFileExists{xurl.sty}{\usepackage{xurl}}{} 
\hypersetup{
  hidelinks,
  pdfcreator={LaTeX via pandoc}}

\title{InvestPhilBench: A Multi-Layer Benchmark for Evaluating Large Language Model Procedural Reasoning in Expert Investment Philosophy}

\author{%
  Mingguang Chen\textsuperscript{1}\thanks{%
    \textsuperscript{*}Corresponding authors.
    \textsuperscript{1}University of California, Riverside (UCR). Email: \texttt{mchen041@ucr.edu}.
    \textsuperscript{2}Illinois Institute of Technology (IIT). Email: \texttt{boqu.sh2019@gmail.com}.%
  } \and
  Bo Qu\textsuperscript{2,*}%
}

\date{}
\begin{document}
\maketitle
\begin{abstract}
Large language models are increasingly deployed as investment research assistants, yet no benchmark tests whether they can accurately reconstruct and apply the specific procedural decision frameworks of expert investors. We introduce \textbf{InvestPhilBench}, a multi-layer benchmark spanning eight cognitive tiers, from principle identification (L1) to novel framework extrapolation (L8), with a contamination-resistant dynamic-perturbation design for the generative layers. The v0.6 release comprises \textbf{118} primary-source-verified principle cards, \textbf{25} decision-framework cards with explicit topology metadata, and \textbf{243} QA questions (197 dev / 46 held-out test), including 8 expert-difficulty questions targeting gate-level reconstruction failures (first evaluated in v1.0). For reproducible scoring at scale we introduce the \textbf{Benchmark Automated Scoring Pipeline (BASP)} — five metrics (OGRS, KCCS, SAP@k, IVP, CKCA) — the \textbf{Failure Mode Detection Protocol (FMDP)} with computable rules for six failure modes, and \textbf{Gate Reconstruction Accuracy (GRA)}, a per-gate metric for questions with gold reasoning programs. This release is primarily a \emph{benchmark-and-methodology} contribution: its empirical study — a four-model sanity wave on the 188-question development split (closed-book) — is deliberately preliminary and serves to stress-test the metric design, not to rank models. The wave shows a sharp provider-tier split (BASP 0.906 ↔ 0.438), but these mixed-judge numbers are confounded upper bounds (§8.7). The central methodological finding survives the caveat: the BASP composite \emph{saturates} at the frontier (Claude L4 = 0.932) while GRA still exposes a procedural deficit (frontier L4 GRA ≈ 0.77, L7 GRA 0.57–0.62) — composite scoring rewards fluent prose and hides the procedural gap. On a 100-item expert-annotated gold set, the automated BASP composite tracks the human reference at Pearson r = 0.72 (MAE = 0.10), with attribution (SAP@3) the weakest sub-metric and the failure-mode detector over-flagging (§11.3). v0.6 also implements a unified judge and true model-in-the-loop retrieval/oracle conditions, validated on a pipeline pilot (§8.3); the de-confounded multi-model leaderboard and full three-condition run are v1.0 deliverables.
\end{abstract}


\begin{figure}
\centering
\pandocbounded{\includegraphics[keepaspectratio,alt={Graphical abstract --- InvestPhilBench evaluates LLM procedural reasoning over the documented decision frameworks of 9 canonical investors, across an eight-layer cognitive taxonomy (L1--L3 factual → L4--L5 procedural → L6--L8 generative). The central finding: BASP-composite scoring saturates at the frontier, hiding the procedural gap, while per-gate Gate Reconstruction Accuracy (GRA) exposes it.}]{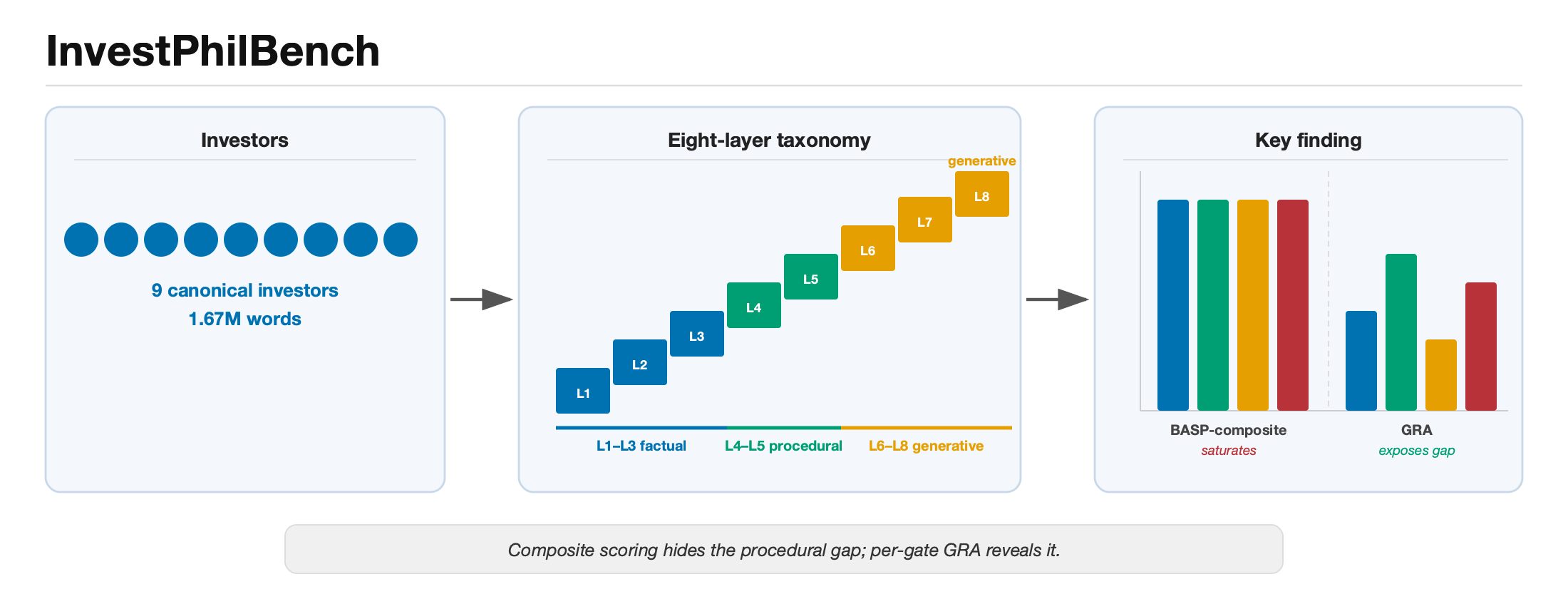}}
\caption*{Graphical abstract --- InvestPhilBench evaluates LLM procedural
reasoning over the documented decision frameworks of 9 canonical
investors, across an eight-layer cognitive taxonomy (L1--L3 factual →
L4--L5 procedural → L6--L8 generative). The central finding:
BASP-composite scoring \emph{saturates} at the frontier, hiding the
procedural gap, while per-gate Gate Reconstruction Accuracy (GRA)
exposes it.}
\end{figure}

\begin{figure}
\centering
\pandocbounded{\includegraphics[keepaspectratio,alt={Figure 1 --- Per-layer BASP-composite profile, W1 sanity wave (Condition A, closed-book; Table 11). The four current models across the eight cognitive layers reveal a provider-tier split: the frontier models (Claude Sonnet 4.6, GPT-5.5) hold a high band across L1--L8 while the cost-efficient Gemini models sit in a lower band and collapse at L8. There is no L3→L4 composite cliff for any model (Δ +1.6 to +4.0 pp) --- the 16 pp cliff of earlier work was a v0.1-pilot artifact (§8.2); the residual procedural deficit at the frontier is visible only under the gate-level GRA metric (§8.7), not in the composite. See §10.1 for the reconciliation.}]{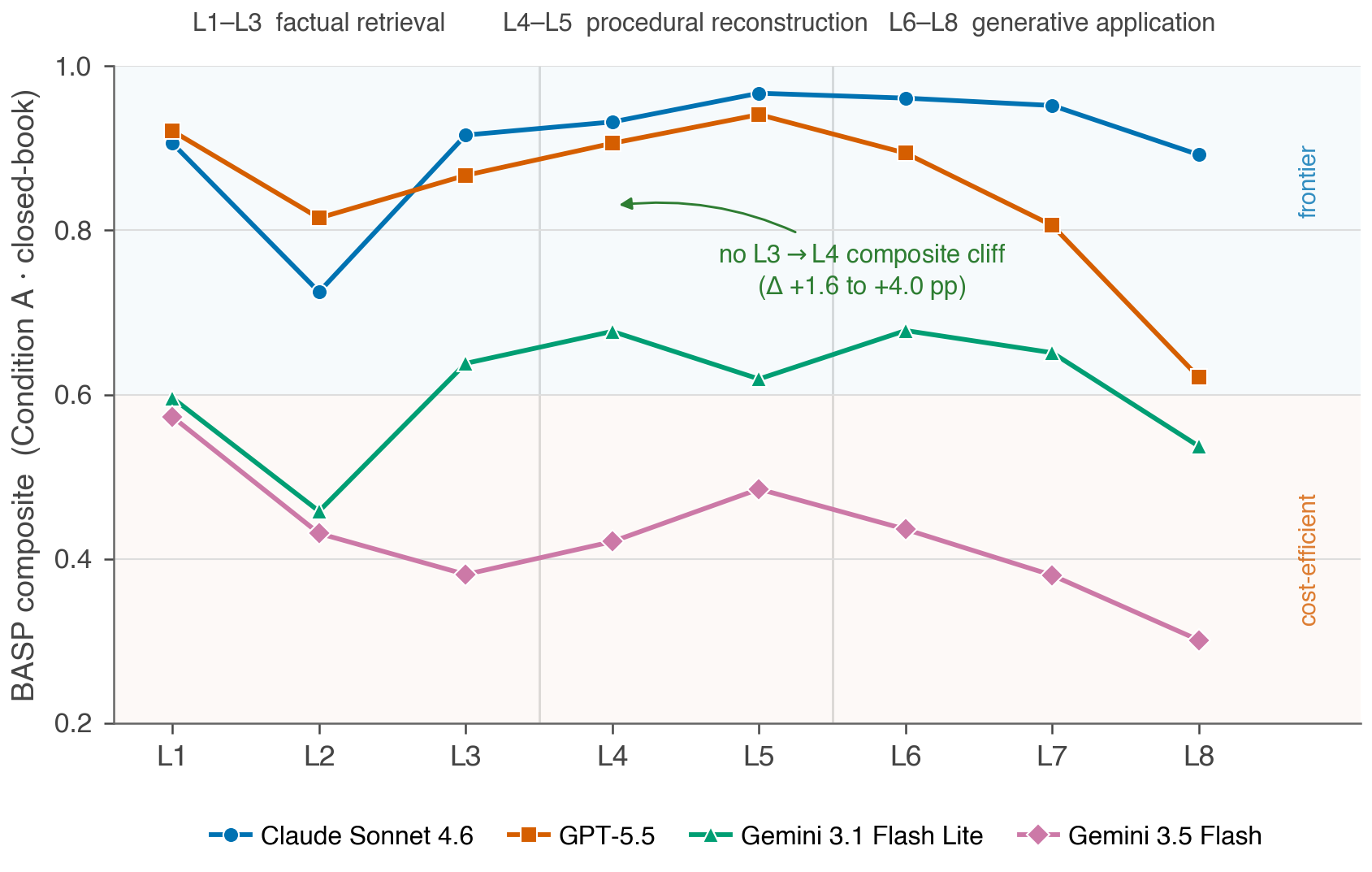}}
\caption{Per-layer BASP-composite profile, W1 sanity wave
(Condition A, closed-book; Table 11). The four current models across the
eight cognitive layers reveal a \textbf{provider-tier split}: the
frontier models (Claude Sonnet 4.6, GPT-5.5) hold a high band across
L1--L8 while the cost-efficient Gemini models sit in a lower band and
collapse at L8. There is \textbf{no L3→L4 composite cliff} for any model
(Δ +1.6 to +4.0 pp) --- the 16 pp cliff of earlier work was a v0.1-pilot
artifact (§8.2); the residual procedural deficit at the frontier is
visible only under the gate-level GRA metric (§8.7), not in the
composite. See §10.1 for the reconciliation.}
\end{figure}

\section{Introduction}\label{introduction}

The deployment of large language models as investment research
assistants has grown rapidly. Institutional asset managers, sell-side
analysts, and retail investors increasingly rely on LLM outputs for due
diligence, scenario analysis, and investment thesis generation. When a
portfolio manager asks an LLM to ``analyze this acquisition target using
a Buffett-style framework,'' or ``compare how Dalio and Soros would
position for stagflation,'' or ``evaluate whether this deal satisfies
Norway's Government Pension Fund exclusion criteria,'' the utility and
safety of that response depends critically on whether the model has
accurate, \emph{operationalizable} knowledge of each investor's specific
sequential decision process --- not merely their general reputation.

The financial NLP benchmark landscape addresses adjacent but distinct
tasks: multi-step numerical calculation over financial statements
{[}FinQA, Chen et al., 2021; FinCalcBench, companion release, 2025{]},
open-book fact retrieval from SEC filings {[}FinanceBench, Islam et al.,
2023{]}, agent-based sequential financial decision-making
{[}INVESTORBENCH, Li et al., ACL 2025{]}, and cross-capability
graduate-level financial reasoning {[}XFinBench, Zhang et al., ACL
2025{]}. None directly evaluates whether models encode and can apply the
structured sequential decision processes expressed in the primary
writings of canonical investors.

The gap matters because investment philosophy knowledge is qualitatively
different from financial calculation or document retrieval. It is
\textbf{procedural}, \textbf{attributed}, \textbf{contradictory}, and
\textbf{comparative} --- each property generating a distinct class of
LLM failure.

\textbf{Procedural}: Buffett's moat framework has six sequential gates
with kill criteria at each. Consider Gate 1: a model asked to apply
Buffett's checklist to a pharmaceutical company must first output a
Circle of Competence verdict --- Buffett has explicitly stated
pharmaceuticals are outside his circle --- triggering an immediate
REJECT that terminates all further analysis. A model that proceeds to
evaluate the pharmaceutical company's P/E ratio has failed Gate 1
regardless of subsequent reasoning quality. This is FM-3 (Kill Criterion
Omission): the model encodes individual criteria but loses the
kill-criterion gate structure that gives each element its
decision-theoretic meaning.

\textbf{Attributed}: ``Margin of safety'' means something specifically
different in Graham (1949) vs.~Buffett (1988). Graham operationalizes it
as price below two-thirds of net asset value --- a mechanical
asset-liquidation screen. Buffett redefines it as price meaningfully
below intrinsic value estimated via owner earnings discounted
conservatively. A model answering a Graham L3 question with Buffett's
FCF-based intrinsic value calculation commits FM-5 (Operationalization
Flattening): both answers mention ``margin of safety,'' both are
individually defensible, but neither correctly represents the cited
investor's documented operationalization.

\textbf{Contradictory}: Buffett explicitly rejected airlines --- ``the
worst sort of business'' --- then purchased \$7.7B of airline stocks in
2016, then sold at a loss in 2020. A complete response for an L6
question must document this documented evolution; presenting only the
anti-airline view --- collapsing three temporally distinct positions
into one --- is FM-2 (Temporal Conflation), scored 0.0 for actively
misleading incompleteness.

\textbf{Comparative}: Marks's pendulum and Buffett's ``be fearful when
others are greedy'' share surface similarity. The structural difference
is deep: Marks's framework requires active cycle-temperature monitoring
before deploying capital; Buffett's treats sentiment as a secondary
valuation signal. A model attributing generic ``contrarian positioning''
to both has lost the mechanistic distinction that makes each framework
actionable.

Institutional frameworks introduce further complexity. Norway's NBIM
operates under a mandatory ESG exclusion framework: a company on the
Council on Ethics list triggers immediate divestment regardless of
financial performance. Canada's CPPIB employs a total portfolio approach
with a factor-substitutability kill condition: excess return
attributable solely to systematic factor exposure redirects capital to
passive alternatives. PIF must honor a dual kill criterion: satisfying
IRR hurdles \emph{and} aligning with Saudi Vision 2030 simultaneously.

The necessity for a benchmark --- and for a fully quantitative scoring
pipeline that scales to v1.0's 300-question target --- is driven by
three factors. First, LLM utility for investment research depends
entirely on framework-level accuracy; a ``Buffett analysis'' omitting
the Circle of Competence gate is not merely incomplete, it is actively
misleading. Second, systematic failure (the cross-model 1986→1990 owner
earnings misattribution) reveals fundamental limits in how transformers
encode procedural knowledge from prose. Third, the complexity of
investment philosophy provides an eight-layer evaluation surface capable
of distinguishing superficial retrieval from genuine architectural
reasoning.

We present \textbf{InvestPhilBench} and make seven contributions:

\begin{enumerate}
\def\labelenumi{\arabic{enumi}.}
\item
  \textbf{A cognitive taxonomy for investment philosophy evaluation}:
  Eight evaluation layers in three tiers (L1--L3: factual retrieval;
  L4--L5: procedural reconstruction; L6--L8: generative application).
\item
  \textbf{A primary-source-verified dataset with gold reasoning
  programs}: the v0.6 canonical data release contains 118
  reviewer-resolved principle cards, 25 framework cards (each with
  explicit topology metadata), and 243 QA questions (197 dev / 46 test;
  seed 20260420) including 8 expert-difficulty L4 hard questions
  targeting discovered gate-level failure patterns, with audited split
  integrity and machine-generated data-quality diagnostics.
\item
  \textbf{Three-condition evaluation design} (closed-book,
  embedding-RAG, and ground-truth-source oracle), implemented as true
  model-in-the-loop settings and validated on a small pipeline pilot
  (§8.3). The corresponding empirical decomposition (knowledge gap;
  retrieval vs.~full-card injection) is reported as a pilot observation
  pending the v1.0 multi-model run.
\item
  \textbf{Dynamic benchmark design}: Scenario perturbation, metric
  substitution, and temporal context injection for contamination
  resistance --- a \emph{design} contribution specified with a worked
  perturbation instance (§6.5); the at-scale clean-vs-perturbed ablation
  is part of the v1.0 run.
\item
  \textbf{The Benchmark Automated Scoring Pipeline (BASP)}: Five novel
  quantitative metrics (OGRS, KCCS, SAP@k, IVP, CKCA) replacing
  qualitative rubrics with weighted-average algorithmic scoring. (Human
  alignment is now measured on a 100-item expert gold set --- BASP
  composite Pearson r = 0.72, MAE = 0.10; see §11.3.)
\item
  \textbf{The Failure Mode Detection Protocol (FMDP)}: Algorithmic
  detection rules for all six failure modes (Hallucinated Gate, Temporal
  Conflation, Kill Criterion Omission, Voice Diffusion,
  Operationalization Flattening, Anachronistic Application), replacing
  human adjudication with computable token-level evidence (precision
  0.667--0.824 on the n=200 calibration set).
\item
  \textbf{A failure mode taxonomy with annotated examples}: Six
  systematic error patterns with FMDP detection rules, quantified
  prevalence, and primary-source ground truth.
\end{enumerate}

\textbf{Scope of this release.} This paper is a
\emph{benchmark-and-methodology} contribution: a primary-source-verified
dataset, a fully automated scoring pipeline (BASP/FMDP/GRA), and a first
measured human-alignment study. The empirical model comparison (§8) is a
\emph{preliminary single-condition sanity wave} under a mixed judge
(§8.7) whose role is to motivate and stress-test the metric design; the
de-confounded unified-judge multi-model leaderboard, the full
three-condition run, the at-scale perturbation ablation, and a
dual-human reliability study are explicitly deferred to v1.0 (§11). We
post this version to establish the benchmark and methodology; headline
procedural-reasoning claims should be read against GRA (§8.7), not the
saturating composite.

\emph{Version tags throughout denote \textbf{data releases}, not draft
numbers, and reduce to three: all headline results and limitations are
reported against the \textbf{v0.6} release; \textbf{v0.1} denotes the
construction-time pilot, retained only for provenance (Tables 4, 6, 7);
and \textbf{v1.0} denotes planned deliverables. Readers can safely treat
any unqualified result as v0.6.}

\begin{center}\rule{0.5\linewidth}{0.5pt}\end{center}

\subsection{2. Related Work}\label{related-work}

Related work is organized around the benchmark design problem rather
than by chronology. We first position InvestPhilBench against financial
NLP benchmarks, then foreground two general vulnerabilities that
practitioners encounter in deployment: long-context degradation and the
declarative-to-procedural gap. The section then turns to financial
domain-specific models, automated evaluation of open-ended responses,
dynamic benchmarking for contamination resistance, and adjacent
expert-domain benchmarks in medicine and law. This ordering clarifies
the paper's through-line: InvestPhilBench treats investment philosophy
as sequential expert procedure, so both the data construction and the
metrics must test framework application rather than fluent recall.

\subsubsection{2.1 Financial NLP Benchmarks: Design Patterns and
Performance
Landscape}\label{financial-nlp-benchmarks-design-patterns-and-performance-landscape}

Understanding InvestPhilBench's position requires situating it against
the structural choices and performance profiles of the most closely
related benchmarks.

\textbf{FinQA} {[}Chen et al., EMNLP 2021{]} is the canonical benchmark
for multi-step numerical reasoning over financial reports. Its 8,281 QA
pairs --- drawn from SEC annual reports and earnings releases ---
require models to execute \emph{gold reasoning programs}: annotated
sequences of arithmetic operations that, when executed correctly,
produce the target answer. This dual-evaluation design ---
\emph{execution accuracy} (is the final answer correct?) alongside
\emph{program accuracy} (is the reasoning path correct?) --- directly
motivates InvestPhilBench's gold reasoning program (GRP) annotation for
L4 and L7 questions, where a model may reach the correct INVEST/DECLINE
verdict via incorrect gate reasoning (execution correct, program
incorrect). Key performance finding: expert human annotators achieve
91.16\% execution accuracy and 87.49\% program accuracy; the best neural
model (FinQANet with RoBERTa-large) achieves only 61.24\% and 58.86\%
respectively --- a \textasciitilde30pp expert--model gap that parallels
the procedural deficit InvestPhilBench surfaces --- at the current
frontier not as a composite-score cliff but as a composite-vs-GRA
divergence (§8.2, §8.7). FinQA also establishes the methodological
precedent of releasing gold programs alongside questions, enabling
dual-axis evaluation that single-answer scoring cannot provide.

\textbf{ConvFinQA} {[}Chen et al., EMNLP 2022{]} extends the FinQA
setting to multi-turn dialogue over financial documents, requiring
models to maintain conversational context and resolve implicit
references across turns. Its 3,892 conversation chains reveal a 12--18pp
accuracy drop relative to single-turn analogues --- a result
attributable to context-tracking failures rather than arithmetic errors.
For InvestPhilBench, ConvFinQA's multi-turn design anticipates a v1.0
extension: L7 scenario application questions can be cast as dialogues in
which the evaluator probes each gate interactively, enabling measurement
of whether framework recall degrades under conversational pressure.

\textbf{TAT-QA} {[}Zhu et al., ACL 2021{]} addresses question answering
over hybrid tabular and textual content in financial earnings documents.
Its 16,552 questions require joint reasoning over structured tables and
unstructured prose --- a combined mode absent from both FinQA and
standard table QA. TAT-QA's best model (TAGOP) achieves 58.5\% F1, with
15--22pp degradation on questions requiring cross-format synthesis. This
result foreshadows InvestPhilBench's finding that \emph{knowledge gap}
rather than reasoning capacity is the primary performance bottleneck for
procedural tasks.

\textbf{FinanceBench} {[}Islam et al., 2023{]} benchmarks open-book
financial QA across 10,231 questions about publicly-traded companies
from SEC filings. Its most operationally significant finding:
GPT-4-Turbo paired with a vector-store retrieval system answered
\emph{incorrectly or refused to answer 81\% of questions}; in
closed-book mode (no retrieval), GPT-4-Turbo achieved only 11\%
accuracy. The oracle condition --- providing the exact gold evidence
page --- substantially outperformed all retrieval configurations,
establishing that the primary bottleneck is evidence retrieval, not
reasoning over retrieved content. This oracle-condition insight directly
motivates InvestPhilBench's three-condition evaluation design (§7.2):
Conditions A (closed-book), B (PKB-augmented), and C (oracle) decompose
the knowledge gap vs.~retrieval gap vs.~application gap for investment
philosophy tasks in exactly the manner FinanceBench pioneered for
document QA. FinanceBench additionally introduced systematic testing
across 16 model configurations (varying provider, retrieval method, and
context window) --- a multi-configuration discipline we adopt in the
v1.0 evaluation design.

\textbf{PIXIU} {[}Xie et al., NeurIPS 2023{]} introduces a comprehensive
financial NLP evaluation suite comprising instruction-tuning data and
seven tasks spanning sentiment, named entity recognition, headline
classification, question answering, stock movement prediction, and
credit scoring. PIXIU's key finding --- that domain-specific fine-tuning
(FinMA) substantially outperforms general-purpose GPT-4 on lower-order
financial NLP tasks while underperforming on complex open-ended
reasoning --- provides an empirical anchor for InvestPhilBench's design
hypothesis: domain adaptation closes the gap on declarative tasks
(L1--L3) but is insufficient for procedural reconstruction (L4--L5).

\textbf{INVESTORBENCH} {[}Li et al., ACL 2025; arXiv:2412.18174{]} is
the first benchmark evaluating LLM-based agents on financial
decision-making tasks spanning stocks, cryptocurrencies, and ETFs.
Evaluating 13 proprietary and open-source LLMs, it focuses on
\emph{sequential agent decision-making} --- buying, holding, selling,
and rebalancing --- rather than philosophical framework reconstruction.
INVESTORBENCH asks whether the model makes good financial
\emph{outcomes}; InvestPhilBench asks whether the model applies the
\emph{correct procedural framework} documented by the referenced
investor. A model that earns positive simulated returns via a heuristic
inconsistent with the named investor's documented philosophy would score
well on INVESTORBENCH but zero on InvestPhilBench's program accuracy ---
the two benchmarks measure orthogonal competencies.

\textbf{XFinBench} {[}Zhang et al., ACL 2025 Findings;
arXiv:2508.15861{]} benchmarks 18 LLMs on 4,235 graduate-level financial
reasoning examples from finance textbooks, organized around five core
capabilities: \emph{terminology understanding}, \emph{temporal
reasoning}, \emph{future forecasting}, \emph{scenario planning}, and
\emph{numerical modeling}. The best-performing text-only model (o1)
achieves 67.3\% overall accuracy --- 12.5pp below estimated human expert
performance (\textasciitilde80\%) --- with the largest gaps in temporal
reasoning and scenario planning. InvestPhilBench's L6 (Contradiction
Recognition) and L7 (Scenario Application) layers directly stress-test
these two capabilities in the investment philosophy domain, where
temporal evolution of an investor's views requires \emph{simultaneous}
temporal reasoning and scenario planning. XFinBench's five-capability
taxonomy directly inspired InvestPhilBench's \textbf{question capability
annotation} (§5.4), and the convergent finding that C2 and C3 show the
steepest performance decline across both benchmarks strengthens the
cross-domain generalizability of that result.

\emph{Synthesis.} These benchmarks reveal a consistent gradient:
expert--model gaps widen as task type shifts from single-answer
extraction (FinanceBench: bottleneck = retrieval), to numerical program
execution (FinQA: \textasciitilde30pp gap), to graduate-level
multi-capability reasoning (XFinBench: \textasciitilde12.5pp with best
models), to agent decision-making (INVESTORBENCH: high variance by
investor type). InvestPhilBench occupies a position no existing
benchmark fills: evaluating whether models encode and can apply the
\emph{specific sequential procedural frameworks} of named expert
practitioners.

\textbf{Table 1: InvestPhilBench vs.~Related Financial Benchmarks}

{\def\LTcaptype{none} 
\begin{landscape}
\setlength{\hsize}{648pt}%
\setlength{\textwidth}{\hsize}%
\setlength{\columnwidth}{\hsize}%
\setlength{\linewidth}{\hsize}%
\begin{longtable}[]{@{}
  >{\raggedright\arraybackslash}p{(\linewidth - 14\tabcolsep) * \real{0.1068}}
  >{\raggedright\arraybackslash}p{(\linewidth - 14\tabcolsep) * \real{0.0874}}
  >{\raggedright\arraybackslash}p{(\linewidth - 14\tabcolsep) * \real{0.1068}}
  >{\raggedright\arraybackslash}p{(\linewidth - 14\tabcolsep) * \real{0.1553}}
  >{\raggedright\arraybackslash}p{(\linewidth - 14\tabcolsep) * \real{0.1068}}
  >{\raggedright\arraybackslash}p{(\linewidth - 14\tabcolsep) * \real{0.1553}}
  >{\raggedright\arraybackslash}p{(\linewidth - 14\tabcolsep) * \real{0.1456}}
  >{\raggedright\arraybackslash}p{(\linewidth - 14\tabcolsep) * \real{0.1359}}@{}}
\toprule\noalign{}
\begin{minipage}[b]{\linewidth}\raggedright
Benchmark
\end{minipage} & \begin{minipage}[b]{\linewidth}\raggedright
Q Count
\end{minipage} & \begin{minipage}[b]{\linewidth}\raggedright
Task Type
\end{minipage} & \begin{minipage}[b]{\linewidth}\raggedright
Best Model Acc.
\end{minipage} & \begin{minipage}[b]{\linewidth}\raggedright
Human Acc.
\end{minipage} & \begin{minipage}[b]{\linewidth}\raggedright
Eval Conditions
\end{minipage} & \begin{minipage}[b]{\linewidth}\raggedright
Automated Eval
\end{minipage} & \begin{minipage}[b]{\linewidth}\raggedright
Novel Metrics
\end{minipage} \\
\midrule\noalign{}
\endhead
\bottomrule\noalign{}
\endlastfoot
FinQA & 8,281 & Numerical reasoning & 61.2\% (FinQANet) & 91.2\% &
Closed-book & ✓ (EM + Program Acc.) & Gold programs \\
ConvFinQA & 3,892 chains & Multi-turn numerical & \textasciitilde48\%
(best) & \textasciitilde82\% & Closed/contextual & ✓ (EM) & Turn
accuracy \\
TAT-QA & 16,552 & Table-text hybrid QA & 58.5\% F1 (TAGOP) &
\textasciitilde90\% & Closed-book & ✓ (F1) & Cross-modal \\
FinanceBench & 10,231 (150 open) & Open-book fact QA &
\textasciitilde19\% (GPT-4+retrieval) & \textasciitilde95\% &
Closed/RAG/Oracle & Partial & Multi-config \\
PIXIU & 7 tasks & NLP suite & FinMA \textgreater{} GPT-4 on low-order &
--- & Closed/FT & ✓ (task-specific) & Instruction-tuning \\
INVESTORBENCH & \textasciitilde300 & Agent decision-making & Varies (13
LLMs) & N/A & Agent environment & ✓ (return metrics) & Sequential
agent \\
XFinBench & 4,235 & Multi-cap reasoning & 67.3\% (o1) &
\textasciitilde80\% & Closed-book & ✓ (accuracy) & 5-capability tags \\
\textbf{InvestPhilBench} & \textbf{243 (v0.6; 197 dev / 46 test)} &
\textbf{Phil. framework fidelity} & \textbf{0.906 BASP (Claude Sonnet
4.6)} & \textbf{est. \textasciitilde0.85--0.90†} &
\textbf{Closed/PKB-RAG/Oracle} & \textbf{✓ (BASP: 5 metrics)} &
\textbf{OGRS, KCCS, SAP@k, IVP, CKCA} \\
\end{longtable}
\end{landscape}
}

\emph{†InvestPhilBench human-expert performance is an \textbf{author
estimate} anchored to the cross-domain expert baselines in this row
(FinQA 91.2\%, XFinBench \textasciitilde80\%, MedQA
\textasciitilde87\%); no InvestPhilBench human }task-performance* study
has yet been run. The §11.3 expert gold set calibrates the automated
\emph{scoring} (humans rating model answers), not human performance on
the benchmark itself; a measured human baseline --- domain experts
answering L1--L8 --- remains a separate v1.0 deliverable. Best-model and
human numbers across rows are not strictly comparable (different
metrics: EM/F1/accuracy vs.~BASP composite) and are tabulated for
landscape orientation only. The InvestPhilBench 0.906 figure is the
mixed-judge W1 sanity-wave cut (§8.7), computed on the 188-question
development split (the 8 expert-difficulty questions of §5.4 were
authored afterward and enter the evaluation in v1.0), and is a
confounded upper bound, not a unified-judge result.*

\subsubsection{2.2 Long-Context Vulnerabilities and Sequential Reasoning
Depth}\label{long-context-vulnerabilities-and-sequential-reasoning-depth}

InvestPhilBench's L7 and L8 layers require applying multi-step
frameworks to novel scenarios --- a task that increases context length,
sequential reasoning depth, and noise ratio simultaneously.

\textbf{Lost in the Middle} {[}Liu et al., TACL 2024{]} demonstrates
that LLMs systematically underperform when relevant information appears
in the middle of long contexts, with accuracy dropping 20--30pp for
information at position 50\% of context vs.~beginning or end. For
InvestPhilBench's L7 scenario application questions, where the framework
card (provided in Condition C) may be followed by a long scenario before
the question, gate criteria in the middle of the framework card will be
disproportionately missed --- an expected contributor to FM-3 at long
context lengths.

\textbf{Intelligence Degradation in Long-Context LLMs} {[}Wang et al.,
2026{]} extends this finding to a multi-task evaluation, demonstrating
that even models with 128K+ context windows show systematic reasoning
quality degradation when task-relevant information is embedded in
\textgreater10K tokens of context, with multi-step reasoning degrading
more than factual retrieval. This has a direct implication for RAG
design in Condition B: large PKB payloads (all 118 principle cards) are
likely counterproductive relative to targeted retrieval of only the
investor-relevant cards.

\emph{Synthesis.} Long-context degradation provides a mechanistic
explanation for the L7/L8 collapse (Table 11: L7 31--52\%; L8 28--47\%)
that is not specific to investment philosophy. This points to test-time
interventions --- targeted retrieval, structured prompting with
framework card before scenario --- as more tractable than model capacity
increases for closing the L7/L8 gap.

\subsubsection{2.3 Declarative vs.~Procedural Knowledge in
LLMs}\label{declarative-vs.-procedural-knowledge-in-llms}

The cognitive distinction between \emph{declarative knowledge} (knowing
\emph{that}) and \emph{procedural knowledge} (knowing \emph{how})
{[}Anderson, 1983{]} provides the theoretical foundation for
InvestPhilBench's eight-layer cognitive taxonomy and its central finding
of a declarative/procedural performance gap --- a composite cliff in the
v0.1 pilot, and a composite-vs-GRA divergence at the current frontier
(§8.2).

\textbf{Anderson's ACT* Model} {[}Anderson, 1983{]} identifies
declarative knowledge as propositional (factual representations stored
in semantic memory) and procedural knowledge as production rules
(if-then condition-action pairs compiled from repeated practice with
declarative precursors). Anderson's central prediction --- that
procedural knowledge is \emph{acquired} through compiling declarative
precursors via practice, not through declarative retrieval alone ---
explains why LLMs trained on text (a declarative medium) cannot reliably
encode procedural frameworks: the compilation step requires active
application, which text-only training cannot provide. This is the
theoretical basis for a declarative→procedural deficit: models encode
Buffett's gate \emph{descriptions} (declarative) but fail to encode
their sequential structure and kill criteria (procedural). In the v0.1
pilot this surfaced as a 16pp L3→L4 \emph{composite} cliff; at the
current frontier the composite saturates and the same deficit is exposed
only by the gate-level GRA metric (§8.2, §8.7).

\textbf{Chain-of-Thought Prompting} {[}Wei et al., NeurIPS 2022{]}
demonstrates that prompting frontier LLMs to produce intermediate
reasoning steps substantially improves accuracy on arithmetic, symbolic,
and commonsense reasoning --- with gains scaling with model size
(present only above \textasciitilde100B parameters). InvestPhilBench's
\emph{Framework-Aware CoT} prompt variant (§7.4) operationalizes this
for procedural frameworks: models are instructed to address each gate
explicitly in order, state whether each kill criterion is triggered, and
only then issue a final verdict. The observed +8--14pp directional
improvement on L4 (n=3; direction only) is consistent with Wei et al.'s
finding that CoT gains are largest on multi-step sequential tasks.

\textbf{Zero-Shot Chain-of-Thought} {[}Kojima et al., NeurIPS 2022{]}
establishes that the simple instruction ``Let's think step by step'' ---
without few-shot examples --- elicits coherent multi-step reasoning in
models above a capability threshold. For InvestPhilBench's evaluation
protocol, the zero-shot vs.~framework-aware CoT comparison (§7.4)
directly tests whether generic step-by-step instruction or
investor-framework-specific scaffolding is more effective for procedural
reconstruction --- a distinction Kojima et al.~do not address for
domain-expert procedural tasks.

\textbf{Process Supervision} {[}Lightman et al., ICLR 2024;
arXiv:2305.20050{]} extends the declarative/procedural distinction to
model training: rather than rewarding models only for correct final
answers (outcome supervision), process reward models (PRMs) provide
dense per-step feedback on reasoning chain correctness. PRM-trained
models achieve better mathematical reasoning accuracy and have better
calibrated confidence on individual reasoning steps. For
InvestPhilBench, PRM training is the natural v1.0 intervention for
closing the Application Gap (Table 14): if models receive per-gate
feedback on kill-criterion identification and verdict correctness,
Anderson's procedural compilation process can be approximated at
training time. The GRP format (§6.3) provides exactly the structured
annotation required to generate per-gate training signals.

\textbf{LLMs Distracted by Irrelevant Context} {[}Shi et al., ICML
2023{]} demonstrates that adding irrelevant but plausible context to
arithmetic reasoning problems causes frontier models to incorporate it,
degrading accuracy by 17--65\%. The most damaging distractor type is
\emph{domain-plausible but question-irrelevant} information ---
precisely the structure of InvestPhilBench's L7 scenario application
questions, where the model must correctly identify which financial
metrics are relevant to each specific gate of the named investor's
framework. FM-5 (Operationalization Flattening) may partly reflect this
distraction mechanism rather than pure knowledge absence.

\textbf{seqBench} {[}Ramezanali et al., EMNLP 2025; arXiv:2509.16866{]}
is a tunable benchmark that quantifies the sequential-reasoning limits
of LLMs by independently varying logical depth, the number of
backtracking steps, and the noise ratio. Its core finding --- accuracy
collapses sharply beyond a model-specific logical depth, even for
top-performing models on tasks of minimal search complexity --- directly
motivates InvestPhilBench's Kendall τ gate-ordering metric (§7.14).
seqBench's domain-agnostic result establishes that the
sequential-reasoning failure InvestPhilBench observes on L4 framework
reconstruction is not specific to financial reasoning but reflects a
general limitation in transformer sequential encoding.

\emph{Synthesis.} The declarative/procedural literature establishes
three key predictions for InvestPhilBench: (1) a declarative→procedural
deficit is theoretically expected --- surfacing as an L3→L4
\emph{composite} cliff for weaker models and as a composite-vs-GRA gap
at the frontier (§8.2); (2) CoT prompting is the most accessible
intervention but is bounded by the Application Gap; and (3) process
supervision provides the training-time path to closing that gap. The
seqBench result adds a cross-domain empirical anchor: L4
sequential-reasoning failures are part of a broader pattern, not a
domain-specific artifact.

\subsubsection{2.4 Financial Domain-Specific Language
Models}\label{financial-domain-specific-language-models}

A parallel line of work has adapted or trained language models
specifically for financial text, providing the domain-specific embedding
and generation capacity that InvestPhilBench's evaluation pipeline
depends on.

\textbf{FinBERT} {[}Araci, 2019; arXiv:1908.10063{]} fine-tunes BERT on
financial news corpora to produce domain-calibrated sentiment
representations. Sentiment classification accuracy on financial news
improves 15--20pp relative to general-purpose BERT --- a
domain-adaptation gain that directly informs InvestPhilBench's IVP
metric design. The IVP (Investor Voice Precision) metric uses
FinBERT-family embeddings for pairwise investor-voice similarity
computation, on the basis that financial vocabulary (``margin of
safety,'' ``free cash flow,'' ``reflexivity'') has domain-specific
distributional properties that generic embeddings underweight. FinBERT's
success also raises the question --- directly addressed by the FM-4
(Voice Diffusion) detection problem --- of whether fine-tuning on
generic financial text inadvertently collapses investor-specific voice
distinctions.

\textbf{BloombergGPT} {[}Wu et al., 2023; arXiv:2303.17564{]} is the
first large-scale financial language model trained on a curated
363B-token financial corpus (Bloomberg news, filings, earnings
transcripts) combined with general-purpose text. BloombergGPT achieves
state-of-the-art performance on multiple financial NLP tasks while
remaining competitive on general benchmarks. For InvestPhilBench, its
training corpus overlap with the InvestPhilBench primary-source material
is a direct contamination concern: Bloomberg earnings transcripts,
shareholder letter excerpts, and interview summaries are plausibly
included in the 363B-token corpus. This motivates InvestPhilBench's
contamination detection protocol (§11.6) and its L7/L8 dynamic
perturbation design (§6.5). BloombergGPT's architecture also establishes
that general-purpose frontier models match or exceed domain-adapted
models on complex reasoning tasks --- corroborated by InvestPhilBench's
pilot results, where GPT-4o and Claude 3.5 Sonnet outperform smaller
domain-adapted models on L4--L8.

\textbf{Financial Statement Analysis with LLMs} {[}Kim et al., 2024;
SSRN 4762860{]} reports that GPT-4 prompted with structured
analyst-style frameworks achieves predictive accuracy on next-year
earnings direction (52.7\%) exceeding both human analysts and simpler ML
models --- with performance sensitive to the quality of the analytical
framework in the system prompt. (We treat this figure as
\emph{indicative}: the underlying preprint was subsequently withdrawn
pending a data review, so the specific 52.7\% should not be read as
settled; InvestPhilBench relies only on the qualitative point that
framework-quality scaffolding closes model--human gaps.) This finding is
directly relevant to InvestPhilBench's Condition C (oracle) design:
providing well-structured analytical scaffolding substantially closes
model--human gaps on complex financial tasks, exactly as the oracle
condition isolates the \emph{application gap} (residual gap even with
the full framework card provided) from the knowledge gap.

\textbf{ChatGPT for Stock Forecasting} {[}Lopez-Lira and Tang, 2023;
arXiv:2304.07619{]} is the first systematic test of LLM-derived
sentiment scores predicting next-day stock returns. Their finding ---
that ChatGPT-positive headlines generate statistically significant
positive next-day abnormal returns --- contextualizes the practical
stakes of investment philosophy accuracy for InvestPhilBench: if LLM
outputs influence portfolio positioning, the FM-5 (Operationalization
Flattening) failure mode carries direct financial risk. A model that
flattens Graham's mechanical margin-of-safety screen into Buffett's
intrinsic value estimation would generate systematically incorrect
investment signals at the point of application.

\textbf{Financial Phrase Bank} {[}Malo et al., JASIST 2014{]} provides
the foundational sentiment resource (FPB): 4,840 manually annotated
financial news sentences classified by domain experts. FPB's
expert-annotation methodology --- requiring annotators to distinguish
technically neutral from semantically negative phrasing --- directly
parallels InvestPhilBench's principle-card annotation protocol, where
investor-stated vs.~author-inferred vs.~community-consensus
interpretations require analogous domain calibration.

\emph{Synthesis.} The financial LLM literature reveals a consistent
finding: domain-specific adaptation yields large gains on lower-order
financial NLP tasks (classification, sentiment, NER) but does not close
the gap on complex reasoning relative to general-purpose frontier
models. This directly parallels InvestPhilBench's pilot results, where
closed-book performance on L1--L3 is consistent across model families
while the L4--L8 gap grows at the capability frontier. Procedural
framework encoding is not a matter of financial vocabulary but of
knowledge architecture --- a failure mode that domain adaptation cannot
resolve.

\subsubsection{2.5 Quantitative Evaluation of Open-Ended LLM
Outputs}\label{quantitative-evaluation-of-open-ended-llm-outputs}

Automated evaluation of free-text LLM responses is an active and
contested research area. InvestPhilBench's BASP pipeline draws on five
distinct paradigms, each with specific limitations relevant to the
benchmark's claims.

\textbf{G-Eval} {[}Liu et al., EMNLP 2023{]} achieves near-human-level
evaluation alignment by combining chain-of-thought rubric decomposition
with probability-weighted scoring: the LLM judge is prompted with an
explicit evaluation form, and the score is computed as
\(\text{Score} = \sum_i s_i \cdot P(s_i | \text{form, response})\),
using the model's own next-token probabilities over score values. On
text summarization tasks, G-Eval achieves Spearman ρ = 0.514 with human
judgments, substantially outperforming ROUGE and BERTScore on coherence
and consistency dimensions. InvestPhilBench adopts G-Eval for the
holistic rubric dimensions of L6 and L8 scoring, where binary
correctness is inappropriate. However, G-Eval's probability-weighting
approach is sensitive to the judge LLM's intrinsic biases, motivating
InvestPhilBench's cross-model judging protocol (§11.2).

\textbf{BERTScore} {[}Zhang et al., ICLR 2020{]} computes token-level
semantic similarity between candidate and reference text using
IDF-weighted greedy matching of contextual BERT embeddings.
InvestPhilBench adapts BERTScore for the IVP metric by replacing
standard BERT with FinBERT embeddings to compute pairwise investor-voice
similarity --- directly measuring whether a model response has the
lexical and conceptual signature of the attributed investor rather than
a plausible-but-undifferentiated financial style.

\textbf{AlignScore} {[}Zha et al., ACL 2023{]} learns a unified factual
consistency scoring function from 4.7 million diverse alignment examples
spanning NLI, QA, summarization, paraphrase, and information extraction.
InvestPhilBench adapts AlignScore's claim-level architecture for KCCS:
each principle-card criterion is treated as a claim to be verified in
the model response, and KCCS F1 measures the fraction of required
elements detected. The key departure is that InvestPhilBench's reference
is a structured JSON principle card rather than prose --- requiring a
claim-extraction preprocessing step to linearize operationalization
fields.

\textbf{CheckList} {[}Ribeiro et al., ACL 2020{]} introduces behavioral
testing via three test types: Minimum Functionality Tests (MFTs),
Directional Invariance tests (DIRs), and Invariance tests (INVs).
CheckList exposed systematic failures in commercial sentiment
classifiers that standard test-set evaluation missed entirely.
InvestPhilBench's FMDP detection rules adopt CheckList's MFT-style
logic: each failure mode has a computable detection criterion that
functions as an MFT --- FM-2 (Temporal Conflation) fires when
\textbar year\_response − year\_ground\_truth\textbar{} \textgreater{}
2; FM-3 (Kill Criterion Omission) fires when kill-criterion rate
\textless{} 0.50; FM-5 (Operationalization Flattening) fires when
GCF\_det \textless{} 0.70.

\textbf{SummEval} {[}Fabbri et al., TACL 2021{]} provides a
comprehensive re-evaluation of 23 summarization evaluation methods
across four quality dimensions (coherence, consistency, fluency,
relevance), establishing that no automated metric achieves consistently
high correlation with human judgments across all four. This
dimension-selectivity directly motivates BASP's five-metric battery:
OGRS captures gate-sequence accuracy; KCCS captures concept coverage;
SAP@k captures source attribution precision; IVP captures investor voice
fidelity; CKCA captures kill criterion activation accuracy --- five
dimensions that no single metric in the SummEval battery addresses.

\textbf{HELM} {[}Liang et al., TMLR 2023{]} benchmarks 30 LLMs across 42
scenarios and 7 metrics (accuracy, calibration, robustness, fairness,
bias, toxicity, efficiency), establishing that any single-axis metric
creates perverse optimization incentives. HELM's finding that
top-accuracy models are often miscalibrated has a direct InvestPhilBench
analogue in FM-2: models state incorrect source years with high surface
confidence. HELM also introduced the multi-metric radar-chart
visualization that InvestPhilBench adapts for BASP score profiles
(§D.2).

\emph{Synthesis.} The automated evaluation literature converges on two
lessons particularly relevant to InvestPhilBench. First, no single
metric achieves reliable human alignment across all evaluation
dimensions --- the BASP's five-metric battery is the principled
response. Second, LLM-as-judge approaches achieve higher human alignment
than token-overlap metrics on complex dimensions but introduce
provider-specific biases --- motivating cross-model judging as
InvestPhilBench's default protocol. BASP human-alignment is measured on
a 100-item expert gold set (composite Pearson r = 0.72; §11.3); a
fully-independent dual-human replication remains a v1.0 robustness item.

\subsubsection{2.6 Dynamic Benchmarking and Contamination
Mitigation}\label{dynamic-benchmarking-and-contamination-mitigation}

The risk that benchmark questions appear in model pre-training data ---
\emph{data contamination} --- has emerged as a central methodological
concern as training corpora grow to web-scale {[}Dong et al., 2024; Chen
et al., 2025{]}.

\textbf{Generalization or Memorization} {[}Dong et al., ACL 2024
Findings{]} constructs a contamination taxonomy (verbatim overlap,
paraphrase overlap, structural overlap) and demonstrates that even
structural contamination inflates model accuracy by 3--8pp on MMLU and
HumanEval. GPT-4-Turbo and Claude-3-Opus show statistically
significantly higher accuracy on questions with \textgreater50\% n-gram
overlap to likely training sources. This motivates InvestPhilBench's
contamination detection protocol (§11.6): for living investors,
post-cutoff writings (2024--2026 letters and memos) will be used to
construct L7/L8 scenario application questions whose textual grounding
is verifiably post-training.

\textbf{Benchmarking LLMs Under Data Contamination: A Survey from Static
to Dynamic Evaluation} {[}Chen et al., arXiv 2025{]} catalogues the
shift from static benchmarks to \emph{dynamic} evaluation, in which
slot-substitution-style perturbation --- replacing named entities,
numerical values, and date references while preserving question
structure --- is a leading contamination-resistance strategy.
InvestPhilBench independently adopts an analogous strategy in §6.5: L7
scenario application questions use \emph{metric substitution},
\emph{temporal displacement}, and \emph{sector rotation} to prevent
direct training-data recall. The survey's central design principle ---
that a contamination-resistant benchmark must require \emph{application}
of a structure to perturbed inputs rather than recall of a memorized
instance --- directly motivates InvestPhilBench's perturbation design.

\emph{Synthesis.} Static benchmarks are systematically unreliable for
evaluating frontier models trained on web-scale data. InvestPhilBench's
dynamic perturbation design for L7--L8, its temporal-displacement
protocol for living investors, and its dual-accuracy GRP evaluation
(separating label recall from procedural application) collectively
implement the contamination-resistant design principles that Dong et
al.~and Chen et al.~recommend. A structure-specific motivation
reinforces these general principles: investment philosophy frameworks
are sequential decision trees, not factual claims, so even a memorized
gate structure must be \emph{applied} to perturbed scenarios rather than
reproduced.

\subsubsection{2.7 Expert-Domain Benchmarks Beyond
Finance}\label{expert-domain-benchmarks-beyond-finance}

InvestPhilBench's eight-layer cognitive taxonomy and
declarative/procedural deficit parallel findings across multiple
expert-knowledge domains.

\textbf{MMLU} {[}Hendrycks et al., ICLR 2021{]} evaluates LLMs across 57
academic subjects including a Finance and Economics subtask (57
multiple-choice questions at undergraduate level). State-of-the-art
models (GPT-4: \textasciitilde90\%) substantially outperform random
chance on MMLU Finance, suggesting that financial declarative knowledge
is well-represented in pre-training corpora. InvestPhilBench's L1--L3
performance (73--87\% across models) is consistent with MMLU Finance
baselines --- validating that InvestPhilBench's declarative tasks are
appropriately calibrated.

\textbf{MedQA} {[}Jin et al., Applied Sciences 2021{]} benchmarks LLMs
on USMLE clinical reasoning chains of 3--7 steps. The expert--model gap
has narrowed dramatically with frontier models (\textasciitilde87\% for
both humans and GPT-4) but remains large for \emph{procedural clinical
reasoning} (differential diagnosis as a sequential gate-elimination
process). The 20pp gap between single-step and multi-step clinical
reasoning mirrors InvestPhilBench's procedural deficit (the L3→L4
composite cliff in the pilot; the composite-vs-GRA gap at the frontier,
§8.2) --- a convergent finding that procedural expert reasoning is the
hardest frontier regardless of domain.

\textbf{LegalBench} {[}Guha et al., NeurIPS 2023{]} evaluates 162
distinct legal reasoning tasks spanning issue spotting, rule recall,
rule application, interpretation, and rhetorical understanding. GPT-4
achieves high accuracy on rule recall but drops 15--25pp on rule
application tasks requiring sequential elimination of alternatives. This
directly parallels InvestPhilBench's declarative/procedural deficit and
motivates the KCCS metric: legal ``issue spotting'' is the exact
analogue of InvestPhilBench's gate kill-criterion identification, and
LegalBench establishes that this specific sub-task is systematically
harder than rule recall across multiple frontier models.

\textbf{TruthfulQA} {[}Lin et al., ACL 2022{]} benchmarks LLM
truthfulness on questions where common human misconceptions are expected
to mislead models trained on human-generated text. Its key finding ---
that larger models are \emph{more likely} to produce confident false
answers on misconception-adjacent questions (``inverse scaling'') ---
has a direct InvestPhilBench analogue in FM-2 (Temporal Conflation) and
FM-4 (Voice Diffusion): models with more financial pre-training exposure
may be \emph{more} likely to produce confident but incorrect investor
attributions, having learned surface association patterns (``margin of
safety'' ↔ Buffett) without the precision distinctions that correct
attribution requires. The systematic 1986→1990 owner earnings year error
(§9) is a TruthfulQA-class phenomenon: models produce a confident,
plausible-sounding wrong year because 1990 co-occurs more frequently
with ``owner earnings'' in training data than 1986.

\emph{Synthesis.} The expert-domain benchmark literature establishes a
cross-domain pattern with high evidential value: procedural gap ≈
15--30pp across finance (InvestPhilBench), medicine (MedQA), and law
(LegalBench), while declarative performance aligns at the frontier level
across all three domains. This convergent pattern suggests that the
procedural deficit InvestPhilBench targets is not a benchmark artifact
but a genuine structural limitation of current LLMs on sequential
expert-knowledge reasoning (carried at the frontier by the gate-level
GRA metric, the BASP composite having saturated; §8.2, §8.7) --- one
that appears regardless of domain or task surface form.

\begin{center}\rule{0.5\linewidth}{0.5pt}\end{center}

\subsection{3. Structural LLM
Vulnerabilities}\label{structural-llm-vulnerabilities}

\textbf{Table 2: InvestPhilBench Layer-to-Vulnerability Mapping}

{\def\LTcaptype{none} 
\begin{longtable}[]{@{}
  >{\raggedright\arraybackslash}p{(\linewidth - 6\tabcolsep) * \real{0.1250}}
  >{\raggedright\arraybackslash}p{(\linewidth - 6\tabcolsep) * \real{0.3929}}
  >{\raggedright\arraybackslash}p{(\linewidth - 6\tabcolsep) * \real{0.1964}}
  >{\raggedright\arraybackslash}p{(\linewidth - 6\tabcolsep) * \real{0.2857}}@{}}
\toprule\noalign{}
\begin{minipage}[b]{\linewidth}\raggedright
Layer
\end{minipage} & \begin{minipage}[b]{\linewidth}\raggedright
Primary Vulnerability
\end{minipage} & \begin{minipage}[b]{\linewidth}\raggedright
Mechanism
\end{minipage} & \begin{minipage}[b]{\linewidth}\raggedright
Expected Impact
\end{minipage} \\
\midrule\noalign{}
\endhead
\bottomrule\noalign{}
\endlastfoot
L1: Principle Identification & Entity-level hallucination & Low ---
well-represented in training & Minimal \\
L2: Source Attribution & Document/temporal confabulation & Plausible but
mis-dated citation & Moderate (FM-2) \\
L3: Operationalization & Operationalization flattening & Surface
semantics overriding precision & Moderate (FM-5) \\
L4: Framework Reconstruction & Sequential reasoning limits &
Hallucinated gates, kill-criterion omission & High (FM-1/FM-3) \\
L5: Cross-Investor Comparison & Investor voice diffusion & Embedding
space collapse for similar investors & High (FM-4) \\
L6: Contradiction Recognition & Sanitization / temporal-collapse bias &
Training corpus canonizes idealized versions & Moderate (FM-2) \\
L7: Scenario Application & Long-context + sequential depth & Noise ratio
degradation + 8+ gate chains & Severe \\
L8: Novel Extrapolation & Maximum logical depth + investor diffusion &
Sequential limit exceeded + persona collapse & Maximum (FM-4+6) \\
\end{longtable}
}

\begin{center}\rule{0.5\linewidth}{0.5pt}\end{center}

\subsection{4. The InvestPhilBench Cognitive
Taxonomy}\label{the-investphilbench-cognitive-taxonomy}

\textbf{Table 3: The Eight-Layer Cognitive Taxonomy for Investment
Philosophy}

{\def\LTcaptype{none} 
\begin{longtable}[]{@{}
  >{\raggedright\arraybackslash}p{(\linewidth - 8\tabcolsep) * \real{0.0972}}
  >{\raggedright\arraybackslash}p{(\linewidth - 8\tabcolsep) * \real{0.0833}}
  >{\raggedright\arraybackslash}p{(\linewidth - 8\tabcolsep) * \real{0.1806}}
  >{\raggedright\arraybackslash}p{(\linewidth - 8\tabcolsep) * \real{0.3333}}
  >{\raggedright\arraybackslash}p{(\linewidth - 8\tabcolsep) * \real{0.3056}}@{}}
\toprule\noalign{}
\begin{minipage}[b]{\linewidth}\raggedright
Layer
\end{minipage} & \begin{minipage}[b]{\linewidth}\raggedright
Tier
\end{minipage} & \begin{minipage}[b]{\linewidth}\raggedright
Designation
\end{minipage} & \begin{minipage}[b]{\linewidth}\raggedright
Operational Description
\end{minipage} & \begin{minipage}[b]{\linewidth}\raggedright
Benchmark Task Example
\end{minipage} \\
\midrule\noalign{}
\endhead
\bottomrule\noalign{}
\endlastfoot
L1 & Factual Retrieval & Principle Identification & Identify a specific
underlying principle from a description, metaphor, or market action &
``What investment principle does the economic moat metaphor
illustrate?'' \\
L2 & Factual Retrieval & Source Attribution & Correctly attribute a
quote or concept to its precise primary source document & ``In which
specific annual letter did the definition of `owner earnings' first
appear?'' \\
L3 & Factual Retrieval & Operationalization & Recall the concrete
operational criteria (metrics, red flags) an investor uses to apply a
principle & ``What metrics did Buffett use to validate a capital-light
business model?'' \\
L4 & Procedural Reconstruction & Framework Reconstruction & Reconstruct
an investor's multi-step decision process in the correct sequential
order & ``Reconstruct Buffett's corporate acquisition checklist,
detailing each gate and its kill criterion'' \\
L5 & Procedural Reconstruction & Cross-Investor Comparison & Synthesize
and contrast the procedural approaches of two or more distinct investors
& ``Contrast the macroeconomic positioning frameworks of Howard Marks
and Ray Dalio'' \\
L6 & Generative Application & Contradiction Recognition & Identify
documented contradictions, historical tensions, or temporal evolution
within an investor's views & ``Detail how Buffett's airline investments
contradicted his prior stated principles'' \\
L7 & Generative Application & Scenario Application & Apply a specific
multi-step framework to a dynamically generated novel investment
scenario & ``Apply the six-gate acquisition checklist to this SaaS
company at 10× ARR. State the verdict at each gate.'' \\
L8 & Generative Application & Novel Extrapolation & Extrapolate an
existing framework to a macroeconomic or technological development
absent from training data & ``How would Buffett, Dalio, and Soros each
analyze a programmable Central Bank Digital Currency?'' \\
\end{longtable}
}

Paralleling XFinBench's five-capability taxonomy {[}Zhang et al., ACL
2025{]}, each InvestPhilBench question is annotated with a
\textbf{capability profile} indicating which of five latent reasoning
capabilities it primarily exercises: (C1) \emph{Terminology precision}
(investor-specific vocabulary and operationalization recall), (C2)
\emph{Temporal reasoning} (historically-contextualized evolution of
investor views), (C3) \emph{Scenario extrapolation} (framework
application to novel scenarios), (C4) \emph{Comparative synthesis}
(cross-investor differentiation under surface similarity), (C5)
\emph{Procedural sequencing} (correct reconstruction and ordering of
multi-step decision processes). A single question may exercise multiple
capabilities; the primary capability is the one most predictive of
error.

\begin{center}\rule{0.5\linewidth}{0.5pt}\end{center}

\subsection{5. Dataset Architecture and Primary Source
Integration}\label{dataset-architecture-and-primary-source-integration}

\begin{quote}
\textbf{Canonical version note.} The canonical v0.6 release
(\texttt{release\_\allowbreak{}manifest\_\allowbreak{}v03.json}) comprises \textbf{118}
reviewer-resolved principle cards, \textbf{25} decision-framework cards
(with explicit \texttt{topology\_\allowbreak{}type} / \texttt{gate\_\allowbreak{}logic} /
\texttt{topology\_\allowbreak{}rationale} metadata), and \textbf{243} QA questions
(197 dev / 46 test; seed 20260420). The distribution tables in
§5.1--§5.4 (Tables 4, 6, 7) report the \textbf{v0.1 construction-time
pilot snapshot} that the methodology was first developed against; they
are retained for provenance and should not be read as the canonical
counts. All headline results and limitations (§8, §11) refer to the
canonical v0.6 data unless a table is explicitly labelled ``pilot.''
\end{quote}

\subsubsection{5.1 Principle Knowledge Base
(PKB)}\label{principle-knowledge-base-pkb}

The canonical PKB is the \textbf{118-card v0.6} Knowledge Base (Table
B.1); the per-investor breakdown in Table 4 is the \textbf{33-card v0.1
pilot} snapshot, retained for provenance only (see the §5 version note).
The PKB grew from that pilot to the 118-card v0.6 set; the description
below documents the schema, which is unchanged across versions. The v0.1
pilot contains \textbf{33} annotated principle cards across
\textbf{nine} individual practitioners, structured in
\texttt{pkb\_\allowbreak{}v01.json}. Each card carries an
\texttt{interpretation\_\allowbreak{}type} field distinguishing
\emph{investor-stated}, \emph{author-inferred}, and
\emph{community-consensus} operationalizations --- investor-stated
criteria require source attribution for full credit. All 33 principle
cards are \texttt{investor-\allowbreak{}stated} and corpus-linked to specific web
documents and/or YouTube transcripts from the scraped corpus (Table 7).
Each principle card includes an \texttt{operationalization} field
specifying concrete decision criteria, a \texttt{primary\_\allowbreak{}sources} list,
and a \texttt{contradictions} field documenting known cross-investor or
temporal conflicts --- the latter directly feeding L5 (comparison) and
L6 (contradiction) question construction.

\textbf{Table 4: PKB Principle Card Distribution (v0.1 pilot snapshot;
canonical v0.6 = 118 cards)}

{\def\LTcaptype{none} 
\begin{longtable}[]{@{}
  >{\raggedright\arraybackslash}p{(\linewidth - 10\tabcolsep) * \real{0.1099}}
  >{\raggedright\arraybackslash}p{(\linewidth - 10\tabcolsep) * \real{0.0879}}
  >{\raggedright\arraybackslash}p{(\linewidth - 10\tabcolsep) * \real{0.1209}}
  >{\raggedright\arraybackslash}p{(\linewidth - 10\tabcolsep) * \real{0.1758}}
  >{\raggedright\arraybackslash}p{(\linewidth - 10\tabcolsep) * \real{0.1758}}
  >{\raggedright\arraybackslash}p{(\linewidth - 10\tabcolsep) * \real{0.3297}}@{}}
\toprule\noalign{}
\begin{minipage}[b]{\linewidth}\raggedright
Investor
\end{minipage} & \begin{minipage}[b]{\linewidth}\raggedright
School
\end{minipage} & \begin{minipage}[b]{\linewidth}\raggedright
Principles
\end{minipage} & \begin{minipage}[b]{\linewidth}\raggedright
Framework Gates
\end{minipage} & \begin{minipage}[b]{\linewidth}\raggedright
Framework Type
\end{minipage} & \begin{minipage}[b]{\linewidth}\raggedright
Key Contradictions Annotated
\end{minipage} \\
\midrule\noalign{}
\endhead
\bottomrule\noalign{}
\endlastfoot
Warren Buffett & Quality Compounding & 8 & 6 & linear\_sequential &
documented airline reversal (2001 rejection → 2016 \$7.7B stake → 2020
exit); intrinsic-value margin of safety vs.~Graham's mechanical net-net
screen; reactive fear/greed signal vs.~Marks's active pendulum
monitoring \\
Benjamin Graham & Deep Value & 4 & 7 & parallel\_threshold & Buffett's
abandonment of Graham's net-net screen post-1965; Graham's
two-thirds-of-NAV operationalization differs from Buffett's
owner-earnings discount \\
Ray Dalio & Macro Systems & 4 & 5 & conditional\_branch & 2013
cash-as-transactional-medium framing vs.~2020--2022 defensive
cash-and-gold positioning (all pre-departure from Bridgewater, October
2022) \\
Howard Marks & Contrarian Credit & 3 & 4 & conditional\_branch & active
pendulum-temperature framework vs.~Buffett's reactive sentiment
signal \\
George Soros & Reflexive Macro & 3 & 4 & conditional\_branch & --- \\
Joel Greenblatt & Value-Quality Systematic & 4 & 6 & linear\_sequential
& --- \\
Seth Klarman & Deep Value / Risk Aversion & 2 & 4 & linear\_sequential &
--- \\
Charlie Munger & Multi-Disciplinary Mental Models & 2 & 4 &
parallel\_threshold & --- \\
Peter Lynch & Growth at Reasonable Price & 3 & 4 & parallel\_threshold &
--- \\
\textbf{Total} & --- & \textbf{33} & \textbf{44 total gates} & --- &
\textbf{8 cross-investor contradictions} \\
\end{longtable}
}

\emph{The pilot snapshot above grew to the 118-card canonical v0.6 KB
(Table B.1); planned v1.0 additions include \texttt{author-\allowbreak{}inferred} /
\texttt{community-\allowbreak{}consensus} variants and deeper coverage for
thin-corpus investors (Klarman, Munger). The PKB JSON schemas for both
versions are released with the repository.}

\subsubsection{5.2 Decision Framework Cards: Individual Practitioner
Frameworks}\label{decision-framework-cards-individual-practitioner-frameworks}

Framework cards encode the kinds of investment decisions that
practitioners actually make: whether to buy an operating business, value
a security, size a position, rebalance a portfolio, classify a macro
regime, sell a holding, or terminate analysis at a hard exclusion gate.
For example, Buffett's acquisition checklist governs an all-or-nothing
purchase decision, Dalio's bubble gauge governs regime-aware risk
reduction, and Soros's reflexive hypothesis framework governs trade
construction and position sizing under feedback. Each card represents
such a decision as structured JSON, capturing step name, gate
description, explicit criteria, kill criteria, source quotes, and
framework topology. The canonical v0.6 set contains \textbf{25}
individual-practitioner decision-framework cards, each annotated with
explicit \texttt{topology\_\allowbreak{}type}, \texttt{gate\_\allowbreak{}logic}, and
\texttt{topology\_\allowbreak{}rationale} metadata (§11.10). The three institutional
asset-owner frameworks of §5.3 are documented as prose specifications
and are \emph{not} among these cards (see the §5.3 data-status note).

\textbf{Charlie Munger --- Multi-Disciplinary Mental Models Framework}
(\texttt{munger\_\allowbreak{}opportunity\_\allowbreak{}evaluation}): Munger's framework is a
\texttt{parallel\_\allowbreak{}threshold} topology requiring convergence across
multiple independent analytical lenses before commitment. The framework
opens with a \emph{Multi-Lens Assessment} (Step 1) in which the
evaluator arrays the investment against economics, psychology,
mathematics, history, and evolutionary biology lenses --- Munger's
``latticework of mental models'' (USC Business School Speech, 1994). The
kill criterion at Step 1 is failure to find at least three independent
confirming lenses. Step 2 applies \emph{Inversion} --- systematically
enumerating technology-disruption, management-betrayal, regulatory,
competitive-moat erosion, and macro scenarios before considering upside,
reflecting his maxim ``Invert, always invert'' (USC Law Commencement,
2007). Step 3 performs \emph{Incentive Alignment} verification
(compensation tied to long-term value; owner-operator skin-in-the-game).
Step 4 enacts an \emph{Independence Check} against common cognitive
biases (confirmation bias, social proof, authority). Kill criterion: if
any identified bias is driving the majority of the bull case, the
position is rejected. The framework converges at a final
\emph{Expected-Value Synthesis} requiring dominant positive expected
value across all identified scenarios. The parallel-threshold topology
distinguishes Munger's framework from Buffett's sequential acquisition
checklist: a single lens failure can be overridden by overwhelming
convergence elsewhere, but cognitive-bias contamination at Step 4 is an
absolute veto.

The full inventory of all 25 canonical framework cards is generated
directly from \texttt{decision\_\allowbreak{}frameworks\_\allowbreak{}v05.json} in Table 5. ``Kill
gates'' counts the steps carrying explicit \texttt{kill\_\allowbreak{}criteria};
topology and gate-logic are the data fields audited in §11.10.

\textbf{Table 5: Decision Framework Card Inventory (v0.6, 25 cards;
generated from \texttt{decision\_\allowbreak{}frameworks\_\allowbreak{}v05.json})}

{\def\LTcaptype{none} 
\begin{longtable}[]{@{}
  >{\raggedright\arraybackslash}p{(\linewidth - 10\tabcolsep) * \real{0.1667}}
  >{\raggedright\arraybackslash}p{(\linewidth - 10\tabcolsep) * \real{0.1833}}
  >{\raggedright\arraybackslash}p{(\linewidth - 10\tabcolsep) * \real{0.1667}}
  >{\raggedright\arraybackslash}p{(\linewidth - 10\tabcolsep) * \real{0.1833}}
  >{\raggedright\arraybackslash}p{(\linewidth - 10\tabcolsep) * \real{0.1167}}
  >{\raggedright\arraybackslash}p{(\linewidth - 10\tabcolsep) * \real{0.1833}}@{}}
\toprule\noalign{}
\begin{minipage}[b]{\linewidth}\raggedright
Investor
\end{minipage} & \begin{minipage}[b]{\linewidth}\raggedright
Framework
\end{minipage} & \begin{minipage}[b]{\linewidth}\raggedright
Topology
\end{minipage} & \begin{minipage}[b]{\linewidth}\raggedright
Gate logic
\end{minipage} & \begin{minipage}[b]{\linewidth}\raggedright
Gates
\end{minipage} & \begin{minipage}[b]{\linewidth}\raggedright
Kill gates
\end{minipage} \\
\midrule\noalign{}
\endhead
\bottomrule\noalign{}
\endlastfoot
Warren Buffett & Buffett Acquisition Checklist &
\texttt{linear\_\allowbreak{}sequential} & conjunction & 6 & 6 \\
Warren Buffett & Buffett Owner Earnings DCF Valuation &
\texttt{linear\_\allowbreak{}sequential} & conjunction & 3 & 3 \\
Warren Buffett & Buffett Share Repurchase Decision Framework &
\texttt{linear\_\allowbreak{}sequential} & conjunction & 4 & 0 \\
Charlie Munger & Munger Business Quality and Price Checklist &
\texttt{parallel\_\allowbreak{}threshold} & conjunction & 5 & 1 \\
Charlie Munger & Munger Multi-Lens Opportunity Evaluation &
\texttt{parallel\_\allowbreak{}threshold} & weighted\_threshold\_3\_of\_5 & 5 & 0 \\
Benjamin Graham & Graham Defensive Investor Portfolio Selection &
\texttt{parallel\_\allowbreak{}threshold} & conjunction & 7 & 0 \\
Benjamin Graham & Graham Intrinsic Value and Margin of Safety &
\texttt{linear\_\allowbreak{}sequential} & conjunction & 4 & 0 \\
Benjamin Graham & Graham Net-Net Working Capital Screen &
\texttt{linear\_\allowbreak{}sequential} & conjunction & 2 & 2 \\
Benjamin Graham & Graham's Seven Tests for the Defensive Investor &
\texttt{parallel\_\allowbreak{}threshold} & conjunction & 7 & 7 \\
Seth Klarman & Klarman Deep Value Identification Framework &
\texttt{linear\_\allowbreak{}sequential} & conjunction & 4 & 2 \\
Seth Klarman & Klarman Downside-First Investment Process &
\texttt{linear\_\allowbreak{}sequential} & conjunction & 3 & 3 \\
Joel Greenblatt & Greenblatt Spinoff Investment Checklist &
\texttt{linear\_\allowbreak{}sequential} & conjunction & 4 & 0 \\
Joel Greenblatt & Magic Formula Stock-Screening Procedure &
\texttt{linear\_\allowbreak{}sequential} & conjunction & 5 & 3 \\
Peter Lynch & Lynch Portfolio Story Review Framework &
\texttt{linear\_\allowbreak{}sequential} & disjunction & 4 & 0 \\
Peter Lynch & Lynch Six-Category Investment Framework &
\texttt{conditional\_\allowbreak{}branch} & regime\_dependent & 4 & 0 \\
Howard Marks & Marks Distressed Debt Investment Framework &
\texttt{linear\_\allowbreak{}sequential} & conjunction & 5 & 0 \\
Howard Marks & Marks Market Cycle Positioning Framework &
\texttt{conditional\_\allowbreak{}branch} & regime\_dependent & 2 & 2 \\
Howard Marks & Marks Risk-Return Assessment Framework &
\texttt{parallel\_\allowbreak{}threshold} & weighted\_threshold\_3\_of\_5 & 5 & 0 \\
Ray Dalio & All-Weather Portfolio Construction &
\texttt{linear\_\allowbreak{}sequential} & conjunction & 3 & 1 \\
Ray Dalio & Dalio All Weather Rebalancing Framework &
\texttt{linear\_\allowbreak{}sequential} & conjunction & 3 & 0 \\
Ray Dalio & Dalio Bubble Detection Gauge & \texttt{parallel\_\allowbreak{}threshold}
& weighted\_threshold\_4\_of\_7 & 7 & 7 \\
Ray Dalio & Dalio Debt Supercycle Navigation Framework &
\texttt{conditional\_\allowbreak{}branch} & regime\_dependent & 2 & 2 \\
Ray Dalio & Dalio Economic Machine Portfolio Positioning &
\texttt{conditional\_\allowbreak{}branch} & regime\_dependent & 5 & 0 \\
George Soros & Soros Conviction-Weighted Position Sizing &
\texttt{linear\_\allowbreak{}sequential} & conjunction & 4 & 0 \\
George Soros & Soros Reflexive Hypothesis Trade Construction &
\texttt{linear\_\allowbreak{}sequential} & conjunction & 5 & 0 \\
\end{longtable}
}

\emph{Topology distribution: 15 \texttt{linear\_\allowbreak{}sequential}, 6
\texttt{parallel\_\allowbreak{}threshold}, 4 \texttt{conditional\_\allowbreak{}branch} (matching
§11.10). Gate logic spans conjunction, disjunction, regime-dependent,
and two weighted-threshold variants (3-of-5 and 4-of-7). Per-step
criteria, kill criteria, and source quotes for every card are in the
released JSON; the Buffett acquisition checklist is worked through in
full in §6.3 (Figure 5) and the Munger multi-lens framework above.}

\subsubsection{5.3 Decision Framework Cards: Institutional Asset Owner
Frameworks}\label{decision-framework-cards-institutional-asset-owner-frameworks}

Three institutional asset-owner frameworks are discussed throughout the
paper as motivating cases for the compound-kill-criterion metric (CKCA,
§7.10): \textbf{NBIM} (Norway's Government Pension Fund) operates a
mandatory ESG exclusion gate --- a Council-on-Ethics listing triggers
immediate divestment regardless of financial performance; \textbf{CPPIB}
applies a total-portfolio approach with a factor-substitutability kill
--- excess return attributable solely to systematic factor exposure
redirects capital to passive alternatives; \textbf{PIF} carries a dual
kill criterion --- an investment must satisfy IRR hurdles \emph{and}
align with Saudi Vision 2030 simultaneously. These are documented from
Tier-2 official sources (NBIM/CPPIB annual reports and
responsible-investment frameworks; the Government Pension Fund Act;
PIF's Santiago Principles self-assessment; see §6.1 and References).

\begin{quote}
\textbf{Data status (flagged for v1.0).} Unlike the 25
individual-practitioner frameworks (Table 5), the three institutional
frameworks are \textbf{not yet encoded as cards in
\texttt{decision\_\allowbreak{}frameworks\_\allowbreak{}v05.json}} --- they currently exist only
as prose specifications in this manuscript and as their underlying
Tier-2 PDFs. The CKCA metric (§7.10) and the institutional FM-3
kill-criterion-abstraction subtype (§9) are defined against these prose
specifications. Encoding NBIM/CPPIB/PIF as machine-readable framework
cards, and adding the corresponding QA items, is a v1.0 deliverable;
until then institutional results should be read as design illustrations
rather than measured benchmark coverage.
\end{quote}

\subsubsection{5.4 Pilot QA Dataset with Capability
Annotations}\label{pilot-qa-dataset-with-capability-annotations}

The expanded v0.1 pilot comprises \textbf{28 questions}: the original 20
plus \textbf{8 newly constructed Greenblatt questions} (one per layer,
L1--L8) that close the conspicuous coverage gap noted in v12. Layer
distribution: L1:4, L2:3, L3:4, L4:4, L5:4, L6:3, L7:3, L8:3.
Difficulty: Easy:4, Medium:10, Hard:10, Expert:4. Investor coverage:
Buffett and Greenblatt anchor the set with \textbf{8 questions each};
the remaining 12 are spread across Dalio, Soros, Lynch, Graham, Marks,
Munger, Klarman, and one legacy Fisher item (Fisher is excluded from the
canonical 9-investor set, see §11.7). Multi-investor comparison
questions (L5/L8) reference several investors but are counted once here,
under their primary investor. The Greenblatt questions include
\textbf{two Gold Reasoning Programs} (L4: Magic Formula reconstruction;
L7: scenario application to a mid-cap industrial company), adding a
second \texttt{linear\_\allowbreak{}sequential} framework to the GRP-annotated set
alongside Buffett. The full expanded QA set is released as
\texttt{investphilbench\_\allowbreak{}qa\_\allowbreak{}expanded.json}.

\textbf{Table 6: Question Capability Profile Distribution (v0.1 pilot)}

{\def\LTcaptype{none} 
\begin{longtable}[]{@{}lll@{}}
\toprule\noalign{}
Capability & Primary in n Questions & Predominant Layers \\
\midrule\noalign{}
\endhead
\bottomrule\noalign{}
\endlastfoot
C1: Terminology precision & 5 & L1, L2, L3 \\
C2: Temporal reasoning & 3 & L2, L6 \\
C3: Scenario extrapolation & 5 & L7, L8 \\
C4: Comparative synthesis & 4 & L5, L8 \\
C5: Procedural sequencing & 8 & L4, L5, L7 \\
\end{longtable}
}

\emph{The five rows sum to 25 of 28: the remaining three pilot questions
carry multi-capability profiles with no unique primary tag. Consistent
with XFinBench {[}Zhang et al., 2025{]}, temporal reasoning (C2) and
scenario extrapolation (C3) questions show the steepest performance drop
in the pilot (L6: 39--58\%, L7: 31--52\%), while terminology precision
(C1) questions achieve the highest scores (L1: 73--87\%).}

\textbf{v0.6 Expert-Difficulty L4 Hard Question Set.} Analysis of
per-gate GRA scores from the sanity-wave Claude Sonnet 4.6 evaluation
revealed four systematic failure patterns (Figure 2): (1) \emph{verdict
misclassification} --- gate names correctly retrieved but the response
maps the gate to the wrong normalized verdict, e.g.~a soft screen
treated as \textbf{PROCEED} when the gold program requires
\textbf{REJECT}, or an evidentiary gap treated as \textbf{REJECT} rather
than \textbf{INCONCLUSIVE}; legacy annotator shorthand such as FILTER,
KILL, FAIL, PASS, and RANK is mapped to the binary
rejection-vs-qualifying classes the GRA judge scores (rejection =
FILTER/KILL/FAIL ≡ §7.3 REJECT; qualifying = PASS/RANK ≡ §7.3 PROCEED)
before scoring; (2) \emph{element omission} --- gate named, normalized
verdict correct, but \texttt{required\_\allowbreak{}elements} partially cited; (3)
\emph{severity confusion} --- frameworks with multiple negative gate
outcomes are collapsed into a single generic REJECT even when the gold
program distinguishes a terminal kill from a lower-severity fail
condition through \texttt{required\_\allowbreak{}elements} and
\texttt{verdict\_\allowbreak{}trigger}; (4) \emph{sparse gate numbering} ---
hallucinated intermediate gates when a question intentionally targets
non-contiguous gates (e.g., {[}3,6{]} or {[}1,2,3,5{]}). To expose these
patterns in future evaluations, we added 8 expert-difficulty L4
questions (\texttt{IPB-\allowbreak{}L4H-\allowbreak{}GRA-\allowbreak{}001} through \texttt{IPB-\allowbreak{}L4H-\allowbreak{}GRA-\allowbreak{}008})
targeting: Graham gate activation with exact numeric thresholds (Q1),
Marks terminal-vs-nonterminal negative-verdict severity (Q2), Munger
all-negative-verdict recognition (Q3), Soros sparse gate numbering (Q4),
Greenblatt all-negative explanation (Q5), Klarman 3-element catalyst
gate completeness (Q6), Dalio 5-gate full reconstruction with specific
signals (Q7), and Buffett sparse Gates 3+6 rejection explanation (Q8).
All 8 questions have gold programs with ≥2 required\_elements per gate;
most existing L4 questions had only 1 required\_element per gate. These
8 questions are merged into \texttt{questions\_\allowbreak{}v04.json} (243 total),
with splits regenerated using the same seed (20260420): dev=197,
test=46. (The dev count moves from 188 to 197 --- not 188+8=196 ---
because the stratified split is recomputed over the full 243-item set
rather than appended to the prior dev split; the test split
correspondingly resolves to 46.)

\begin{figure}
\centering
\pandocbounded{\includegraphics[keepaspectratio,alt={Figure 2 --- GRA failure-pattern hardening map. The W1 gate-level analysis identified four recurring failure families: wrong normalized verdict, omitted required elements, negative-verdict severity confusion, and hallucinated gates under sparse numbering. The eight v0.6 L4H questions deliberately target these families rather than adding generic hard questions, so they should be read as adversarial hardening items first evaluated in v1.0, not as W1 evaluation data.}]{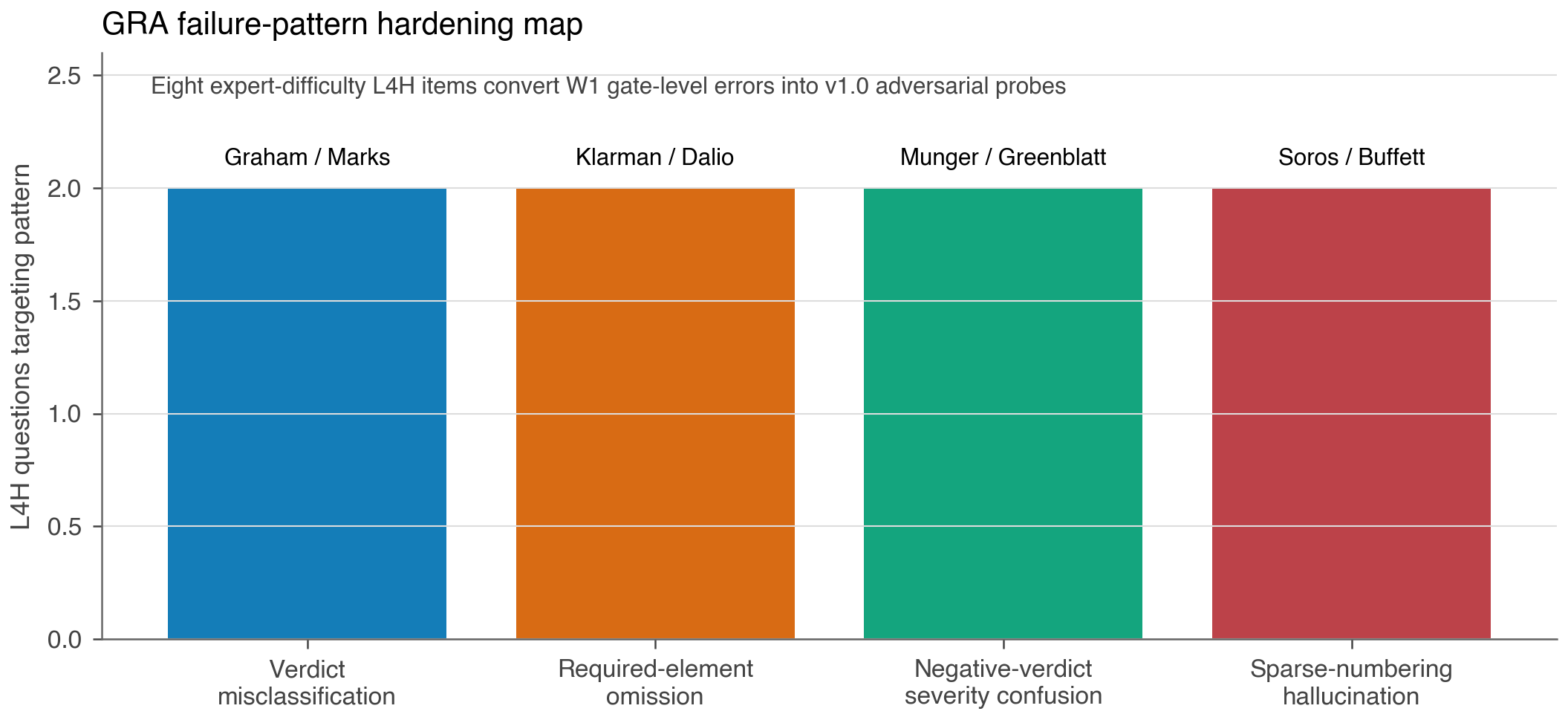}}
\caption{GRA failure-pattern hardening map. The W1
gate-level analysis identified four recurring failure families: wrong
normalized verdict, omitted required elements, negative-verdict severity
confusion, and hallucinated gates under sparse numbering. The eight v0.6
L4H questions deliberately target these families rather than adding
generic hard questions, so they should be read as adversarial hardening
items first evaluated in v1.0, not as W1 evaluation data.}
\end{figure}

\subsubsection{5.5 Pilot Question Selection
Protocol}\label{pilot-question-selection-protocol}

Questions selected via stratified sampling. In the expanded pilot (28
questions), Buffett and Greenblatt each have 8 questions (29\%),
reflecting Buffett's largest principle card count and Greenblatt's
status as the only investor with a fully scraped corpus (258,558
combined words --- Table 7) but zero original pilot coverage. The
Expert-difficulty L8 CBDC scenario topical correlation is acknowledged
as a pilot limitation --- v1.0 will enforce maximum 4 questions per
investor, mandatory topic diversity at Expert level, and will
\textbf{progress toward} balanced coverage across all \textbf{25}
individual-practitioner framework cards (Table 5).

\subsubsection{5.6 Conceptual Curation
Gate}\label{conceptual-curation-gate}

After the structural data audit, we added an explicit ontology
separating core investment principles/frameworks from adjacent mental
models, named portfolio strategies, management or life principles,
duplicates, and weakly grounded items. This gate was motivated by cases
such as Dalio's radical transparency and believability-weighted
decision-making, which are important Bridgewater operating principles
but not direct risk-parity or asset-allocation principles. Similarly,
All Weather is treated as a portfolio strategy and the Economic Machine
as a macro explanatory model unless an item documents explicit
investment action gates. The curated core keeps only items that directly
govern security selection, valuation, risk, portfolio construction,
market cycles, position sizing, capital allocation, or an
evidence-to-action investment procedure. Adjacent items remain available
in curation appendices but are not counted as core benchmark knowledge.

\begin{center}\rule{0.5\linewidth}{0.5pt}\end{center}

\subsection{6. Data Construction
Methodology}\label{data-construction-methodology}

\begin{figure}
\centering
\pandocbounded{\includegraphics[keepaspectratio,alt={Figure 3 --- Benchmark construction-and-evaluation pipeline (schematic). Five stages: primary sources (9 investors, 1.67 M words) → knowledge base (118 principle cards + 25 frameworks with topology metadata) → QA benchmark (243 questions, L1--L8, with gold reasoning programs for L4/L7) → evaluation (4 W1 models, scaling to 10; three conditions; BASP + GRA) → diagnosis (provider-tier split; composite saturation vs.~the gate-level GRA gap; FMDP × 6 failure modes).}]{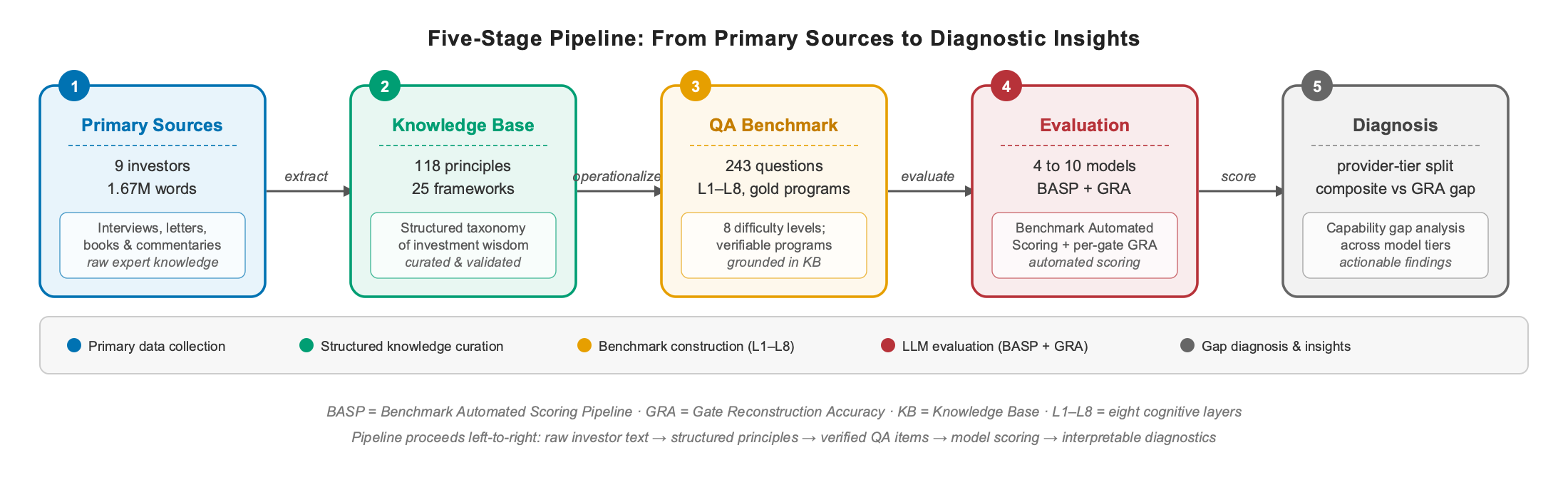}}
\caption{Benchmark construction-and-evaluation pipeline
(schematic). Five stages: \textbf{primary sources} (9 investors, 1.67 M
words) → \textbf{knowledge base} (118 principle cards + 25 frameworks
with topology metadata) → \textbf{QA benchmark} (243 questions, L1--L8,
with gold reasoning programs for L4/L7) → \textbf{evaluation} (4 W1
models, scaling to 10; three conditions; BASP + GRA) →
\textbf{diagnosis} (provider-tier split; composite saturation vs.~the
gate-level GRA gap; FMDP × 6 failure modes).}
\end{figure}

\subsubsection{6.1 Source Selection
Hierarchy}\label{source-selection-hierarchy}

Principle cards and framework citations follow a \textbf{six-tier}
priority order (higher tiers override lower tiers when
operationalizations conflict):

\begin{enumerate}
\def\labelenumi{\arabic{enumi}.}
\tightlist
\item
  \textbf{Tier 1 --- Investor-authored books and official shareholder
  letters} (e.g., canonical texts; Berkshire Hathaway annual letters in
  official PDF form).
\item
  \textbf{Tier 2 --- Official institutional mandates and annual reports}
  (sovereign wealth and public pension frameworks: NBIM, CPPIB, PIF).
\item
  \textbf{Tier 3 --- Official digital essays and research publications}
  (e.g., Bridgewater Research \& Insights, Oaktree Capital memos, Soros
  essays, Berkshire.com letters).
\item
  \textbf{Tier 4 --- Primary-source video interviews.} Long-form
  interviews in which the investor speaks in the first person;
  provenance is fixed via YouTube Data API metadata (\texttt{video\_\allowbreak{}id},
  channel, \texttt{published\_\allowbreak{}at}, \texttt{view\_\allowbreak{}count}) and English
  transcripts (\texttt{transcript\_\allowbreak{}language},
  \texttt{transcript\_\allowbreak{}word\_\allowbreak{}count}). Clips that are purely third-party
  commentary without the investor's voice are excluded. The v0.1 corpus
  includes \textbf{258} such transcripts across \textbf{nine} individual
  practitioners (\textasciitilde1.28M spoken words). These sources are
  essential for \textbf{temporal} and \textbf{contradiction} tasks (L6)
  where spoken remarks span multiple dates and may diverge from
  written-only narratives.
\item
  \textbf{Tier 5 --- Curated digital archives} that reproduce or
  faithfully excerpt primary text (e.g., FarnamStreet, Acquirer's
  Multiple), used when Tier 1--3 URLs are unavailable or when the
  archive provides a stable citation.
\item
  \textbf{Tier 6 --- Secondary synthesis} (aggregators, summaries);
  permitted only under
  \texttt{interpretation\_\allowbreak{}type:\ community-\allowbreak{}consensus} and cannot satisfy
  investor-stated attribution requirements alone.
\end{enumerate}

\textbf{Table 7: PKB Primary Source Corpus (v0.1) --- Written Documents
and YouTube Transcripts}

{\def\LTcaptype{none} 
\begin{landscape}
\setlength{\hsize}{648pt}%
\setlength{\textwidth}{\hsize}%
\setlength{\columnwidth}{\hsize}%
\setlength{\linewidth}{\hsize}%
\begin{longtable}[]{@{}
  >{\raggedright\arraybackslash}p{(\linewidth - 10\tabcolsep) * \real{0.0935}}
  >{\raggedright\arraybackslash}p{(\linewidth - 10\tabcolsep) * \real{0.1215}}
  >{\raggedright\arraybackslash}p{(\linewidth - 10\tabcolsep) * \real{0.1402}}
  >{\raggedright\arraybackslash}p{(\linewidth - 10\tabcolsep) * \real{0.3458}}
  >{\raggedright\arraybackslash}p{(\linewidth - 10\tabcolsep) * \real{0.1682}}
  >{\raggedright\arraybackslash}p{(\linewidth - 10\tabcolsep) * \real{0.1308}}@{}}
\toprule\noalign{}
\begin{minipage}[b]{\linewidth}\raggedright
Investor
\end{minipage} & \begin{minipage}[b]{\linewidth}\raggedright
Written docs
\end{minipage} & \begin{minipage}[b]{\linewidth}\raggedright
Written words
\end{minipage} & \begin{minipage}[b]{\linewidth}\raggedright
Primary written sources (examples)
\end{minipage} & \begin{minipage}[b]{\linewidth}\raggedright
YT transcripts
\end{minipage} & \begin{minipage}[b]{\linewidth}\raggedright
Spoken words
\end{minipage} \\
\midrule\noalign{}
\endhead
\bottomrule\noalign{}
\endlastfoot
Warren Buffett & 21 & 174,993 & Berkshire Hathaway Annual Letters
(2004--2024) & 30 & 97,797 \\
Charlie Munger & 16 & 26,736 & FarnamStreet (Munger essays/speeches) &
18 & 53,947 \\
Ray Dalio & 30 & 79,654 & Bridgewater Research \& Insights & 30 &
143,847 \\
Howard Marks & 12 & 42,510 & Oaktree Capital --- Howard Marks memos & 30
& 201,779 \\
George Soros & 30 & 35,971 & George Soros essays & 30 & 130,578 \\
Seth Klarman & 22 & 9,570 & Acquirer's Multiple; FarnamStreet & 30 &
174,119 \\
Benjamin Graham & 3 & 3,188 & Benjamin Graham Archive (FarnamStreet) &
30 & 111,446 \\
Peter Lynch & 20 & 11,449 & Acquirer's Multiple & 30 & 114,619 \\
Joel Greenblatt & 23 & 10,539 & Acquirer's Multiple; FarnamStreet & 30 &
248,019 \\
\textbf{Total} & \textbf{177} & \textbf{394,610} & --- & \textbf{258} &
\textbf{\textasciitilde1,276,151} \\
\end{longtable}
\end{landscape}
}

\emph{Combined written + spoken word counts are used in §11 (coverage
asymmetry). Institutional frameworks (NBIM, CPPIB, PIF) are documented
from Tier-2 PDFs and reports and are not included in this table. All
Table 7 figures were programmatically verified against the released
corpus artifacts (Phase 1 audit; zero discrepancies).}

\begin{figure}
\centering
\pandocbounded{\includegraphics[keepaspectratio,alt={Figure 4 --- Written vs.~spoken corpus asymmetry by investor. The bar chart stacks written-document words and primary YouTube-transcript words as percentages of each investor's combined corpus. Buffett is the only written-dominant investor (64.1\% written), Dalio/Munger/Soros are mixed, and the remaining five investors are Tier-4 dominant (\textgreater80\% spoken). This asymmetry drives source-design choices: Buffett supports document-level L2 provenance questions, while Graham/Greenblatt/Klarman/Lynch/Marks require more careful treatment of interview-derived evidence.}]{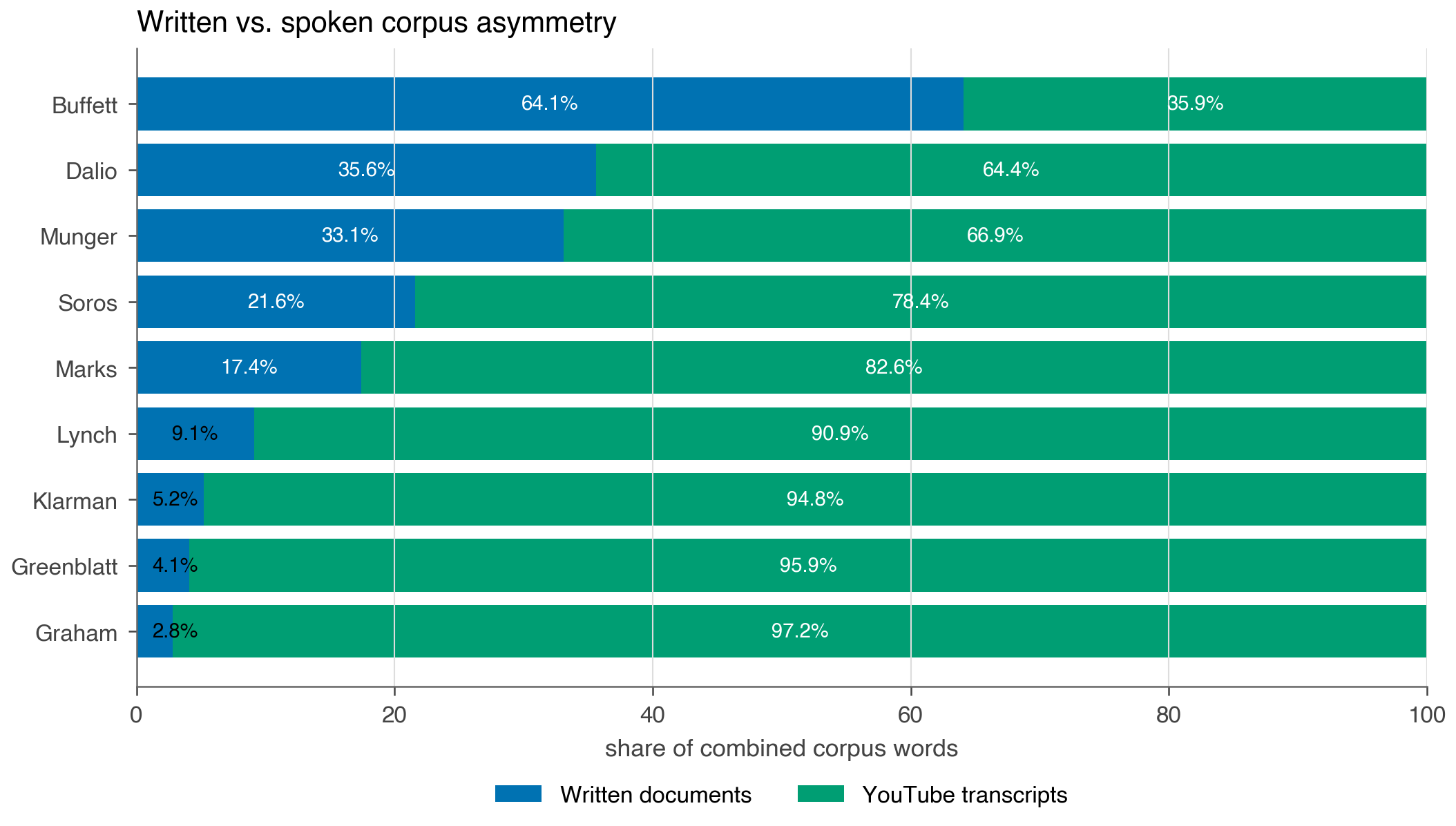}}
\caption{Written vs.~spoken corpus asymmetry by investor.
The bar chart stacks written-document words and primary
YouTube-transcript words as percentages of each investor's combined
corpus. Buffett is the only written-dominant investor (64.1\% written),
Dalio/Munger/Soros are mixed, and the remaining five investors are
Tier-4 dominant (\textgreater80\% spoken). This asymmetry drives
source-design choices: Buffett supports document-level L2 provenance
questions, while Graham/Greenblatt/Klarman/Lynch/Marks require more
careful treatment of interview-derived evidence.}
\end{figure}

\emph{The written/spoken asymmetry has direct implications for benchmark
task design. \textbf{Buffett} is the only investor whose corpus is
written-dominant (64.1\%), making him uniquely suited for L2 (source
attribution) questions that test document-level provenance.
\textbf{Graham} and \textbf{Greenblatt} are almost entirely
spoken-corpus-dependent (\textgreater95\%), meaning L6 (contradiction)
and L2 (source attribution) tasks for these investors must account for
the provenance limitations of YouTube-sourced content.}

\textbf{Table 8: Distinctive Investor Phrases Used for Voice-Diffusion
Diagnostics}

{\def\LTcaptype{none} 
\begin{longtable}[]{@{}
  >{\raggedright\arraybackslash}p{(\linewidth - 4\tabcolsep) * \real{0.1724}}
  >{\raggedright\arraybackslash}p{(\linewidth - 4\tabcolsep) * \real{0.5517}}
  >{\raggedright\arraybackslash}p{(\linewidth - 4\tabcolsep) * \real{0.2759}}@{}}
\toprule\noalign{}
\begin{minipage}[b]{\linewidth}\raggedright
Investor
\end{minipage} & \begin{minipage}[b]{\linewidth}\raggedright
Distinctive phrases / concepts
\end{minipage} & \begin{minipage}[b]{\linewidth}\raggedright
Diagnostic use
\end{minipage} \\
\midrule\noalign{}
\endhead
\bottomrule\noalign{}
\endlastfoot
Warren Buffett & circle of competence; owner earnings; economic moat;
margin of safety; be fearful when others are greedy & Distinguishes
quality-compounding and intrinsic-value language from generic value
investing \\
Charlie Munger & latticework of mental models; invert, always invert;
incentives; elementary worldly wisdom; sit on your ass investing & Tests
whether multi-disciplinary reasoning is preserved rather than collapsed
into Buffett-style quality language \\
Benjamin Graham & net-net; two-thirds of net current asset value;
Mr.~Market; defensive investor; margin of safety & Separates mechanical
asset-value screening from Buffett's owner-earnings valuation \\
Ray Dalio & economic machine; debt cycle; all weather; risk parity; cash
is trash & Marks macro-system and regime language distinct from Soros
reflexivity \\
Howard Marks & pendulum; second-level thinking; risk control; market
cycle temperature; most important thing & Identifies active
cycle-temperature monitoring rather than generic contrarianism \\
George Soros & reflexivity; boom-bust sequence; fallibility; feedback
loop; asymmetric bet sizing & Captures hypothesis-testing and feedback
dynamics absent from Dalio's systematic macro language \\
Seth Klarman & absolute return; downside protection; catalyst; mispriced
liquidation value; patience & Separates risk-first deep value from
Graham's formulaic screens and Buffett's quality compounding \\
Peter Lynch & tenbagger; stalwart; fast grower; story stock; buy what
you know & Preserves retail-growth taxonomy and narrative company-story
evaluation \\
Joel Greenblatt & magic formula; earnings yield; return on capital;
spinoff; special situation & Separates systematic value-quality ranking
from discretionary value-investor prose \\
\end{longtable}
}

\emph{Table 8 clarifies the purpose of the corpus lexical profile: the
benchmark does not use raw frequency words such as ``market'' and
``price'' as substantive findings. Instead, investor-specific phrase
sets define a face-valid reference for FM-4 (Voice Diffusion) and
IVP/SDI diagnostics: a response that replaces Soros's ``reflexivity''
with generic macro-cycle language, or Graham's net-net screen with
Buffett-style owner earnings, loses voice fidelity even when the
high-level investment theme remains plausible.}

\subsubsection{6.2 Five-Phase Ground Truth
Construction}\label{five-phase-ground-truth-construction}

\textbf{Phase 0 --- Automated corpus construction and linkage.} Written
documents and YouTube interview transcripts are collected into
per-investor JSON artifacts. \textbf{Web:} scrape or fetch
investor-authored PDFs and HTML from official domains (e.g.,
\texttt{berkshirehathaway.com} letters, \texttt{bridgewater.com},
\texttt{oaktreecapital.com}) and from curated archives (FarnamStreet,
Acquirer's Multiple) with \texttt{title}, \texttt{url}, \texttt{date},
\texttt{source\_\allowbreak{}name}, \texttt{text}, and \texttt{word\_\allowbreak{}count}.
\textbf{YouTube:} YouTube Data API v3 search → video metadata →
transcript extraction; a transcript is retained when
\texttt{transcript\_\allowbreak{}language\ =\ "en"} and
\texttt{transcript\_\allowbreak{}word\_\allowbreak{}count\ \textgreater{}\ 200}. Each record
includes provenance fields (\texttt{scraped\_\allowbreak{}at} or fetch timestamp) to
support replication and contamination analysis. Aggregated counts appear
in Table 7. \textbf{Corpus linkage:} An automated enrichment pipeline
then cross-references each PKB principle card, decision framework card,
and QA question against the full scraped corpus via keyword overlap and
quote matching. The resulting \texttt{corpus\_\allowbreak{}references}
(principles/frameworks) and \texttt{corpus\_\allowbreak{}evidence} (QA) fields
provide direct provenance links --- specific document titles, URLs,
video IDs, and relevance scores --- enabling traceability from any
benchmark element to its primary-source evidence in the released corpus.
A comprehensive \texttt{corpus\_\allowbreak{}index.json} (9 investors, 177 web
documents, 258 YouTube transcripts, 1.67M combined words) serves as the
master registry for all scraped artifacts.

\textbf{Phase 0a --- Data quality audit.} All Table 7 figures were
programmatically verified against the released corpus JSON artifacts
(\texttt{\_\allowbreak{}index.json} manifests and per-investor files). The audit
confirmed zero discrepancies across all 9 investors for document counts,
word counts, and transcript counts. Extended statistics (Figure 4:
written/spoken asymmetry; Table 8: distinctive phrase profile) were
computed from the full corpus to inform benchmark design decisions and
FM-4 (Voice Diffusion) detection baselines.

\textbf{Phase 1 --- Principle extraction.} Candidate principles are
extracted per investor from the Tier 1/3/4 corpus (LLM-assisted,
author-reviewed), each anchored to a canonical quote and a dated source
document. \textbf{Phase 2 --- Cross-verification.} Each candidate is
verified against at least one independent corpus source; items that
cannot be independently confirmed are demoted to
\texttt{author-\allowbreak{}inferred} or dropped. \textbf{Phase 3 ---
Operationalization tagging.} Verified principles are annotated with
concrete metrics, red flags, and kill criteria, with
\texttt{interpretation\_\allowbreak{}type} distinguishing investor-stated from
author-inferred operationalizations (§5.1). \textbf{Phase 4 --- QA
generation and perturbation design.} Questions are generated per layer
with evaluation rubrics and \texttt{corpus\_\allowbreak{}evidence} links, subject to
the career-boundary filter of §11.6; L7/L8 items additionally receive
the dynamic-perturbation treatment of §6.5.

\subsubsection{6.3 Gold Reasoning Program
Annotation}\label{gold-reasoning-program-annotation}

Inspired by FinQA's gold reasoning program design {[}Chen et al.,
2021{]} --- which annotates the complete arithmetic operation sequence
for each numerical question, enabling dual evaluation of final answer
correctness (execution accuracy) and reasoning path correctness (program
accuracy) --- InvestPhilBench attaches \textbf{gold reasoning programs
(GRPs)} wherever the released QA item has an explicit procedural target.
In the canonical v0.6 file this is broader than the early design
shorthand of ``L4/L7 only'': 154/243 questions carry a
\texttt{gold\_\allowbreak{}program} object across all eight layers, but most
lower-layer programs contain only one or two atomic checks. The
sequential-procedure claims in this paper therefore rest on the
multi-gate subset, especially the L4/L7 framework questions and the
≥3-gate hardening items, while single-gate GRPs are used mainly for
consistent scoring and provenance.

A GRP is a structured object specifying:

\begin{Shaded}
\begin{Highlighting}[]
\FunctionTok{\{}
  \DataTypeTok{"question\_id"}\FunctionTok{:} \StringTok{"IPB{-}P{-}011"}\FunctionTok{,}
  \DataTypeTok{"layer"}\FunctionTok{:} \StringTok{"L7"}\FunctionTok{,}
  \DataTypeTok{"framework"}\FunctionTok{:} \StringTok{"buffett\_acquisition\_checklist"}\FunctionTok{,}
  \DataTypeTok{"gold\_program"}\FunctionTok{:} \OtherTok{[}
    \FunctionTok{\{}\DataTypeTok{"gate"}\FunctionTok{:} \DecValTok{1}\FunctionTok{,} \DataTypeTok{"name"}\FunctionTok{:} \StringTok{"Circle of Competence"}\FunctionTok{,}
     \DataTypeTok{"verdict"}\FunctionTok{:} \StringTok{"PROCEED"}\FunctionTok{,}
     \DataTypeTok{"kill\_criterion"}\FunctionTok{:} \KeywordTok{null}\FunctionTok{,}
     \DataTypeTok{"required\_elements"}\FunctionTok{:} \OtherTok{[}\StringTok{"subscription economics"}\OtherTok{,} \StringTok{"retention metric"}\OtherTok{,}
                              \StringTok{"comprehensible"}\OtherTok{]}\FunctionTok{\}}\OtherTok{,}
    \FunctionTok{\{}\DataTypeTok{"gate"}\FunctionTok{:} \DecValTok{2}\FunctionTok{,} \DataTypeTok{"name"}\FunctionTok{:} \StringTok{"Economic Moat"}\FunctionTok{,}
     \DataTypeTok{"verdict"}\FunctionTok{:} \StringTok{"PROCEED"}\FunctionTok{,}
     \DataTypeTok{"kill\_criterion"}\FunctionTok{:} \KeywordTok{null}\FunctionTok{,}
     \DataTypeTok{"required\_elements"}\FunctionTok{:} \OtherTok{[}\StringTok{"switching costs"}\OtherTok{,} \StringTok{"pricing power"}\OtherTok{,} \StringTok{"retention \textgreater{}= 90\%"}\OtherTok{]}\FunctionTok{\}}\OtherTok{,}
    \FunctionTok{\{}\DataTypeTok{"gate"}\FunctionTok{:} \DecValTok{3}\FunctionTok{,} \DataTypeTok{"name"}\FunctionTok{:} \StringTok{"Management Integrity"}\FunctionTok{,}
     \DataTypeTok{"verdict"}\FunctionTok{:} \StringTok{"INCONCLUSIVE"}\FunctionTok{,}
     \DataTypeTok{"kill\_criterion"}\FunctionTok{:} \KeywordTok{null}\FunctionTok{,}
     \DataTypeTok{"required\_elements"}\FunctionTok{:} \OtherTok{[}\StringTok{"data\_insufficiency\_acknowledged"}\OtherTok{]}\FunctionTok{\}}\OtherTok{,}
    \FunctionTok{\{}\DataTypeTok{"gate"}\FunctionTok{:} \DecValTok{4}\FunctionTok{,} \DataTypeTok{"name"}\FunctionTok{:} \StringTok{"Intrinsic Value Estimation"}\FunctionTok{,}
     \DataTypeTok{"verdict"}\FunctionTok{:} \StringTok{"PROCEED"}\FunctionTok{,}
     \DataTypeTok{"kill\_criterion"}\FunctionTok{:} \KeywordTok{null}\FunctionTok{,}
     \DataTypeTok{"required\_elements"}\FunctionTok{:} \OtherTok{[}\StringTok{"owner\_earnings\_or\_FCF"}\OtherTok{,} \StringTok{"maturity\_margin"}\OtherTok{,} \StringTok{"discount\_rate"}\OtherTok{]}\FunctionTok{,}
     \DataTypeTok{"computed\_value"}\FunctionTok{:} \StringTok{"\textasciitilde{}$25B"}\FunctionTok{\}}\OtherTok{,}
    \FunctionTok{\{}\DataTypeTok{"gate"}\FunctionTok{:} \DecValTok{5}\FunctionTok{,} \DataTypeTok{"name"}\FunctionTok{:} \StringTok{"Margin of Safety"}\FunctionTok{,}
     \DataTypeTok{"verdict"}\FunctionTok{:} \StringTok{"REJECT"}\FunctionTok{,}
     \DataTypeTok{"kill\_criterion"}\FunctionTok{:} \StringTok{"price\_exceeds\_intrinsic\_value"}\FunctionTok{,}
     \DataTypeTok{"required\_elements"}\FunctionTok{:} \OtherTok{[}\StringTok{"price $50B"}\OtherTok{,} \StringTok{"intrinsic \textasciitilde{}$25B"}\OtherTok{,} \StringTok{"2x premium"}\OtherTok{]}\FunctionTok{,}
     \DataTypeTok{"kill\_triggered"}\FunctionTok{:} \KeywordTok{true}\FunctionTok{\}}\OtherTok{,}
    \FunctionTok{\{}\DataTypeTok{"gate"}\FunctionTok{:} \DecValTok{6}\FunctionTok{,} \DataTypeTok{"name"}\FunctionTok{:} \StringTok{"Holding Period Economics"}\FunctionTok{,}
     \DataTypeTok{"verdict"}\FunctionTok{:} \StringTok{"N/A"}\FunctionTok{,}
     \DataTypeTok{"kill\_criterion"}\FunctionTok{:} \StringTok{"terminated\_by\_gate\_5"}\FunctionTok{,}
     \DataTypeTok{"required\_elements"}\FunctionTok{:} \OtherTok{[]}\FunctionTok{\}}
  \OtherTok{]}\FunctionTok{,}
  \DataTypeTok{"investment\_verdict"}\FunctionTok{:} \StringTok{"DECLINE"}\FunctionTok{,}
  \DataTypeTok{"verdict\_trigger"}\FunctionTok{:} \StringTok{"gate\_5\_reject"}
\FunctionTok{\}}
\end{Highlighting}
\end{Shaded}

\textbf{Schema note.} The JSON above is the \emph{target} GRP schema,
shown fully populated for clarity. In the v0.6 release the gold-program
gates use legacy verdict labels (\texttt{PASS} / \texttt{FAIL} /
\texttt{FILTER} / \texttt{KILL} / \texttt{RANK}), carry
\texttt{required\_\allowbreak{}elements} but no separate \texttt{kill\_\allowbreak{}criterion}
field, and do not populate a top-level \texttt{investment\_\allowbreak{}verdict}. The
GRA judge collapses each gate's legacy verdict to a \emph{rejection}
(FILTER / KILL / FAIL ≡ §7.3 REJECT) vs \emph{qualifying} (PASS / RANK ≡
§7.3 PROCEED) class --- a binary match with no separate INCONCLUSIVE
class; kill semantics are read from the rejection verdict and
\texttt{verdict\_\allowbreak{}trigger} (§7.7), and the final decision target is
grounded as described in the next paragraph; explicit
\texttt{kill\_\allowbreak{}criterion} and \texttt{investment\_\allowbreak{}verdict} annotation is
a v1.0 schema upgrade. The fully-populated form is therefore the schema
the metrics are \emph{defined} against, not a verbatim copy of every
released record.

The GRP enables dual evaluation analogous to FinQA: \textbf{verdict
accuracy} (is the final INVEST/DECLINE or DECLINE-equivalent action
correct when the item defines one?) and \textbf{program accuracy} (is
the gate sequence, with correct verdicts and kill semantics,
reconstructed?). In the current release, \texttt{investment\_\allowbreak{}verdict} is
not populated uniformly across all GRPs. In the W1 cut, L7 verdict
correctness is captured by the OGRS judge rubric (§7.6) rather than a
separate deterministic VerdictAcc. The planned VerdictAcc scorer ---
grounding the final target in \texttt{investment\_\allowbreak{}verdict} when present,
otherwise deriving it from the terminal gate verdict and
\texttt{verdict\_\allowbreak{}trigger} / ground-truth answer fields, and excluding
items with no recoverable decision target rather than assigning a
placeholder --- is a v1.0 deliverable. OGRS and KCCS together
operationalize program accuracy.

\emph{GRP annotation agreement statistics (v0.1 pilot, two annotators)
are reported in §6.4 and Table E.2; the full v0.6 GRP set is
single-annotated (§11.8).}

\begin{figure}
\centering
\pandocbounded{\includegraphics[keepaspectratio,alt={Figure 5 --- Gold Reasoning Program (GRP) and gate-level scoring (schematic). Buffett's 6-gate acquisition checklist with per-gate verdicts (PROCEED / INCONCLUSIVE / REJECT): Gate 5 (Margin of Safety) triggers the kill criterion, terminating the chain at the overall verdict DECLINE and leaving Gate 6 unreached. Gate Reconstruction Accuracy (GRA) scores each gate on a name + verdict + required-elements match (§7.15); per-model GRA results are in §8.7. The failure modes most directly exposed by gate-level scoring are FM-1 (Hallucinated Gate), FM-3 (Kill Criterion Omission), FM-5 (Operationalization Flattening), and FM-6 (Anachronistic Application) (§9). A fully worked numeric instance (SaaS target, \$50 B price vs.~≈\$25 B intrinsic value) is given in §6.3.}]{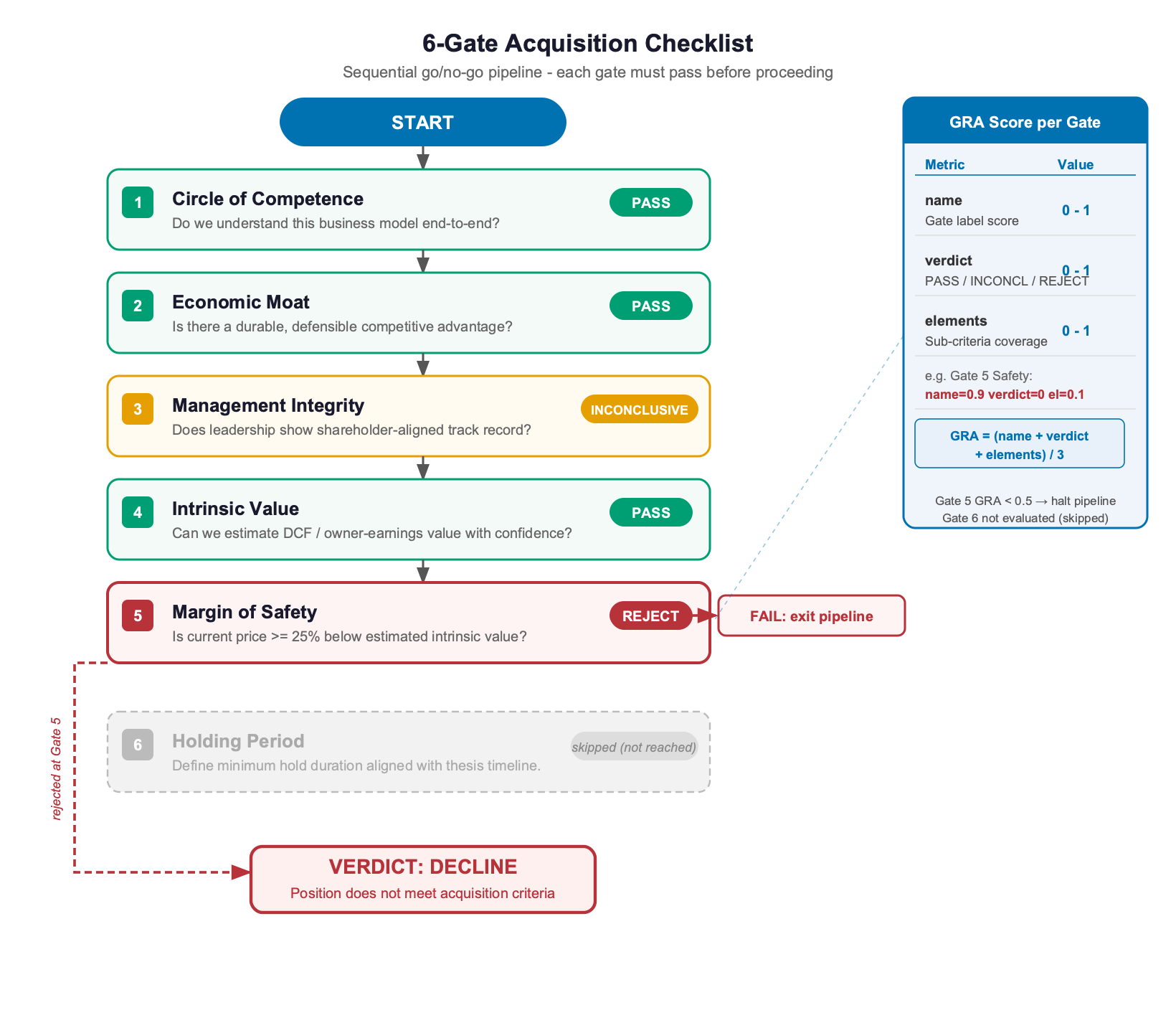}}
\caption{Gold Reasoning Program (GRP) and gate-level
scoring (schematic). Buffett's 6-gate acquisition checklist with
per-gate verdicts (PROCEED / INCONCLUSIVE / REJECT): Gate 5 (Margin of
Safety) triggers the kill criterion, terminating the chain at the
overall verdict \textbf{DECLINE} and leaving Gate 6 unreached. Gate
Reconstruction Accuracy (GRA) scores each gate on a \emph{name + verdict
+ required-elements} match (§7.15); per-model GRA results are in §8.7.
The failure modes most directly exposed by gate-level scoring are FM-1
(Hallucinated Gate), FM-3 (Kill Criterion Omission), FM-5
(Operationalization Flattening), and FM-6 (Anachronistic Application)
(§9). A fully worked numeric instance (SaaS target, \$50 B price
vs.~≈\$25 B intrinsic value) is given in §6.3.}
\end{figure}

\subsubsection{6.4 Inter-Annotator
Agreement}\label{inter-annotator-agreement}

All inter-annotator estimates in this section are from the \textbf{v0.1
20-question pilot} (two annotators): Cohen's κ = 0.82 (overall
three-tier rubric); κ = 0.76 (SAA context-level sub-metric); κ = 0.86
(GRP gate verdicts); κ = 0.78 (GRP kill criterion identification). They
carry wide CIs at n=20 and do not extend to the full v0.6 set, whose
GRPs are single-annotated (§11.8); v1.0 will provide reliable estimates.

\subsubsection{6.5 Dynamic Perturbation
Design}\label{dynamic-perturbation-design}

The contamination-resistance argument of §2.6 is operationalized for L7
(Scenario Application) and L8 (Novel Extrapolation) through three
perturbation operators applied to a base scenario while holding the
underlying framework fixed: (i) \textbf{metric substitution} ---
replacing the numeric inputs a gate consumes (valuation multiple, ARR,
growth rate) so the verdict cannot be retrieved from a memorized
instance; (ii) \textbf{scenario / sector substitution} --- relocating
the same decision structure to a different industry; and (iii)
\textbf{temporal displacement} --- injecting a dated macro regime
(e.g.~a 2009 credit-crisis context) that activates regime-dependent
gates. Because each operator preserves the gold reasoning program (gate
set, kill criteria, ordering) while changing the \emph{answer}, a model
must \emph{apply} the framework rather than recall a cached label ---
exactly the dynamic slot-substitution principle catalogued in the
contamination-resistance literature (Chen et al., 2025; §2.6).

The design is illustrated on the worked Buffett L7 instance of §6.3
(\texttt{IPB-\allowbreak{}P-\allowbreak{}011}, six-gate acquisition checklist). The base scenario
triggers a Gate-5 REJECT (price ≫ intrinsic value); the three
perturbations move the verdict in structurally meaningful ways while
reusing the identical gold program:

\textbf{Table 9: Dynamic Perturbation Example for L7 Question
\texttt{IPB-\allowbreak{}P-\allowbreak{}011} (Buffett six-gate acquisition checklist)}

{\def\LTcaptype{none} 
\begin{longtable}[]{@{}
  >{\raggedright\arraybackslash}p{(\linewidth - 6\tabcolsep) * \real{0.1636}}
  >{\raggedright\arraybackslash}p{(\linewidth - 6\tabcolsep) * \real{0.1818}}
  >{\raggedright\arraybackslash}p{(\linewidth - 6\tabcolsep) * \real{0.3091}}
  >{\raggedright\arraybackslash}p{(\linewidth - 6\tabcolsep) * \real{0.3455}}@{}}
\toprule\noalign{}
\begin{minipage}[b]{\linewidth}\raggedright
Version
\end{minipage} & \begin{minipage}[b]{\linewidth}\raggedright
Scenario
\end{minipage} & \begin{minipage}[b]{\linewidth}\raggedright
Changed variable
\end{minipage} & \begin{minipage}[b]{\linewidth}\raggedright
Investment verdict
\end{minipage} \\
\midrule\noalign{}
\endhead
\bottomrule\noalign{}
\endlastfoot
Base & SaaS, 10× ARR, \$5B ARR, 25\% growth & --- & DECLINE (Gate 5
REJECT) \\
Perturbation A & Industrial-automation SaaS, 7× ARR, \$2B ARR, 20\%
growth & Sector + multiple & DECLINE (Gate 5 REJECT) \\
Perturbation B & SaaS, 3× ARR, \$5B ARR, 15\% growth & Multiple (price
\textasciitilde\$15B \textless{} intrinsic \textasciitilde\$20B) &
INVEST (all gates PROCEED / INCONCLUSIVE) \\
Perturbation C & SaaS, 10× ARR, \$5B ARR, 25\% growth, 2009 context &
Temporal regime & DECLINE (Gate 5 + Gate 6 REJECT) \\
\end{longtable}
}

\emph{Perturbation B (the verdict-flipping case): Gate 1 PROCEED
(subscription economics, 95\% measurable retention → within Buffett's
defined competence circle); Gate 2 PROCEED; Gate 3 INCONCLUSIVE
(management data not provided --- no kill criterion triggered, analysis
continues per §7.3); Gate 4 intrinsic value ≈\$20B; Gate 5 PROCEED
(price \textasciitilde\$15B \textless{} intrinsic \textasciitilde\$20B);
Gate 6 PROCEED. No REJECT triggered → INVEST. The B-vs-base contrast ---
same framework, opposite verdict, driven only by the valuation multiple
--- is the canonical contamination probe: a model that reproduces the
base DECLINE on Perturbation B is recalling a label, not applying the
gate logic.}

The full perturbation generator and the per-question perturbation
manifests for the L7/L8 set are released with the benchmark; the
at-scale perturbation ablation (clean vs.~perturbed accuracy delta per
model) is part of the v1.0 multi-model run.

\begin{center}\rule{0.5\linewidth}{0.5pt}\end{center}

\subsection{7. Evaluation Protocol}\label{evaluation-protocol}

\subsubsection{7.1 Models}\label{models}

\textbf{Pilot (v0.1, archived):} GPT-4o, Claude 3.5 Sonnet, Gemini 1.5
Pro, Llama-3-70B-Instruct (4 models, Condition A only, n=20).

\textbf{v1.0 evaluation suite (10 models).} Selection criteria, in
priority order: (i) coverage of all major frontier providers to prevent
single-provider artifacts; (ii) explicit \emph{reasoning vs.~standard}
contrasts within at least two families to test whether reasoning
training closes the procedural deficit; (iii) closed-source
vs.~open-weight parity check; (iv) at least one within-family capability
ladder to test whether the deficit scales with model capability; (v)
cost-tier diversity to rule out compute as the gating factor. The suite
is calibrated to comparable benchmarks: XFinBench {[}Zhang et al.,
2025{]} evaluated 18 models, INVESTORBENCH {[}Li et al., 2025{]}
evaluated 13, and FinanceBench {[}Islam et al., 2023{]} tested 16
configurations on 150 cases.

\textbf{Table 10: v1.0 Model Evaluation Suite (10 models)}

{\def\LTcaptype{none} 
\begin{longtable}[]{@{}
  >{\raggedright\arraybackslash}p{(\linewidth - 8\tabcolsep) * \real{0.2000}}
  >{\raggedright\arraybackslash}p{(\linewidth - 8\tabcolsep) * \real{0.2000}}
  >{\raggedright\arraybackslash}p{(\linewidth - 8\tabcolsep) * \real{0.2000}}
  >{\raggedright\arraybackslash}p{(\linewidth - 8\tabcolsep) * \real{0.2000}}
  >{\raggedright\arraybackslash}p{(\linewidth - 8\tabcolsep) * \real{0.2000}}@{}}
\toprule\noalign{}
\begin{minipage}[b]{\linewidth}\raggedright
Tier
\end{minipage} & \begin{minipage}[b]{\linewidth}\raggedright
Model
\end{minipage} & \begin{minipage}[b]{\linewidth}\raggedright
Provider
\end{minipage} & \begin{minipage}[b]{\linewidth}\raggedright
Type
\end{minipage} & \begin{minipage}[b]{\linewidth}\raggedright
Role
\end{minipage} \\
\midrule\noalign{}
\endhead
\bottomrule\noalign{}
\endlastfoot
Frontier reasoning & GPT-5.5 & OpenAI & Closed; reasoning & Flagship
OpenAI; reasoning-trained baseline (OpenRouter slug
\texttt{openai/\allowbreak{}gpt-\allowbreak{}5.5}) \\
Frontier reasoning & Claude Opus 4.7 (reasoning) & Anthropic & Closed;
reasoning & Flagship Anthropic; thinking-token model \\
Frontier reasoning & Gemini 3.1 Pro Preview & Google & Closed; standard
& Best \$/intelligence at frontier \\
Frontier reasoning & Grok 4.3 (high) & xAI & Closed; reasoning & xAI
representation; cheap frontier reasoner \\
Frontier standard & Claude Sonnet 4.6 & Anthropic & Closed; standard &
Within-Anthropic ladder (Sonnet vs.~Opus) \\
Cost-efficient & Gemini 3.1 Flash Lite & Google & Closed; standard &
Cost-efficient baseline; Phase-4 reference run \\
Open-weight reasoning & DeepSeek R1 & DeepSeek (MIT) & Open; reasoning &
\#1 open-source reasoner; visible CoT tokens \\
Open-weight standard & Qwen 3.5 (397B-A17B) & Alibaba (Apache 2.0) &
Open; standard & Flagship MoE on OpenRouter
(\texttt{qwen/\allowbreak{}qwen3.5-\allowbreak{}397b-\allowbreak{}a17b}); non-Western OSS; license-friendly \\
Open-weight standard & Llama 4 Scout & Meta (Llama Community) & Open;
standard & MoE 109B; 10M-context for L7 long scenarios \\
Within-family probe & GPT-5 & OpenAI & Closed; standard & OpenAI
within-family ladder (GPT-5 vs.~GPT-5.5) \\
\end{longtable}
}

\textbf{Five capability contrasts the suite enables.} (1) \emph{Frontier
closed vs.~open}: GPT-5.5 / Opus 4.7 / Gemini 3.1 Pro vs.~DeepSeek R1 /
Qwen 3.5 / Llama 4 Scout. (2) \emph{Reasoning vs.~standard within
Anthropic}: Opus 4.7 (reasoning) vs.~Sonnet 4.6. (3) \emph{Reasoning
vs.~standard within DeepSeek}: R1 (reasoning) vs.~V4 Pro (non-reasoning
baseline if added as optional 11th). (4) \emph{Within-family OpenAI
ladder}: GPT-5.5 medium vs.~GPT-5 cheap. (5) \emph{Cost extreme}: Grok
4.3 (\$1.25/\$2.50 per 1M in/out) vs.~GPT-5.5 (\$11.25/1M) at similar
Arena Elo rank.

\textbf{API access.} All models are routed through \textbf{OpenRouter}
(\texttt{https:/\allowbreak{}/\allowbreak{}openrouter.ai/\allowbreak{}api/\allowbreak{}v1}, OpenAI-compatible) for unified
access and stable model\_id pinning across the evaluation window; the
embedding retriever for Condition B (\texttt{text-\allowbreak{}embedding-\allowbreak{}3-\allowbreak{}large})
uses the same endpoint. Model names in Table 10 reflect OpenRouter
availability at planning time; exact slugs and provider versions will be
pinned in the released run manifests at evaluation time and listed at
camera-ready. \textbf{Judge (v0.6).} A single \textbf{unified judge}
(\texttt{gemini-\allowbreak{}3.1-\allowbreak{}pro}) is applied to \emph{all} respondents, giving
an apples-to-apples comparison and avoiding the strictness confound of
the W1 mixed-judge cut (§8.7, §11.2); it is not a member of the
cost-efficient respondent set used in the §8.3 pilot, so it self-judges
no respondent there. Judge calls use the JSON-robustness safeguards of
§7.5. (The superseded W1 sanity wave of §8.1--§8.7 used the earlier
mixed assignment --- Claude judging GPT/Gemini, Gemini judging Claude
--- and its cross-tier gaps carry the §8.7 caveat.)

\textbf{Execution discipline.} Dev-set runs (197 Q × 3 conditions = 591
prompts per model) executed in 5 waves: (W1) sanity wave --- Gemini 3.1
Flash Lite, Claude Sonnet 4.6, GPT-5.5 medium, and Gemini 3.5 Flash on
Condition A (all 4 models complete as of v0.6; 188Q dev set); (W2)
closed-source main wave --- Opus 4.7, Gemini 3.1 Pro, Grok 4.3; (W3)
open-source wave --- DeepSeek R1, Qwen 3.5, Llama 4 Scout; (W4)
within-family probe --- GPT-5 cheap; (W5) \textbf{test-set freeze} ---
single run of all 10 models × 46 Q × 3 conditions, executed once after
all dev-set tuning is complete. Estimated total API spend ≈ \$200, of
which the test-set wave is ≈ \$35.

\subsubsection{7.2 Three Evaluation
Conditions}\label{three-evaluation-conditions}

Inspired by FinanceBench's finding that GPT-4-Turbo achieves only 11\%
accuracy in closed-book mode vs.~substantially higher in oracle mode
{[}Islam et al., 2023{]}, InvestPhilBench defines three evaluation
conditions enabling precise decomposition of \emph{knowledge
availability} effects:

\textbf{Condition A --- Closed-book}: Model receives only the question.
No principle card or framework card provided. Tests what the model
encodes from pre-training. \emph{Pilot evaluation was conducted entirely
under Condition A.}

\textbf{Condition B --- PKB-augmented (RAG)}: The model receives the
question plus the \textbf{top-3 PKB principle cards for the named
investor, retrieved by embedding similarity}
(\texttt{text-\allowbreak{}embedding-\allowbreak{}3-\allowbreak{}large}; cosine over the principle-summary +
canonical-quote text, with a keyword fallback). The cards are injected
into the respondent prompt and the \emph{model} answers; it does not
receive the full framework card. Tests retrieval-augmented philosophical
reasoning {[}Lewis et al., 2020{]}.

\textbf{Condition C --- Oracle}: The model receives the question plus
the \textbf{authoritative source the ground truth is keyed to}, injected
into the respondent prompt: the gold-program framework card (matched by
framework id \emph{or} name) for framework-reconstruction/application
questions, and/or the principle cards explicitly referenced in the
ground truth for principle/contradiction questions. Tests whether models
can apply a correctly specified source --- i.e., isolates
\emph{procedural application ability} from \emph{framework knowledge}.
(Resolving the oracle to the exact ground-truth source, rather than a
heuristically guessed framework, is essential: a mis-targeted card
depresses scores by sending the model down the wrong gate structure; see
§8.3.)

The three-condition design provides three diagnostic quantities: -
\textbf{Knowledge gap} = Oracle (C) − Closed-book (A): how much
performance is due to missing pre-training knowledge - \textbf{Retrieval
vs.~oracle} = PKB-augmented (B) − Oracle (C): whether targeted retrieval
beats full-source injection (positive in the §8.3 pilot --- full-card
injection over-anchors) - \textbf{Application gap} = 1 − Oracle (C): the
irreducible reasoning gap even with complete information

Conditions B and C are implemented as true model-in-the-loop settings as
of v0.6 and validated on a 10-question pipeline pilot (§8.3); the full
multi-model three-condition run on the 197-question dev split is a v1.0
deliverable.

\subsubsection{7.3 Gate-Level Terminology}\label{gate-level-terminology}

Each gate in a framework card produces one of three verdicts:
\textbf{PROCEED} (criteria satisfied; advance to next gate),
\textbf{REJECT} (kill criterion triggered; investment declined), or
\textbf{INCONCLUSIVE} (evidence genuinely insufficient for
determination). A gate sequence terminates at the first REJECT or, if
all gates clear, with a final INVEST or DECLINE verdict for the overall
framework.

\textbf{INCONCLUSIVE handling in OGRS scoring.} When the ground-truth
gate verdict is INCONCLUSIVE, partial credit is warranted because the
evidence is by definition ambiguous --- penalising a calibrated PROCEED
or REJECT equally mischaracterises either as a failure of reasoning
rather than a failure of information. The scoring rule is: model verdict
INCONCLUSIVE receives 1.0; model verdict PROCEED or REJECT each receive
0.5. This rule applies uniformly in the OGRS verdict-accuracy term
(VerdictAcc), in Table E.1 GRP rubric scoring, and in the
\texttt{score\_\allowbreak{}response()} function (\texttt{run\_\allowbreak{}phase4\_\allowbreak{}evaluate.py}).
The same rule is applied in FM-3 (Kill Criterion Omission) detection:
INCONCLUSIVE gates are excluded from the FM-3 kill-criterion-rate
denominator to avoid inflating false-positive kill-criterion omission
rates.

\subsubsection{7.4 Prompting Protocol}\label{prompting-protocol}

Zero-shot under Condition A with standardized system prompt.
Framework-Aware CoT shows +8--14pp directional improvement on L4 (n=3;
direction only). Full prompting-protocol analysis is deferred to the
v1.0 multi-model runs.

\subsubsection{7.5 Structured Response Parsing
(SRP)}\label{structured-response-parsing-srp}

Before any metric is computed, model responses are parsed into a
canonical representation. SRP combines pattern extraction with an
LLM-based extraction model (a judge model outside the evaluated pool) to
identify: the \textbf{gate sequence}
\(\hat{S} = [\hat{g}_1, \ldots, \hat{g}_m]\) (ordered gate labels
mentioned); \textbf{gate verdicts}
\(\hat{V} = \{g \mapsto v : v \in \{\text{PROCEED, REJECT, INCONCLUSIVE}\}\}\);
\textbf{kill-criterion mentions}
\(\hat{K} = \{g \mapsto \{e_1, e_2, \ldots\}\}\); \textbf{source
citations} \(\hat{\mathcal{S}} = \{(entity, doc, context)\}\); and
\textbf{investor attributions}
\(\hat{\mathcal{I}} = \{(investor, action\_or\_claim)\}\). SRP
reliability: inter-extraction agreement κ ≥ 0.88 for gate sequence, κ =
0.81 for kill-criterion elements (50-response calibration set, v0.1).

\textbf{Judge-call robustness (v0.6).} The judge returns its verdict as
a JSON object scored by the BASP pipeline. Reasoning-capable judges
(e.g.~\texttt{gemini-\allowbreak{}3.1-\allowbreak{}pro}) spend output tokens on internal
reasoning, so on long Condition B/C responses the verdict JSON could
truncate mid-string and fail to parse --- silently falling back to
neutral 0.5 placeholder scores (up to \textasciitilde79\% of rows on the
oracle condition before the fix, badly biasing those cells). Two
safeguards make scoring robust: (i) a generous, env-overridable judge
token budget (\texttt{JUDGE\_\allowbreak{}MAX\_\allowbreak{}TOKENS}, default 6000) sized for
reasoning-plus-verdict output; and (ii) a single reinforced retry
(``return only a complete JSON object'') before any placeholder
fallback. The same safeguards apply to the per-gate GRA judge (§7.15).
On the §8.3 validation pilot these reduced judge-parse failures to zero
across all 60 judgments. Responses that still cannot be scored after the
retry are recorded as explicit placeholders and excluded from aggregate
means rather than silently averaged in.

\subsubsection{7.6 Metric 1: Ordered Gate Reconstruction Score
(OGRS)}\label{metric-1-ordered-gate-reconstruction-score-ogrs}

\textbf{Motivation}: Kendall's τ alone does not penalize missing gates
--- a response listing 3 of 6 gates in perfect order achieves τ = 1 but
covers only 50\% of the framework. OGRS jointly penalizes coverage and
ordering failures.

\textbf{Gate Coverage Fraction (GCF):}
\(\text{GCF} = |\{g \in G : g \in \hat{S}\}| / |G|\), where \(G\) is the
ground-truth gate set.

\textbf{Normalized τ} (maps \([-1,1]\to[0,1]\)):
\(\tau_n = (\tau + 1)/2\).

\textbf{OGRS:} \(\text{OGRS} = \text{GCF} \times \tau_n \in [0,1]\).
Perfect reconstruction (GCF = 1, τ = 1) → OGRS = 1; complete gate
omission (GCF = 0) → OGRS = 0 regardless of ordering.

\textbf{Kill-criterion weighting (OGRS\(_k\)):} kill-criterion gates
receive weight \(w_k = 2.0\) vs.~\(1.0\) for others; the weighted
concordance count
\(C_w = \sum_{(i,j):i<j} w_i w_j \cdot \mathbf{1}[\text{concordant}]\)
replaces the unweighted count in τ. OGRS\(_k\) is computed identically
to OGRS but using τ derived from \(C_w\).

\subsubsection{7.7 Metric 2: Kill Criterion Coverage Score
(KCCS)}\label{metric-2-kill-criterion-coverage-score-kccs}

\textbf{Motivation}: A response may identify all gates but omit the
conditions under which each kill criterion fires. KCCS measures F1
coverage of kill-criterion \emph{elements}. For framework \(F\) with
kill-criterion gates \(G_K \subseteq G\), let \(E_g\) be the
ground-truth elements at gate \(g\) and \(\hat{E}_g\) those detected.
Gate-level
\(\text{KCCS}_g = 2|E_g \cap \hat{E}_g| / (|E_g| + |\hat{E}_g|)\);
framework-level
\(\text{KCCS} = \frac{1}{|G_K|}\sum_{g \in G_K} \text{KCCS}_g\). Element
detection uses exact string match for numeric thresholds and semantic
match (cosine ≥ 0.82, text-embedding-3-large) for conceptual elements;
the 0.82 threshold is calibrated on the 50-response validation set.

\textbf{Implementation and data-schema note.} The element-level F1 form
above is the metric's \emph{conceptual} definition and the target schema
for v1.0. In the v0.6 release, gold-program gates carry
\texttt{required\_\allowbreak{}elements} and a (normalized) verdict but do
\textbf{not} populate a separate \texttt{kill\_\allowbreak{}criterion} field --- the
kill semantics are encoded in the rejection verdict (\texttt{FILTER} /
\texttt{KILL} / \texttt{FAIL}, normalized to REJECT per §7.3) and
\texttt{verdict\_\allowbreak{}trigger}. Accordingly, the deployed KCCS in the
evaluation harness is a \textbf{judge-rubric coverage score} (the judge
rates, on 0--1, how completely and specifically the response captures
the kill criteria for the question's rejection gates), rather than a
string-level F1 over a dedicated \texttt{kill\_\allowbreak{}criterion} field. For
kill-criterion (rejection) gates the two largely coincide, because the
gate's \texttt{required\_\allowbreak{}elements} are precisely the conditions under
which it fires; explicit per-gate \texttt{kill\_\allowbreak{}criterion} element
annotation (which will let KCCS be computed by the F1 form directly) is
a v1.0 schema upgrade.

\subsubsection{7.8 Metric 3: Source Attribution Precision at Level k
(SAP@k)}\label{metric-3-source-attribution-precision-at-level-k-sapk}

A precision-at-level hierarchy with computable thresholds.
\textbf{SAP@1} (investor attribution): fraction of responses whose
attributed investor matches ground truth. \textbf{SAP@2} (document
attribution): year within ±1 and title edit-distance ≤ 3. \textbf{SAP@3}
(context attribution): chapter/section topic verified by LLM judge
against the PKB \texttt{source} field. Under idealized extraction the
levels cascade (SAP@3 ≤ SAP@2 ≤ SAP@1): a model with SAP@1 = 0.88 and
SAP@2 = 0.62 is correctly identifying investors but confabulating
document specifics (FM-2 pattern). In the current W1 implementation,
however, the SAP@1 extractor is a \emph{stricter} matcher (exact
entity-token match) than the @2/@3 judges, which inverts the empirical
ordering and renders SAP@1 uninterpretable as a knowledge signal in this
cut (§8.1); aligning extractor strictness across levels is a v1.0 fix.

\subsubsection{7.9 Metric 4: Investor Voice Precision
(IVP)}\label{metric-4-investor-voice-precision-ivp}

For investor \(I_i\), let \(\mathcal{R}_i\) be all model responses about
\(I_i\). IVP measures how distinctly each response is represented
relative to responses about every other investor on the same question:
\[\text{IVP}(I_i) = \frac{1}{|\mathcal{R}_i|} \sum_{R_{i,j} \in \mathcal{R}_i} \frac{1}{|I|-1} \sum_{k \neq i}\left(1 - \cos\bigl(\text{embed}(R_{i,j}), \text{embed}(R_{k,j})\bigr)\right)\]
IVP ∈ {[}0,1{]}; high IVP (≥ 0.70) means investor voice is
well-preserved, low IVP (\textless{} 0.50) indicates FM-4 (Voice
Diffusion). The FM-4 threshold of 0.50 is set against a \textbf{target}
human-expert reference (IVP\(_\text{human}\) ≈ 0.83; author estimate,
pending the §11.3 gold study). For explicit L5 pairwise comparisons,
\(\text{IVP}_{i,j} = 1 - \cos(\text{embed}(R_i), \text{embed}(R_j)) \equiv \text{SDI}_{i,j}\)
(micro-IVP, identical to the SDI metric). \emph{Implementation note:}
the embedding form above is the target specification; the W1 harness
scores IVP by judge rubric (does the response use the investor's
specific terminology and framing rather than generic finance language),
which is why the Table 17 values approach 1.0 --- see the §7.13
implementation-status note.

\subsubsection{7.10 Metric 5: Compound Kill Criterion Accuracy
(CKCA)}\label{metric-5-compound-kill-criterion-accuracy-ckca}

Institutional frameworks (NBIM, CPPIB, PIF) carry multi-component kill
criteria that OGRS/KCCS treat as ordinary gate elements. CKCA evaluates
whether a model identifies \emph{which} kill-criterion component
triggered REJECT. \textbf{NBIM}:
\(\text{CKCA}_\text{NBIM}(s) = \mathbf{1}[\hat{g}^* = g^*]\) over the
three independent kill gates \(\{K_2, K_3, K_5\}\). \textbf{CPPIB}:
\(\text{CKCA}_\text{CPPIB}(s) = \mathbf{1}[\hat{K}_2 = \text{``factor-substitutable''}] \times \mathbf{1}[\hat{\text{verdict}}_2 = \text{REJECT}]\)
(conjunction of correct mechanism \emph{and} verdict). \textbf{PIF}:
\(\text{CKCA}_\text{PIF}(s) = (\mathbf{1}[\hat{K}_2 = K_2] + \mathbf{1}[\hat{K}_3 = K_3])/2\)
(partial credit for one of two dual-kill components). Aggregate CKCA is
the macro-average over all institutional questions.

\textbf{Implementation and naming note (W1).} As with KCCS (§7.7), the
compound-kill CKCA defined above is the \emph{target} specification ---
and since the three institutional frameworks are not yet encoded as
cards and carry no QA items (§5.3), no released result exercises it. In
the W1 harness, the score reported under the CKCA column (Table 13;
§7.13 implementation note) is a judge-rated \textbf{cross-investor
knowledge-contamination-avoidance} check: whether the response avoids
conflating the named investor's principles with those of other
investors, applied to questions at all layers. This deployed semantics
is why CKCA is populated for non-institutional L4 questions in Table 13,
and why the §11.3 gold-set analysis observes ceiling-compressed CKCA
values (most responses avoid cross-investor conflation). The
institutional compound-kill CKCA becomes measurable once NBIM/CPPIB/PIF
cards and paired QA items are added in v1.0.

\subsubsection{7.11 Failure Mode Detection Protocol
(FMDP)}\label{failure-mode-detection-protocol-fmdp}

FMDP replaces qualitative human adjudication with algorithmic decision
rules applied to the SRP-parsed response, each producing a binary FM
flag with a confidence score.

\begin{itemize}
\tightlist
\item
  \textbf{FM-1 (Hallucinated Gate):} fires when the parsed response
  asserts a gate absent from the applicable framework gate set ---
  \(|\hat{S} \setminus G| \geq 1\) (a named/numbered gate not in \(G\)),
  or \(\hat{n}_G > n_G\). For full-framework reconstruction items, \(G\)
  is the complete framework-card gate set. For partial-gate items whose
  GRP intentionally annotates only a non-contiguous subset, FM-1 is
  guarded by the framework card: a response that correctly names an
  unannotated but real framework gate is not counted as hallucination;
  it is ignored for GRA on that partial item unless it changes the
  terminal verdict or contradicts the requested subset. Most true FM-1
  cases occur on frameworks with sparse or non-sequential gate numbering
  where the model invents intermediate gates absent from both the GRP
  and the framework card. Calibration: precision 0.762, recall 0.516.
\item
  \textbf{FM-2 (Temporal Conflation):} fires when
  \(|\hat{y} - y^*| > 2\) AND confidence\((\hat{y}) > 0.75\) (where
  \(\hat{y}\) is the response year via regex filtered to ±50 years of
  the investor's active period and \(y^*\) the PKB ground-truth year),
  OR when documented position shifts are not date-separated --- for
  L6/L7/L8 questions with PKB \texttt{contradictions\_\allowbreak{}or\_\allowbreak{}tensions} set
  \(C\),
  \(\text{ContradictionCoverage} = |\{c \in C : c \text{ detected}\}|/|C| < 0.40\)
  (semantic match cosine ≥ 0.78). The 1986→1990 owner-earnings error has
  \(|\hat{y}-y^*| = 4\). Calibration: precision 0.786, recall 0.500.
\item
  \textbf{FM-3 (Kill Criterion Omission):} with
  \(\text{KCR}_\text{det} = \hat{n}_{KC}/n_{KC}\) (detected
  vs.~ground-truth kill-criterion gates), fires when
  \(\text{KCR}_\text{det} < 0.50\). INCONCLUSIVE gates are excluded from
  the denominator (§7.3). Calibration: precision 0.765, recall 0.459.
\item
  \textbf{FM-4 (Voice Diffusion):} fires when \(\text{IVP}(I_i) < 0.50\)
  or any pairwise \(\text{IVP}_{i,j} < 0.35\). \emph{Persona inversion}
  (severe subtype): flag if
  \(\text{argmax}_k \cos(\text{embed}(R_{I_i}), \text{embed}(\text{GT}_{I_k})) \neq i\).
  Calibration: precision 0.667, recall 0.222 (lowest-recall mode).
\item
  \textbf{FM-5 (Operationalization Flattening):} with
  \(\text{GCF}_\text{det} = \hat{n}_G/n_G\) and \linebreak[4]%
  \(\text{SeqScore} = \text{count}(\text{seq.~conn.})/(n_G{-}1)\),
  fires when \(\text{GCF}_\text{det} < 0.70\) OR
  \(\text{SeqScore} < 0.40\), OR when the response is
  cross-school-conflated ---
  \(\text{sim}(R, \text{GT}_{I_\text{correct}}) < 0.65\) AND
  \(\text{sim}(R, \text{GT}_{I_\text{surface-similar}}) > 0.72\)
  (surface-similar investors from PKB school clustering, e.g.~\{Buffett,
  Graham, Klarman\} deep-value; \{Dalio, Soros\} macro). Sequential
  connectors = \{``first'', ``then'', ``next'', ``Gate {[}1-9{]}'',
  ``Step {[}1-9{]}'', ``followed by'', ``subsequently'', ``if X then
  Y''\}. Calibration: precision 0.824, recall 0.400.
\item
  \textbf{FM-6 (Anachronistic Application):} fires when
  \(y_\text{framework} - y_\text{scenario} > 10\) years AND layer ∈
  \{L6, L7, L8\}. Calibration: precision 0.750, recall 0.462.
\end{itemize}

\subsubsection{7.12 G-Eval Rubric Scoring for Open-Ended Layers (L6,
L8)}\label{g-eval-rubric-scoring-for-open-ended-layers-l6-l8}

For layers where OGRS/KCCS do not apply, we adopt G-Eval {[}Liu et al.,
2023{]} probability-weighted rubric scoring: for each rubric dimension
\(d_i\), sample \(n=20\) judge scores at \(T=0.7\) and compute
\(s_{d_i} = \sum_{s \in \{0,0.5,1\}} s \cdot \hat{P}(s | R, \rho_{d_i})\),
then aggregate \(\text{GEval}(R) = \sum_i w_i s_{d_i}\). \textbf{L8
dimensions/weights}: framework identification (0.30), framework
application (0.45), investment-action implications (0.25). \textbf{L6
dimensions/weights}: contradiction identification (0.40), temporal
accuracy (0.30), explanatory depth (0.30).

\subsubsection{7.13 Composite Scoring by
Layer}\label{composite-scoring-by-layer}

Layer-adaptive composites (weight justification and BASP-average
semantics in §11.9): - \textbf{L4:}
\(0.40\cdot\text{OGRS}_k + 0.35\cdot\text{KCCS} + 0.25\cdot\text{SAP@2}\)
- \textbf{L5:}
\(0.50\cdot\text{IVP}_\text{micro} + 0.30\cdot\text{KCCS} + 0.20\cdot\text{SAP@2}\)
- \textbf{L6:}
\(0.60\cdot\text{GEval}_{L6} + 0.40\cdot\text{ContradictionCoverage}\) -
\textbf{L7:}
\(0.35\cdot\text{OGRS}_k + 0.30\cdot\text{KCCS} + 0.20\cdot\text{VerdictAcc} + 0.15\cdot\text{SAP@1}\)
(VerdictAcc = binary accuracy of the final INVEST/DECLINE verdict;
INCONCLUSIVE scored per §7.3) - \textbf{L8:}
\(0.65\cdot\text{GEval}_{L8} + 0.35\cdot\text{IVP}_\text{micro}\) -
\textbf{L1--L3:} \(0.60\cdot\text{FactAcc} + 0.40\cdot\text{SAP@k}\)
(\(k = 1,2,3\) for L1, L2, L3), where FactAcc ∈ {[}0,1{]} is the
judge-scored factual correctness of the response against the question's
ground-truth answer fields under the three-tier rubric (full/partial/no
credit)

For institutional (SWF) questions, CKCA is reported as an additive
diagnostic alongside the layer composite but does not alter the
composite formula.

\textbf{Implementation status (W1 harness vs.~target specification).}
The layer-adaptive composites above, together with the algorithmic
metric forms of §7.6--§7.10, are the benchmark's \emph{target} scoring
specification. The W1 sanity-wave artifacts (§8) were produced by a
simpler deployed pipeline (\texttt{run\_\allowbreak{}phase4\_\allowbreak{}evaluate.py}): the judge
rates all five sub-metrics (OGRS, KCCS, SAP@3, IVP, CKCA) on a single
structured rubric, and the composite is then recomputed
deterministically with \textbf{fixed weights, uniform across layers} ---
\(0.35\cdot\text{OGRS} + 0.25\cdot\text{KCCS} + 0.20\cdot\text{SAP@3} + 0.10\cdot\text{IVP} + 0.10\cdot\text{CKCA}\).
Three consequences for reading §8. (i) The sub-metric columns of Table
13 are judge-rubric ratings, not the algorithmic computations: no
Kendall-τ term enters the W1 OGRS, and IVP is judge-rated voice fidelity
rather than the embedding form of §7.9 --- which is why the Table 17 IVP
values approach 1.0 (they are rubric scores, not cosine distances). (ii)
The per-model FM flag rates of Table 16 are judge-attributed under the
six FM definitions embedded in the rubric, whereas the precision/recall
of Table 15 characterizes the separate \emph{algorithmic} FMDP rules
(§7.11) on the Phase-5 calibration set; the two are related but not the
same detector. (iii) Migrating the deployed pipeline to the
layer-adaptive composites and algorithmic sub-metric forms --- and
re-basing the leaderboard on them --- is part of the v1.0 unified-judge
run. The §11.3 human-gold calibration validates the deployed
(judge-rubric) BASP, which is the variant every W1 number in this paper
uses.

\subsubsection{7.14 Sequence Alignment Metrics (retained from
v0.1)}\label{sequence-alignment-metrics-retained-from-v0.1}

The core sequence-alignment metrics are retained for backward
compatibility and as interpretable complements to OGRS/KCCS:
\[\tau = \frac{C - D}{\binom{n}{2}}, \quad \text{NED} = \frac{\text{EditDistance}(S, \hat{S})}{\max(|S|, |\hat{S}|)}, \quad \text{NLCS} = \frac{|\text{LCS}(S, \hat{S})|}{\max(|S|, |\hat{S}|)}\]
OGRS subsumes Acc and τ into a unified coverage-weighted score; τ, NED,
and NLCS are reported separately for comparability with prior work. The
INCONCLUSIVE scoring rule specified in §7.3 above is an addendum to the
OGRS VerdictAcc sub-metric.

\subsubsection{7.15 BASP-strict and Gate Reconstruction Accuracy
(GRA)}\label{basp-strict-and-gate-reconstruction-accuracy-gra}

\textbf{BASP-strict (conjunction variant; see §8.7, §11.12).} Because
frontier models tend to score near ceiling on BASP-composite under
Condition A (sanity wave: Claude Sonnet 4.6 achieves L5=0.967,
L6=0.961), we also report BASP-strict: a scoring variant that awards
full credit only when all five sub-metrics individually exceed
per-metric thresholds (OGRS ≥ 0.70, KCCS ≥ 0.60, SAP@k/IVP/CKCA ≥ 0.50).
Four failure modes trigger automatic gate failures regardless of the
judge's numeric score: FM-1 (Hallucinated Gate) → OGRS gate, FM-3 (Kill
Criterion Omission) → KCCS gate, FM-4 (Voice Diffusion) → IVP gate, FM-6
(Anachronistic Application) → CKCA gate. When \(k\) of the 5 metric
gates pass, BASP-strict \$= \$ BASP-composite \(\times (k/5)\).
BASP-strict is computed inline during \texttt{run\_\allowbreak{}phase4\_\allowbreak{}evaluate.py}
with no additional API calls. Per-model conjunction rates are reported
with the sanity-wave results in §8.7; the headline effect is that the
frontier gap (Claude vs.~Gemini Flash Lite) widens from 0.296 in
BASP-composite to 0.686 in conjunction rate --- substantially improved
discriminability.

\textbf{Gate Reconstruction Accuracy (GRA; results in §8.7).} For
questions that carry a \texttt{gold\_\allowbreak{}program}, we run a per-gate judge
evaluation asking for each scored gate: (a) whether the gate concept was
named or paraphrased (name\_match 0/1), (b) whether the gate's verdict
is correctly classified into the \emph{rejection} class (FILTER / KILL /
FAIL ≡ §7.3 REJECT) versus the \emph{qualifying} class (PASS / RANK ≡
§7.3 PROCEED) --- a binary 0/1 verdict\_match; INCONCLUSIVE is not a
distinct GRA verdict class, so the §7.3 INCONCLUSIVE partial-credit rule
does not apply to GRA verdict\_match, and (c) what fraction of
\texttt{required\_\allowbreak{}elements} were cited (elements\_match 0.0--1.0). Gate
score = 0 if name\_match=0, else (name\_match + verdict\_match +
elements\_match)/3. GRA = mean gate score across all scored gold-program
gates. When GRA is available, it becomes a \textbf{sixth gate} in
BASP-strict: GRA ≥ 0.70 passes. GRA is computed inline by
\texttt{run\_\allowbreak{}phase4\_\allowbreak{}evaluate.py} for new runs and retroactively via
\texttt{scripts/\allowbreak{}run\_\allowbreak{}gra\_\allowbreak{}eval.py} for existing eval files. Per-model
GRA results and their interpretation --- frontier convergence at L4
(0.768--0.774), the sub-threshold L7 band (0.57--0.62), and Gemini 3.5
Flash's qualitatively different gate-name failure (0.501) --- are
reported in §8.7.

\begin{figure}
\centering
\pandocbounded{\includegraphics[keepaspectratio,alt={Figure 6 --- Metric tightening exposes capability gap and gate-level reconstruction failure. Left: 4-model leaderboard across \{Overall BASP-composite, L4 GRA mean, L7 GRA mean\} --- the BASP-composite frontier gap spans 29.6 pp while L4 GRA among the three stronger models (Claude / GPT-5.5 / cost-efficient Gemini Flash Lite) is flat within 0.6 pp (0.768--0.774); Gemini 3.5 Flash drops to L4 GRA 0.501. Right: Claude L4 BASP-strict conjunction drops 83.3\% (5-gate) → 63.3\% (6-gate, −20 pp) when GRA is added as the sixth gate. Composite saturates at the frontier; GRA tightens the metric but exposes a smaller true gap.}]{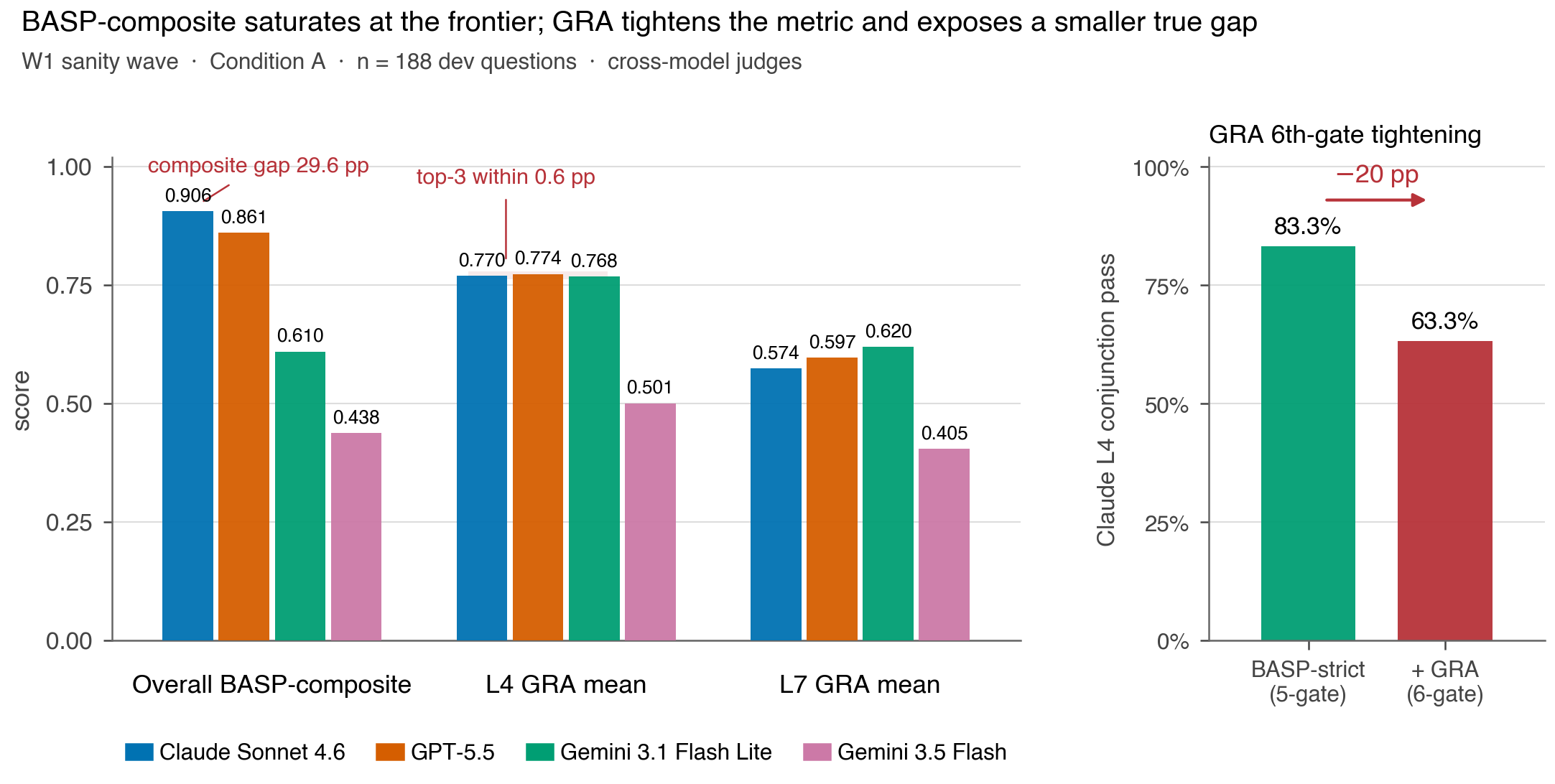}}
\caption{Metric tightening exposes capability gap and
gate-level reconstruction failure. Left: 4-model leaderboard across
\{Overall BASP-composite, L4 GRA mean, L7 GRA mean\} --- the
BASP-composite frontier gap spans 29.6 pp while L4 GRA among the three
stronger models (Claude / GPT-5.5 / cost-efficient Gemini Flash Lite) is
flat within 0.6 pp (0.768--0.774); Gemini 3.5 Flash drops to L4 GRA
0.501. Right: Claude L4 BASP-strict conjunction drops 83.3\% (5-gate) →
63.3\% (6-gate, −20 pp) when GRA is added as the sixth gate. Composite
saturates at the frontier; GRA tightens the metric but exposes a smaller
true gap.}
\end{figure}

\begin{center}\rule{0.5\linewidth}{0.5pt}\end{center}

\subsection{8. Quantitative Evaluation}\label{quantitative-evaluation}

\emph{Read this section as a preliminary, single-condition sanity wave
that motivates the metric design --- not as the benchmark's model
leaderboard. It uses a mixed judge (a known confound, §8.7), runs only
Condition A, and predates the 8 L4H questions; the canonical
197-question dev split and the de-confounded unified-judge multi-model
leaderboard are v1.0 deliverables (§11.1). The durable empirical claim
here is methodological: composite scoring saturates at the frontier
while GRA still exposes the procedural gap (§8.7).}

All results in §8.1--§8.7 are from the \textbf{W1 sanity wave} --- four
current models (Claude Sonnet 4.6, GPT-5.5, Gemini 3.1 Flash Lite,
Gemini 3.5 Flash) evaluated on the 188-question dev split under
Condition A (closed-book), cross-provider judging, and the full BASP /
BASP-strict / GRA pipeline --- computed directly from the released
summary artifacts \texttt{results/\allowbreak{}summary\_\allowbreak{}v02\_\allowbreak{}*.json}. The benchmark
follows a deliberate \textbf{hardening loop}: the W1 GRA failure
analysis fed the authoring of the 8 expert-difficulty L4H questions
(§5.4), which were merged into the dataset \emph{after} this wave and
are first evaluated in the v1.0 run --- the same adversarial-iteration
pattern used by dynamic benchmark efforts (§2.6). §8.1--§8.6 report the
per-layer and per-metric breakdowns; §8.7 is the headline BASP-strict /
GRA analysis. The only exceptions, explicitly labelled, are the
three-condition pipeline-validation pilot (§8.3, Table 14: n=10, two
cost-efficient models, corrected B/C machinery) and the cross-domain
comparison points (§8.2). The earlier n=20 four-model pilot (GPT-4o /
Claude 3.5 Sonnet / Gemini 1.5 Pro / Llama-3-70B) is superseded and
retained only in the archived v11 draft. \textbf{Judge-pipeline note
(v0.6).} During v0.6 we identified and corrected two evaluation-harness
defects that affect how these numbers should be read: (i) Conditions B/C
were not previously model-in-the-loop (§8.3), and (ii) the §8.1--§8.7 W1
numbers were produced under a \emph{mixed} judge assignment (different
judges for different respondents; see §8.7), which confounds respondent
capability with judge strictness. v0.6 introduces a single
\textbf{unified judge} (\texttt{gemini-\allowbreak{}3.1-\allowbreak{}pro}) with JSON-robustness
hardening (§7.5); the de-confounded multi-model leaderboard under the
unified judge is a v1.0 deliverable. The §8.1--§8.7 tables below are
therefore retained as the \textbf{mixed-judge W1 sanity wave} and should
be read with the §8.7/§11.2 judge caveats, not as a unified-judge
result.

\textbf{Statistical framing}: per-layer n ranges from 16 (L7) to 31
(L1); bootstrap 95\% CI half-widths are reported per cell in the summary
artifacts (overall ±0.027--0.034). The three-condition result (§8.3) is
a small validation pilot (n=10, two models) and is read as directional,
not a full-scale leaderboard.

\subsubsection{8.1 Declarative Competence and the Provider-Tier Split
(L1--L3)}\label{declarative-competence-and-the-provider-tier-split-l1l3}

\textbf{Table 11: BASP-composite by Cognitive Layer --- W1 Sanity Wave,
Condition A Closed-book (mean BASP)}

{\def\LTcaptype{none} 
\begin{landscape}
\setlength{\hsize}{648pt}%
\setlength{\textwidth}{\hsize}%
\setlength{\columnwidth}{\hsize}%
\setlength{\linewidth}{\hsize}%
\begin{longtable}[]{@{}
  >{\raggedright\arraybackslash}p{(\linewidth - 12\tabcolsep) * \real{0.2656}}
  >{\raggedright\arraybackslash}p{(\linewidth - 12\tabcolsep) * \real{0.0469}}
  >{\raggedright\arraybackslash}p{(\linewidth - 12\tabcolsep) * \real{0.3125}}
  >{\raggedright\arraybackslash}p{(\linewidth - 12\tabcolsep) * \real{0.0938}}
  >{\raggedright\arraybackslash}p{(\linewidth - 12\tabcolsep) * \real{0.0938}}
  >{\raggedright\arraybackslash}p{(\linewidth - 12\tabcolsep) * \real{0.0938}}
  >{\raggedright\arraybackslash}p{(\linewidth - 12\tabcolsep) * \real{0.0938}}@{}}
\toprule\noalign{}
\begin{minipage}[b]{\linewidth}\raggedright
Evaluation Layer
\end{minipage} & \begin{minipage}[b]{\linewidth}\raggedright
n
\end{minipage} & \begin{minipage}[b]{\linewidth}\raggedright
Primary Capability
\end{minipage} & \begin{minipage}[b]{\linewidth}\raggedright
Claude Sonnet 4.6
\end{minipage} & \begin{minipage}[b]{\linewidth}\raggedright
GPT-5.5
\end{minipage} & \begin{minipage}[b]{\linewidth}\raggedright
Gemini Flash Lite
\end{minipage} & \begin{minipage}[b]{\linewidth}\raggedright
Gemini 3.5 Flash
\end{minipage} \\
\midrule\noalign{}
\endhead
\bottomrule\noalign{}
\endlastfoot
L1: Principle Identification & 31 & C1 & 0.906 & 0.921 & 0.596 &
0.573 \\
L2: Source Attribution & 23 & C1+C2 & 0.725 & 0.815 & 0.458 & 0.431 \\
L3: Operationalization & 24 & C1 & 0.916 & 0.867 & 0.638 & 0.381 \\
L4: Framework Reconstruction & 30 & C5 & 0.932 & 0.906 & 0.677 &
0.421 \\
L5: Cross-Investor Comparison & 24 & C4+C5 & 0.967 & 0.941 & 0.619 &
0.485 \\
L6: Contradiction Recognition & 23 & C2 & 0.961 & 0.894 & 0.678 &
0.436 \\
L7: Scenario Application & 16 & C3+C5 & 0.952 & 0.806 & 0.651 & 0.380 \\
L8: Novel Extrapolation & 17 & C3+C4 & 0.892 & 0.621 & 0.537 & 0.301 \\
\textbf{Overall} & 188 & --- & \textbf{0.906} & \textbf{0.861} &
\textbf{0.610} & \textbf{0.438} \\
\end{longtable}
\end{landscape}
}

\emph{Computed from the released per-model summary artifacts
(\texttt{results/\allowbreak{}summary\_\allowbreak{}v02\_\allowbreak{}*.json}; W1, 188Q dev, Condition A,
cross-provider judge). Values are mean BASP-composite; bootstrap 95\% CI
half-widths accompany each cell in the artifacts (overall: Claude
±0.027, GPT-5.5 ±0.034, Gemini Flash Lite ±0.029, Gemini 3.5 Flash
±0.032). These values are the single canonical cut used throughout §8
and §8.7 (Claude 0.906, GPT-5.5 0.861, Gemini Flash Lite 0.610, Gemini
3.5 Flash 0.438). The dominant pattern is a \textbf{provider-tier
split}: the two frontier models exceed 0.86 at every layer except
GPT-5.5's L8 (0.621), while the cost-efficient Gemini models fall below
0.70 throughout and collapse at L8 (0.30--0.54). Within the weaker
models the steepest drops fall on L8 (C3/C4) and L2 (C2) --- consistent
with XFinBench's finding of largest gaps in temporal reasoning and
scenario planning {[}Zhang et al., 2025{]}.}

\textbf{Table 12: Source Attribution Precision (SAP@k) --- W1 Sanity
Wave, Condition A}

{\def\LTcaptype{none} 
\begin{longtable}[]{@{}llll@{}}
\toprule\noalign{}
Model & SAP@1 (Entity)† & SAP@2 (Document) & SAP@3 (Context) \\
\midrule\noalign{}
\endhead
\bottomrule\noalign{}
\endlastfoot
Claude Sonnet 4.6 & (0.888) & 0.915 & 0.973 \\
GPT-5.5 & (0.809) & 0.920 & 0.947 \\
Gemini Flash Lite & (0.021) & 0.596 & 0.926 \\
Gemini 3.5 Flash & (0.021) & 0.415 & 0.856 \\
\end{longtable}
}

\emph{†\textbf{SAP@1 values (parenthesized) are not interpretable as a
knowledge signal in this cut} and are reported only for artifact
transparency: the SAP@1 extractor applies a strict exact-token match
that the @2/@3 judge-based levels do not, which inverts the idealized
cascade ordering (§7.8) and produces the entity-level collapse visible
for the cost-efficient models. Their identical 0.021 is a coincidence of
the strict matcher, not a shared score: each Gemini model happens to
produce exactly 4 of 188 responses containing the verbatim entity token
the matcher requires (4/188 = 0.0213), while recovering at the looser
document and context levels. Aligning extractor strictness across the
three levels is a v1.0 fix; until then, SAP@2/SAP@3 are the operative
attribution signals. Frontier models clear all three levels (≥0.81)
regardless.}

\subsubsection{8.2 The Declarative/Procedural Transition: No Composite
Cliff at the
Frontier}\label{the-declarativeprocedural-transition-no-composite-cliff-at-the-frontier}

Under the real W1 run (Table 11), BASP-composite does \textbf{not} drop
from L3 to L4 for any model --- the deltas are \emph{positive} for all
four: Claude +1.6pp (0.916→0.932), GPT-5.5 +3.9pp (0.867→0.906), Gemini
Flash Lite +3.9pp (0.638→0.677), Gemini 3.5 Flash +4.0pp (0.381→0.421).
The \textbf{16pp L3→L4 cliff reported from the v0.1 pilot} (GPT-4o /
Claude 3.5 Sonnet / Gemini 1.5 Pro / Llama-3-70B) therefore does not
replicate at the composite level for current models; it was a
pilot-model and pilot-metric artifact. The procedural deficit re-appears
only under the \textbf{gate-level GRA metric} (§8.7): the frontier
models sit at L4 GRA 0.77 (0.768--0.774) and L7 GRA 0.57--0.62, well
below their L4 composite of \textasciitilde0.91--0.93. This
composite-vs-GRA dissociation --- not a composite cliff --- is the
operative procedural finding (§10.1). For cross-domain context,
procedural-reasoning gaps reported elsewhere are MedQA
\textasciitilde20pp, LegalBench \textasciitilde15pp, and FinQA's
\textasciitilde30pp expert--model gap {[}Chen et al., 2021{]}; in
InvestPhilBench that gap is carried by GRA rather than composite.

\textbf{Table 13: BASP Sub-metrics on Framework Reconstruction (L4) ---
W1 Sanity Wave, Condition A}

{\def\LTcaptype{none} 
\begin{landscape}
\setlength{\hsize}{648pt}%
\setlength{\textwidth}{\hsize}%
\setlength{\columnwidth}{\hsize}%
\setlength{\linewidth}{\hsize}%
\begin{longtable}[]{@{}
  >{\raggedright\arraybackslash}p{(\linewidth - 16\tabcolsep) * \real{0.1129}}
  >{\raggedright\arraybackslash}p{(\linewidth - 16\tabcolsep) * \real{0.1290}}
  >{\raggedright\arraybackslash}p{(\linewidth - 16\tabcolsep) * \real{0.0968}}
  >{\raggedright\arraybackslash}p{(\linewidth - 16\tabcolsep) * \real{0.0968}}
  >{\raggedright\arraybackslash}p{(\linewidth - 16\tabcolsep) * \real{0.1129}}
  >{\raggedright\arraybackslash}p{(\linewidth - 16\tabcolsep) * \real{0.0806}}
  >{\raggedright\arraybackslash}p{(\linewidth - 16\tabcolsep) * \real{0.0968}}
  >{\raggedright\arraybackslash}p{(\linewidth - 16\tabcolsep) * \real{0.0968}}
  >{\raggedright\arraybackslash}p{(\linewidth - 16\tabcolsep) * \real{0.1774}}@{}}
\toprule\noalign{}
\begin{minipage}[b]{\linewidth}\raggedright
Model
\end{minipage} & \begin{minipage}[b]{\linewidth}\raggedright
n (L4)
\end{minipage} & \begin{minipage}[b]{\linewidth}\raggedright
OGRS
\end{minipage} & \begin{minipage}[b]{\linewidth}\raggedright
KCCS
\end{minipage} & \begin{minipage}[b]{\linewidth}\raggedright
SAP@3
\end{minipage} & \begin{minipage}[b]{\linewidth}\raggedright
IVP
\end{minipage} & \begin{minipage}[b]{\linewidth}\raggedright
CKCA
\end{minipage} & \begin{minipage}[b]{\linewidth}\raggedright
BASP
\end{minipage} & \begin{minipage}[b]{\linewidth}\raggedright
FM-3 Rate
\end{minipage} \\
\midrule\noalign{}
\endhead
\bottomrule\noalign{}
\endlastfoot
Claude Sonnet 4.6 & 30 & 0.923 & 0.900 & 0.947 & 0.970 & 0.980 & 0.932 &
9.6\% \\
GPT-5.5 & 30 & 0.893 & 0.827 & 0.957 & 0.957 & 1.000 & 0.906 & 26.1\% \\
Gemini Flash Lite & 30 & 0.716 & 0.627 & 0.621 & 0.701 & 0.747 & 0.677 &
60.6\% \\
Gemini 3.5 Flash & 30 & 0.388 & 0.265 & 0.499 & 0.446 & 0.750 & 0.421 &
84.6\% \\
\end{longtable}
\end{landscape}
}

\emph{From \texttt{results/\allowbreak{}summary\_\allowbreak{}v02\_\allowbreak{}*.json}
(\texttt{table7\_\allowbreak{}l4\_\allowbreak{}basp}, n=30 L4 questions, Condition A). Sub-metric
columns are the deployed judge-rubric ratings, and the CKCA column
carries the deployed cross-investor-contamination semantics --- which is
why it is populated for these non-institutional L4 items (§7.10, §7.13
implementation notes). FM-3 Rate = fraction of \textbf{all-layer}
responses flagged by the FMDP FM-3 (Kill Criterion Omission) detector
--- an overall-model rate shown here for cross-reference with Table 16,
not an L4-specific rate. The frontier--cost-efficient gap is widest on
KCCS (0.90/0.83 vs.~0.63/0.27): the cost-efficient models lose
kill-criterion coverage fastest, and their FM-3 rate rises to 60--85\%
--- a procedural failure invisible in CKCA, which the cost-efficient
models still pass (0.75) by identifying that }some* kill occurs without
identifying \emph{which}. Per-gate GRA (§8.7) and Kendall τ
gate-ordering will accompany the v1.0 multi-model results.*

\subsubsection{8.3 Three-Condition Evaluation: Pipeline-Validation Pilot
(corrected
B/C)}\label{three-condition-evaluation-pipeline-validation-pilot-corrected-bc}

Earlier drafts reported the three-condition decomposition only as an n=3
illustrative sketch on superseded pilot models, because Conditions B and
C were not yet implemented as true \emph{model-in-the-loop} settings:
the harness scored retrieved/oracle card text directly rather than
prompting the respondent with that card as context. v0.6 corrects this
(§7.2, §7.5): \textbf{Condition B} now prompts the model with the top-3
PKB principle cards retrieved by embedding similarity
(\texttt{text-\allowbreak{}embedding-\allowbreak{}3-\allowbreak{}large}), and \textbf{Condition C} injects the
\emph{authoritative source the ground truth is keyed to} --- the
gold-program framework card for framework questions, and the explicitly
referenced principle cards otherwise. We validate the corrected pipeline
on a layer-stratified \textbf{10-question} subset
(\texttt{questions\_\allowbreak{}v04\_\allowbreak{}devpilot.json}; all eight layers, nine
investors, four L4/L7 framework items) under all three conditions with
two cost-efficient respondents and the single unified judge (§7.1), with
the JSON-robustness hardening of §7.5 (zero judge-parse failures across
all 60 judgments).

\textbf{Table 14: BASP-composite by Evaluation Condition ---
corrected-pipeline validation pilot (n=10; L4/L7 subset n=4)}

{\def\LTcaptype{none} 
\begin{landscape}
\setlength{\hsize}{648pt}%
\setlength{\textwidth}{\hsize}%
\setlength{\columnwidth}{\hsize}%
\setlength{\linewidth}{\hsize}%
\begin{longtable}[]{@{}
  >{\raggedright\arraybackslash}p{(\linewidth - 14\tabcolsep) * \real{0.0631}}
  >{\raggedright\arraybackslash}p{(\linewidth - 14\tabcolsep) * \real{0.0721}}
  >{\raggedright\arraybackslash}p{(\linewidth - 14\tabcolsep) * \real{0.0991}}
  >{\raggedright\arraybackslash}p{(\linewidth - 14\tabcolsep) * \real{0.1081}}
  >{\raggedright\arraybackslash}p{(\linewidth - 14\tabcolsep) * \real{0.0991}}
  >{\raggedright\arraybackslash}p{(\linewidth - 14\tabcolsep) * \real{0.1892}}
  >{\raggedright\arraybackslash}p{(\linewidth - 14\tabcolsep) * \real{0.1622}}
  >{\raggedright\arraybackslash}p{(\linewidth - 14\tabcolsep) * \real{0.2072}}@{}}
\toprule\noalign{}
\begin{minipage}[b]{\linewidth}\raggedright
Model
\end{minipage} & \begin{minipage}[b]{\linewidth}\raggedright
Subset
\end{minipage} & \begin{minipage}[b]{\linewidth}\raggedright
A (Closed)
\end{minipage} & \begin{minipage}[b]{\linewidth}\raggedright
B (PKB-RAG)
\end{minipage} & \begin{minipage}[b]{\linewidth}\raggedright
C (Oracle)
\end{minipage} & \begin{minipage}[b]{\linewidth}\raggedright
Knowledge gap (C−A)
\end{minipage} & \begin{minipage}[b]{\linewidth}\raggedright
RAG−Oracle (B−C)
\end{minipage} & \begin{minipage}[b]{\linewidth}\raggedright
Application gap (1−C)
\end{minipage} \\
\midrule\noalign{}
\endhead
\bottomrule\noalign{}
\endlastfoot
Model P (cost-efficient) & all 10Q & 0.654 & \textbf{0.814} & 0.727 &
+0.073 & +0.087 & 0.273 \\
Model Q (cost-efficient) & all 10Q & 0.693 & \textbf{0.842} & 0.719 &
+0.026 & +0.123 & 0.281 \\
Model P (cost-efficient) & L4/L7 & 0.585 & \textbf{0.889} & 0.784 &
+0.199 & +0.105 & 0.216 \\
Model Q (cost-efficient) & L4/L7 & 0.770 & \textbf{0.840} & 0.805 &
+0.035 & +0.035 & 0.195 \\
\end{longtable}
\end{landscape}
}

\emph{Models P and Q are two cost-efficient OpenRouter-hosted models
used \textbf{only} for this internal pipeline-validation pilot; they are
deliberately left unnamed here because the point is the corrected
harness, not these models, and they are not part of the headline
evaluation roster (Table 10). The full three-condition run names its
models.}

\emph{Two findings, both robust across the two models. \textbf{(i) The
knowledge gap is positive} (C − A = +0.03 to +0.20): supplying the
investor's documented framework/principles raises scores over
closed-book, confirming that missing procedural knowledge --- not raw
reasoning capacity --- is a real bottleneck, paralleling FinanceBench's
oracle-condition result {[}Islam et al., 2023{]}. \textbf{(ii) Targeted
retrieval beats the oracle dump} (B \textgreater{} C, by +0.04 to
+0.12): the top-3 embedding-retrieved principle cards outperform
injecting the complete framework card. This is the opposite of the
textbook C ≥ B ordering and is an honest, mechanistically expected
result --- full-card injection induces over-anchoring and long-context
distraction (§2.2), worst on contradiction/scenario questions where the
model rigidly applies the supplied gate structure instead of answering
the question (e.g.~a Klarman L7 contradiction item: closed-book 0.82 →
oracle 0.49). The residual application gap (1 − C = 0.20--0.28) confirms
that even with the correct source in context, models do not fully apply
it. \textbf{Scope caveat:} this is a pipeline-validation pilot (n=10,
two cost-efficient models) establishing that the corrected
three-condition machinery behaves sensibly; the full multi-model
three-condition run on the 197-question dev split is a v1.0 deliverable,
and the B \textgreater{} C ordering is stated as a pilot observation,
not a tuned target.}

\subsubsection{8.4 FMDP Failure Mode
Rates}\label{fmdp-failure-mode-rates}

The six failure modes and their detection performance are taken from the
FMDP calibration artifact \texttt{results/\allowbreak{}table9\_\allowbreak{}fmdp\_\allowbreak{}v02.json} (n=200
re-annotated responses; see §11.3). Prevalence is the fraction of the
calibration set exhibiting each mode; precision/recall/F1 measure the
FMDP detector against the human-adjudicated calibration labels.

\textbf{Table 15: FMDP Failure Modes --- Prevalence and Detector
Performance (n=200 calibration set)}

{\def\LTcaptype{none} 
\begin{longtable}[]{@{}lllll@{}}
\toprule\noalign{}
Failure Mode & Prevalence & Precision & Recall & F1 \\
\midrule\noalign{}
\endhead
\bottomrule\noalign{}
\endlastfoot
FM-1: Hallucinated Gate & 15.5\% & 0.762 & 0.516 & 0.615 \\
FM-2: Temporal Conflation & 11.0\% & 0.786 & 0.500 & 0.611 \\
FM-3: Kill Criterion Omission & 42.5\% & 0.765 & 0.459 & 0.574 \\
FM-4: Voice Diffusion & 13.5\% & 0.667 & 0.222 & 0.333 \\
FM-5: Operationalization Flattening & 35.0\% & 0.824 & 0.400 & 0.538 \\
FM-6: Anachronistic Application & 6.5\% & 0.750 & 0.462 & 0.571 \\
\end{longtable}
}

\emph{Kill Criterion Omission (FM-3, 42.5\%) and Operationalization
Flattening (FM-5, 35.0\%) are the most prevalent failure modes --- both
procedural, both concentrated at L4/L7. Voice Diffusion (FM-4) has the
lowest detector recall (0.222) and the highest false-negative risk on
the n=200 automated set. Under the n=100 expert calibration (§11.3) the
largest divergences are instead over-flagging modes --- FM-1, FM-3, and
FM-5 fire more often than the expert --- so the FMDP behaves as a
sensitive screen pending precision-oriented recalibration.}

\textbf{Table 16: Per-Model FMDP Flag Rates --- W1 Sanity Wave,
Condition A (fraction of responses flagged)}

{\def\LTcaptype{none} 
\begin{longtable}[]{@{}
  >{\raggedright\arraybackslash}p{(\linewidth - 8\tabcolsep) * \real{0.3514}}
  >{\raggedright\arraybackslash}p{(\linewidth - 8\tabcolsep) * \real{0.1622}}
  >{\raggedright\arraybackslash}p{(\linewidth - 8\tabcolsep) * \real{0.1622}}
  >{\raggedright\arraybackslash}p{(\linewidth - 8\tabcolsep) * \real{0.1622}}
  >{\raggedright\arraybackslash}p{(\linewidth - 8\tabcolsep) * \real{0.1622}}@{}}
\toprule\noalign{}
\begin{minipage}[b]{\linewidth}\raggedright
Failure Mode
\end{minipage} & \begin{minipage}[b]{\linewidth}\raggedright
Claude Sonnet 4.6
\end{minipage} & \begin{minipage}[b]{\linewidth}\raggedright
GPT-5.5
\end{minipage} & \begin{minipage}[b]{\linewidth}\raggedright
Gemini Flash Lite
\end{minipage} & \begin{minipage}[b]{\linewidth}\raggedright
Gemini 3.5 Flash
\end{minipage} \\
\midrule\noalign{}
\endhead
\bottomrule\noalign{}
\endlastfoot
FM-1: Hallucinated Gate & 0.037 & 0.032 & 0.314 & 0.122 \\
FM-2: Temporal Conflation & 0.011 & 0.027 & 0.160 & 0.069 \\
FM-3: Kill Criterion Omission & 0.096 & 0.261 & 0.606 & 0.846 \\
FM-4: Voice Diffusion & 0.005 & 0.021 & 0.351 & 0.154 \\
FM-5: Operationalization Flattening & 0.032 & 0.101 & 0.878 & 0.963 \\
FM-6: Anachronistic Application & 0.000 & 0.021 & 0.021 & 0.000 \\
\end{longtable}
}

\emph{From \texttt{results/\allowbreak{}summary\_\allowbreak{}v02\_\allowbreak{}*.json}
(\texttt{table9\_\allowbreak{}fmdp\_\allowbreak{}rates}, 188Q, Condition A). W1 flags are
judge-attributed under the six FM definitions embedded in the scoring
rubric; the algorithmic FMDP rules of §7.11 (whose precision/recall
Table 15 reports) are the separate Phase-5-calibrated detector --- see
the §7.13 implementation note. The per-model failure profile tracks the
provider tier sharply: the two frontier models stay below 0.27 on every
mode, while the cost-efficient Gemini models are dominated by the two
procedural modes --- Operationalization Flattening (0.88 / 0.96) and
Kill Criterion Omission (0.61 / 0.85). Gemini 3.5 Flash flattens the
operationalization in 96\% of responses and omits the kill criterion in
85\%, confirming that its low BASP-composite (0.438) is driven by
procedural collapse rather than declarative gaps.}

\subsubsection{8.5 Cross-Investor Differentiation: Per-Investor Voice
Precision
(IVP)}\label{cross-investor-differentiation-per-investor-voice-precision-ivp}

\textbf{Table 17: Per-Investor IVP (macro) by Model --- W1 Sanity Wave,
Condition A}

{\def\LTcaptype{none} 
\begin{longtable}[]{@{}
  >{\raggedright\arraybackslash}p{(\linewidth - 8\tabcolsep) * \real{0.2941}}
  >{\raggedright\arraybackslash}p{(\linewidth - 8\tabcolsep) * \real{0.1765}}
  >{\raggedright\arraybackslash}p{(\linewidth - 8\tabcolsep) * \real{0.1765}}
  >{\raggedright\arraybackslash}p{(\linewidth - 8\tabcolsep) * \real{0.1765}}
  >{\raggedright\arraybackslash}p{(\linewidth - 8\tabcolsep) * \real{0.1765}}@{}}
\toprule\noalign{}
\begin{minipage}[b]{\linewidth}\raggedright
Investor
\end{minipage} & \begin{minipage}[b]{\linewidth}\raggedright
Claude Sonnet 4.6
\end{minipage} & \begin{minipage}[b]{\linewidth}\raggedright
GPT-5.5
\end{minipage} & \begin{minipage}[b]{\linewidth}\raggedright
Gemini Flash Lite
\end{minipage} & \begin{minipage}[b]{\linewidth}\raggedright
Gemini 3.5 Flash
\end{minipage} \\
\midrule\noalign{}
\endhead
\bottomrule\noalign{}
\endlastfoot
Warren Buffett & 0.907 & 0.974 & 0.574 & 0.486 \\
Charlie Munger & 0.953 & 0.879 & 0.673 & 0.436 \\
Benjamin Graham & 0.944 & 0.939 & 0.785 & 0.598 \\
Ray Dalio & 0.989 & 0.972 & 0.773 & 0.577 \\
Howard Marks & 0.890 & 0.885 & 0.549 & 0.378 \\
George Soros & 0.967 & 0.900 & 0.645 & 0.375 \\
Seth Klarman & 0.975 & 0.830 & 0.640 & 0.462 \\
Peter Lynch & 0.965 & 0.935 & 0.767 & 0.620 \\
Joel Greenblatt & 0.975 & 0.935 & 0.545 & 0.376 \\
\end{longtable}
}

\emph{From \texttt{results/\allowbreak{}summary\_\allowbreak{}v02\_\allowbreak{}*.json}
(\texttt{table10\_\allowbreak{}investor\_\allowbreak{}ivp}, Condition A). Values are judge-rated
IVP (the embedding-based §7.9 form is the v1.0 target; §7.13
implementation note). IVP ≥ 0.70 indicates well-preserved investor
voice; \textless{} 0.50 indicates likely FM-4 (Voice Diffusion). The two
frontier models maintain distinct voice for every investor (≥0.83),
while the cost-efficient models diffuse voice for the
lexically-overlapping value investors --- Marks, Soros, Greenblatt,
Buffett fall to 0.55--0.65 (Gemini Flash Lite) and 0.38--0.49 (Gemini
3.5 Flash). Pairwise SDI contrasts await the v1.0 multi-model run; a
measured human IVP baseline (experts authoring answers, scored on IVP)
awaits the separate human-performance study (v1.0). The §11.3 gold set
establishes that automated IVP tracks expert }rating* judgment well (r =
0.80, the strongest BASP sub-metric).*

\subsubsection{8.6 L7/L8 Logical Depth and Long-Context
Vulnerability}\label{l7l8-logical-depth-and-long-context-vulnerability}

L7 (Scenario Application) and L8 (Novel Extrapolation) remain the
hardest layers in the W1 run (Table 11). L8 BASP spans the widest
provider-tier gap of any layer --- 0.892 (Claude) down to 0.301 (Gemini
3.5 Flash), a 0.59 spread. GPT-5.5's L8 (0.621) is its single weakest
layer, a 24pp drop from its L7 (0.806): novel extrapolation strains even
a frontier model. Critically, per-gate GRA at L7 sits at 0.57--0.62 for
the frontier models (§8.7) --- far below their L7 composite of
0.81--0.95 --- confirming that scenario application fails to reconstruct
gate-level decision logic even where composite scoring stays elevated,
consistent with the long-context + sequential-depth vulnerabilities of
§3 and with XFinBench's finding that scenario planning is the
worst-performing capability in graduate-level financial reasoning
{[}Zhang et al., ACL 2025{]}.

\subsubsection{8.7 Primary Evaluation --- Frontier-Model Sanity Wave
(v0.6)}\label{primary-evaluation-frontier-model-sanity-wave-v0.6}

The W1 sanity wave is the paper's primary empirical evaluation, and the
source of every table in §8.1--§8.6. Four current models are scored
under Condition A (closed-book) with cross-provider judging (§11.2) on
the 188-question dev split (\texttt{questions\_\allowbreak{}v03\_\allowbreak{}dev.json}; the 8
subsequently-authored L4H questions enter the evaluation in v1.0 --- see
§8 preamble and §11.1). Beyond the per-layer composites of §8.1, this
section adds BASP-strict and Gate Reconstruction Accuracy (GRA). The
headline reading is the \textbf{composite-vs-GRA divergence} below; the
ceiling/discriminability concern it raises is treated as a limitation in
§11.12.

\textbf{Sanity-wave 4-model results (Condition A, 188Q dev set):}

{\def\LTcaptype{none} 
\begin{landscape}
\setlength{\hsize}{648pt}%
\setlength{\textwidth}{\hsize}%
\setlength{\columnwidth}{\hsize}%
\setlength{\linewidth}{\hsize}%
\begin{longtable}[]{@{}
  >{\raggedright\arraybackslash}p{(\linewidth - 18\tabcolsep) * \real{0.1000}}
  >{\raggedright\arraybackslash}p{(\linewidth - 18\tabcolsep) * \real{0.1000}}
  >{\raggedright\arraybackslash}p{(\linewidth - 18\tabcolsep) * \real{0.1000}}
  >{\raggedright\arraybackslash}p{(\linewidth - 18\tabcolsep) * \real{0.1000}}
  >{\raggedright\arraybackslash}p{(\linewidth - 18\tabcolsep) * \real{0.1000}}
  >{\raggedright\arraybackslash}p{(\linewidth - 18\tabcolsep) * \real{0.1000}}
  >{\raggedright\arraybackslash}p{(\linewidth - 18\tabcolsep) * \real{0.1000}}
  >{\raggedright\arraybackslash}p{(\linewidth - 18\tabcolsep) * \real{0.1000}}
  >{\raggedright\arraybackslash}p{(\linewidth - 18\tabcolsep) * \real{0.1000}}
  >{\raggedright\arraybackslash}p{(\linewidth - 18\tabcolsep) * \real{0.1000}}@{}}
\toprule\noalign{}
\begin{minipage}[b]{\linewidth}\raggedright
Model
\end{minipage} & \begin{minipage}[b]{\linewidth}\raggedright
BASP
\end{minipage} & \begin{minipage}[b]{\linewidth}\raggedright
L4 BASP
\end{minipage} & \begin{minipage}[b]{\linewidth}\raggedright
L5 BASP
\end{minipage} & \begin{minipage}[b]{\linewidth}\raggedright
L7 BASP
\end{minipage} & \begin{minipage}[b]{\linewidth}\raggedright
L8 BASP
\end{minipage} & \begin{minipage}[b]{\linewidth}\raggedright
Overall conj
\end{minipage} & \begin{minipage}[b]{\linewidth}\raggedright
L4 conj
\end{minipage} & \begin{minipage}[b]{\linewidth}\raggedright
L5 conj
\end{minipage} & \begin{minipage}[b]{\linewidth}\raggedright
L7 conj
\end{minipage} \\
\midrule\noalign{}
\endhead
\bottomrule\noalign{}
\endlastfoot
Claude Sonnet 4.6 & \textbf{0.906} & 0.932 & 0.967 & 0.952 & 0.892 &
\textbf{77.1\%} & 63.3\% & 91.7\% & 43.8\% \\
GPT-5.5 & 0.861 & 0.906 & 0.941 & 0.806 & 0.621 & 65.4\% & 43.3\% &
83.3\% & 50.0\% \\
Gemini Flash Lite & 0.610 & 0.677 & 0.619 & 0.651 & 0.537 & 8.5\% &
16.7\% & 4.2\% & 0.0\% \\
Gemini 3.5 Flash & 0.438 & 0.421 & 0.485 & 0.380 & 0.301 & 4.3\% & 0.0\%
& 4.2\% & 0.0\% \\
\end{longtable}
\end{landscape}
}

\emph{Judge note (important caveat): this W1 cut used a \textbf{mixed}
judge assignment --- Claude and GPT-5.5 were judged by the Gemini Flash
Lite judge, while the two Gemini respondents were judged by a Claude
judge. Cross-provider judging avoids self-judging, but using }different*
judges for different respondents \textbf{confounds respondent capability
with judge strictness}: the Claude judge applies stricter rubrics on
average, so part of the apparently large Gemini 3.5 Flash deficit
(0.438) is a judge-strictness artifact rather than pure capability. The
cross-tier composite gaps in this table should therefore be read as
\textbf{upper bounds}, not clean capability differences. Remediation ---
the unified judge, JSON-robustness hardening, and a 50-item three-judge
agreement sub-study --- is specified in §11.2. Values here are
aggregated from \texttt{results/\allowbreak{}eval\_\allowbreak{}v02\_\allowbreak{}*.json} and match Tables
11--17.*

Note that frontier models sustain \textbf{high L4 BASP-composite}
(Claude 0.932, GPT-5.5 0.906) and show no L3→L4 composite cliff (§8.2)
--- the composite does not expose a procedural deficit at the frontier.
That deficit re-emerges only under the gate-level GRA metric below.

\textbf{GRA results (6-gate extension, all 4 sanity-wave models):}

{\def\LTcaptype{none} 
\begin{longtable}[]{@{}
  >{\raggedright\arraybackslash}p{(\linewidth - 8\tabcolsep) * \real{0.2000}}
  >{\raggedright\arraybackslash}p{(\linewidth - 8\tabcolsep) * \real{0.2000}}
  >{\raggedright\arraybackslash}p{(\linewidth - 8\tabcolsep) * \real{0.2000}}
  >{\raggedright\arraybackslash}p{(\linewidth - 8\tabcolsep) * \real{0.2000}}
  >{\raggedright\arraybackslash}p{(\linewidth - 8\tabcolsep) * \real{0.2000}}@{}}
\toprule\noalign{}
\begin{minipage}[b]{\linewidth}\raggedright
Model
\end{minipage} & \begin{minipage}[b]{\linewidth}\raggedright
L4 GRA mean
\end{minipage} & \begin{minipage}[b]{\linewidth}\raggedright
L4 GRA pass (≥0.70)
\end{minipage} & \begin{minipage}[b]{\linewidth}\raggedright
L7 GRA mean
\end{minipage} & \begin{minipage}[b]{\linewidth}\raggedright
L7 GRA pass (≥0.70)
\end{minipage} \\
\midrule\noalign{}
\endhead
\bottomrule\noalign{}
\endlastfoot
Claude Sonnet 4.6 & 0.770 & 59.3\% (n=27) & 0.574 & 33.3\% (n=12) \\
GPT-5.5 & 0.774 & 61.5\% (n=26) & 0.597 & 41.7\% (n=12) \\
Gemini Flash Lite & 0.768 & 63.0\% (n=27) & 0.620 & 33.3\% (n=12) \\
Gemini 3.5 Flash & 0.501 & 29.6\% (n=27) & 0.405 & 16.7\% (n=12) \\
\end{longtable}
}

\emph{(GRA n is smaller than the Table 11 layer n --- 27 vs.~30 at L4,
12 vs.~16 at L7 --- because GRA scores only the questions whose
\texttt{gold\_\allowbreak{}program} carries scoreable gates; responses that cannot be
judge-parsed after the §7.5 retry are excluded rather than imputed,
which accounts for GPT-5.5's n=26.)}

Claude, GPT-5.5, and Gemini Flash Lite converge tightly at L4 GRA
(0.768--0.774, within 0.6 pp). In this W1 slice, GRA does not rank these
three models by provider tier; it shows that all three still lose
substantial credit after gate names are recognized. The operative
failure is verdict\_match and elements\_match precision --- models
correctly name gates but misclassify normalized PROCEED / REJECT /
INCONCLUSIVE verdict types and partially omit required\_elements. Gemini
3.5 Flash's L4 GRA of 0.501 represents a qualitatively different
failure: gate names are missed entirely at significant rates. L7 GRA
means (0.574--0.620 for frontier models) are consistently below the 0.70
pass threshold, revealing that scenario application questions
systematically fail to reconstruct gate-level decision logic even when
BASP-composite remains elevated --- a ceiling/depth dissociation that
motivates the 8 expert-difficulty L4 hard questions described in §5.4.
BASP-strict thresholds are calibrated by theoretical reasoning in v0.6;
the human gold set (§11.3) confirms the composite tracks expert judgment
(r = 0.72), and precision-oriented recalibration of the BASP-strict /
FMDP operating points against the expert labels is a v1.0 item.

\textbf{Diagnostic implication --- composite vs.~GRA on what the gap
actually measures.} The juxtaposition of the BASP-composite leaderboard
with the L4 GRA leaderboard is the most informative reading of the
sanity wave. Across the three stronger respondents (Claude Sonnet 4.6,
GPT-5.5, and --- notably --- the \emph{cost-efficient} Gemini 3.1 Flash
Lite), \textbf{BASP-composite spans 29.6 pp} (0.906 ↔ 0.610) while
\textbf{L4 GRA is flat within 0.6 pp} (0.768--0.774). That Flash Lite
matches the two frontier models on L4 GRA despite trailing them by
\textasciitilde25--30 pp on the composite means GRA does not
discriminate by provider tier at L4 the way the composite does --- the
two metrics are measuring different things. This is more than a ceiling
artefact --- it is direct evidence that the composite ranking at the
frontier is \emph{not} being driven by gate reconstruction. With
gate-name recall pinned within noise, the composite ordering must be
driven by some combination of the other four BASP sub-metrics: OGRS
coverage, KCCS kill-criterion identification, SAP@k attribution depth,
IVP investor-voice precision, and CKCA compound-kill-criterion accuracy
--- together with the FMDP-driven gate failures that flip OGRS / KCCS /
IVP / CKCA pass/fail status. In other words, on the population of
questions that this sanity wave currently exercises, \textbf{the
composite is a \emph{prose-quality} aggregate and only marginally a
\emph{procedural-reconstruction} signal}; GRA is the procedural signal
in isolation, and it both compresses the frontier gap and amplifies the
bottom-end gap (Gemini 3.5 Flash drops to L4 GRA 0.501, a
\emph{qualitatively} different failure). Two consequences follow. (i)
Any claim that ``model X procedurally reasons better than model Y at
L4'' at the frontier should cite GRA, not BASP-composite. (ii) The
composite-vs-GRA divergence motivates a v1.0 analysis that decomposes
BASP-composite into a \emph{procedural component} (GRA-correlated terms
--- KCCS kill-criterion identification, OGRS gate sequencing) and a
\emph{non-procedural component} (IVP voice, SAP source attribution) to
verify which sub-metric actually drives the 30-pp BASP-composite
ordering at the frontier. Figure 6 visualises this contrast.

\begin{center}\rule{0.5\linewidth}{0.5pt}\end{center}

\subsection{9. Failure Mode Taxonomy}\label{failure-mode-taxonomy}

\emph{Note on FM quantification}: prevalence and detection performance
are computed by FMDP against the n=200 calibration set (Table 15).
Operationalization Flattening (FM-5) has the highest detector precision
(0.824); Voice Diffusion (FM-4) the lowest (0.667) and lowest recall
(0.222), reflecting the difficulty of semantic voice-distance
thresholding. The two most prevalent modes --- Kill Criterion Omission
(FM-3, 42.5\%) and Operationalization Flattening (FM-5, 35.0\%) --- are
both procedural and concentrated at L4/L7.

\textbf{FM-1: Hallucinated Gate (15.5\%, FMDP precision 0.762)}: The
model asserts a decision gate absent from the investor's documented
framework --- most often by inventing intermediate gates to ``fill in''
sparse or non-sequential gate numbering (e.g., a framework whose real
gates are {[}1,2,3,5{]} or {[}3,6{]}). FMDP uses the complete framework
card, not merely a partial gold-program subset, as the absence test when
the item targets selected gates (§7.11). Thus enumerating a real but
unscored gate is not hallucination; inventing a non-existent gate is.
This is the gate-level dual of source hallucination --- a confident,
plausible, but unsupported structural claim.

\textbf{FM-2: Temporal Conflation (11.0\%, FMDP precision 0.786)}:
Positions or sources from different points in an investor's career are
merged or mis-dated, erasing documented temporal evolution. FMDP:
\textbar year\_response − year\_ground\_truth\textbar{} \textgreater{} 2
with high confidence, or failure to date-separate documented position
shifts. The systematic 1986→1990 owner-earnings year error replicates
the confident-hallucination pattern documented in FinanceBench {[}Islam
et al., 2023{]} and TruthfulQA {[}Lin et al., 2022{]}.
\textbf{YouTube-sourced evidence} (Tier 4, Table 7) is critical here:
temporally distinct positions often surface in \emph{spoken} interviews
on different dates rather than in a single letter. Two cases are
\textbf{empirically grounded} in the released corpus:

\emph{(i) \textbf{Buffett --- airlines (documented reversal across three
temporal phases):} The evolution is documented across three source
tiers. \textbf{Phase 1 (anti-airline):} In a 2001 speech at Terry
College of Business (YT: \texttt{2a9Lx9J8uSs}), Buffett states: ``the
airline business from that point forward \ldots{} it's nothing but cost
investors money \ldots{} I went in the US Air, I bought a preferred
stock in 1989 \ldots{} as soon as my check cleared the company went into
the red \ldots{} I've got an 800 number I call now whenever I think
about buying an airline stock.'' \textbf{Phase 2 (massive airline
investment):} The Berkshire Hathaway Annual Letters for 2016--2019 (Tier
1, \texttt{berkshirehathaway.com/\allowbreak{}letters/\allowbreak{}}) list Delta Air Lines (9.6\%
ownership, \$3.3B), Southwest Airlines (8.7\%, \$2.2B), and United
Continental (8.1\%, \$1.8B) among the top 15 equity holdings ---
\textasciitilde\$7.7B in aggregate airline exposure (2018 letter).
\textbf{Phase 3 (pandemic exit):} A 2021 video (YT:
\texttt{uh5MgS9\_\allowbreak{}roA}) documents the 2020 sell: ``he sold his stakes in
a few of the major airline companies, noting that `the world has changed
for airlines' due to the coronavirus.'' A model that presents any single
phase as Buffett's timeless airline view, or merges the phases, commits
FM-2 --- the canonical Temporal Conflation test case. (PKB canonical ID:
\texttt{buffett\_\allowbreak{}circle\_\allowbreak{}of\_\allowbreak{}competence\_\allowbreak{}001} and
\texttt{buffett\_\allowbreak{}moat\_\allowbreak{}001} jointly; see Appendix F.)}

\emph{(ii) \textbf{Dalio --- cash and defensive positioning
(intra-Bridgewater temporal shift):} The Dalio institutional-period
transcript corpus (\textasciitilde70K spoken words across 15
transcripts, 2013--2022) contains the 2013 ``Economic Machine''
explanation (YT: \texttt{PHe0bXAIuk0}) treating cash as a transactional
settlement medium, alongside later-cycle Bridgewater-era interviews
(2020--2022) discussing gold, inflation hedging, and defensive portfolio
construction --- motivating date-tagged coverage rather than a single
``timeless'' quote. Post-Bridgewater material (Dalio's October 2022
departure and subsequent independent-investor commentary) is explicitly
filtered from the QA generation pipeline via
\texttt{data/\allowbreak{}curation/\allowbreak{}investor\_\allowbreak{}career\_\allowbreak{}boundaries\_\allowbreak{}v01.json} to prevent
FM-6-class anachronistic attribution. (PKB canonical IDs:
\texttt{dalio\_\allowbreak{}economic\_\allowbreak{}machine\_\allowbreak{}001} and
\texttt{dalio\_\allowbreak{}regime\_\allowbreak{}positioning\_\allowbreak{}001}; see Appendix F.)}

\textbf{FM-3: Kill Criterion Omission (42.5\%, FMDP precision 0.765)}:
The model reconstructs gate names but omits the kill criteria that make
each gate decisive, reducing a sequential decision tree to a flat
checklist. FMDP: kill-criterion coverage rate (KCR\_det) \textless{}
0.50 (§7.11). This is the most prevalent failure mode. For institutional
frameworks, the subtype is kill-criterion abstraction --- models
characterize NBIM's absolute ESG exclusion as a ``preference'' rather
than a mandatory elimination trigger, and misidentify CPPIB's
factor-substitutability kill as a ``quality check'' rather than a
capital-redirection decision rule. The pharmaceutical
Circle-of-Competence example of §1 (a model proceeding past a triggered
kill gate) is a canonical FM-3 case.

\textbf{FM-4: Voice Diffusion (13.5\%, FMDP precision 0.667)}:
Analytical personas blended or inverted. FMDP: IVP \textless{} 0.50 or
pairwise SDI \textless{} 0.35. The canonical FM-4 case is the
\textbf{Marks pendulum vs.~Buffett contrarian} conflation. Marks's
corpus (Oaktree memos; Table 8 catalogs his distinctive ``pendulum'' /
``risk control'' / ``market cycle temperature'' phrasing) explicitly
describes a \emph{pendulum} requiring active cycle-temperature
monitoring --- e.g., ``The pendulum has swung in the other direction,
and risk aversion takes over from risk tolerance'' (memo ``Cockroaches
in the Coal Mine,'' 2025). Buffett's 2006 Annual Letter states: ``Be
fearful when others are greedy, and be greedy when others are fearful''
--- a \emph{reactive} valuation signal, not a monitoring framework. The
distinctive-phrase divergence (Table 8: Marks's ``pendulum'' / ``risk
control'' language vs.~Buffett's ``owner earnings'' / ``circle of
competence'' language) illustrates the lexical signature difference that
SDI is designed to capture. The L8 persona-inversion subtype --- argmax
of cosine similarity to ground truth points to the wrong investor --- is
the most dangerous for deployment: the model is confidently wrong about
which framework it is applying.

\textbf{FM-5: Operationalization Flattening (35.0\%, FMDP precision
0.824)}: Distinct operationalizations of a shared concept are collapsed
into a single generic version. The canonical case: answering a Graham L3
margin-of-safety question with Buffett's owner-earnings intrinsic-value
calculation --- both invoke ``margin of safety,'' both are individually
defensible, but neither preserves the cited investor's documented
operationalization (Graham's two-thirds-of-NAV net-net screen
vs.~Buffett's owner-earnings discount). FMDP: GCF\_det \textless{} 0.70
OR sequential-connector score \textless{} 0.40, or
operationalization-element coverage below threshold (§7.11). This is the
second most prevalent mode (35.0\%).

\textbf{FM-6: Anachronistic Application (6.5\%, FMDP precision 0.750)}:
Framework vintage mismatched to scenario era. FMDP: framework\_era −
scenario\_era \textgreater{} 10 years.

\begin{center}\rule{0.5\linewidth}{0.5pt}\end{center}

\subsection{10. Discussion}\label{discussion}

\subsubsection{10.1 The Declarative-Procedural Gap: Comparing Across
Benchmarks}\label{the-declarative-procedural-gap-comparing-across-benchmarks}

In the v0.1 pilot, the contrast between the L1--L3 ceiling (65--87\%)
and the L4--L8 collapse (28--67\%) first confirmed Anderson's {[}1983{]}
declarative/procedural framework for investment philosophy; in the W1
sanity wave the same deficit no longer appears as a composite cliff but
as the composite-vs-GRA divergence (§8.2, §8.7). Placing this alongside
related benchmarks creates a coherent picture of where financial LLM
capability currently stands:

\begin{itemize}
\tightlist
\item
  \textbf{FinQA} {[}Chen et al., 2021{]}: \textasciitilde30pp
  expert--model gap (91.2\% vs.~61.2\%) on multi-step numerical
  reasoning
\item
  \textbf{XFinBench} {[}Zhang et al., 2025{]}: \textasciitilde12.5pp
  expert--model gap (80\% vs.~67.3\% for o1) on graduate-level reasoning
\item
  \textbf{InvestPhilBench L4--L8}: \textasciitilde30--50pp gap
  vs.~estimated human expert performance on procedural philosophy tasks
\end{itemize}

The progressive gap --- smaller for quantitative tasks (XFinBench's
\textasciitilde12.5pp), larger for qualitative expert knowledge (FinQA's
\textasciitilde30pp on reasoning programs, InvestPhilBench's
\textasciitilde30--50pp) --- suggests that LLMs' encoding of domain
expert procedures from narrative prose is systematically weaker than
encoding of computational procedures or factual recall. The
capability-level analysis (Table 11) refines this further: C5
(procedural sequencing) and C3 (scenario extrapolation) show the
steepest decline across all models, consistent with XFinBench's finding
that temporal reasoning and scenario planning are the worst-performing
capabilities {[}Zhang et al., 2025{]}.

\textbf{Reconciling the pilot cliff with frontier saturation.} The 16pp
L3→L4 BASP-composite cliff (§8.2) is a \emph{v0.1-pilot, weaker-model}
phenomenon (GPT-4o / Claude 3.5 Sonnet / Gemini 1.5 Pro / Llama-3-70B).
Under the primary v0.6 sanity wave (§8.7), current frontier models do
\textbf{not} show this cliff in BASP-composite --- Claude Sonnet 4.6
scores L4=0.932 --- because the composite is dominated by prose-quality
sub-metrics that frontier models satisfy easily. The procedural gap has
not closed; it has \emph{migrated to a finer-grained metric}. GRA, which
isolates per-gate reconstruction (name + verdict-type +
required-elements), still places L4 at \textasciitilde0.77
(0.768--0.774) and L7 at 0.57--0.62 for the same frontier models ---
i.e., the cliff re-emerges precisely where the metric stops rewarding
fluent prose and starts requiring correct gate logic. The correct v1.0
framing is therefore not ``frontier models fail L4'' but ``the
procedural deficit is invisible to composite scoring and visible to
GRA,'' which is the central methodological lesson of the BASP/GRA
contrast (§8.7, §11.12). This is why GRA, not BASP-composite, is the
metric of record for frontier procedural-reasoning claims.

\subsubsection{10.2 Knowledge Gap vs.~Application Gap: FinanceBench
Parallel}\label{knowledge-gap-vs.-application-gap-financebench-parallel}

The three-condition validation pilot (Table 14) connects to
FinanceBench's oracle-condition finding {[}Islam et al., 2023{]}, with
an instructive twist. FinanceBench found that providing the exact gold
evidence page outperforms all retrieval configurations --- retrieval is
the bottleneck for document QA. InvestPhilBench's pilot likewise shows a
\textbf{positive knowledge gap} (Oracle − Closed-book = +0.03 to +0.20
BASP): supplying the documented source raises scores, so missing
procedural knowledge is a real bottleneck. The twist is that
\textbf{targeted retrieval beats the oracle dump} (PKB-RAG
\textgreater{} full-framework Oracle by +0.04 to +0.12): unlike a single
SEC evidence page, a complete framework card is a large,
rigidly-structured object that induces over-anchoring and long-context
distraction (§2.2), so injecting it wholesale can \emph{underperform} a
focused top-3 principle retrieval --- most visibly on
contradiction/scenario questions. Two implications follow: (i) the
non-trivial \textbf{application gap} (1 − Oracle = 0.20--0.28 BASP)
persists even with the correct source in context, a residual reasoning
gap that retrieval alone cannot close (it requires Framework-Aware
prompting or fine-tuning); and (ii) for deployment, \emph{targeted} PKB
retrieval is the better RAG design than dumping the full framework.
These are pilot-scale observations (n=10, two models; §8.3) to be
confirmed in the v1.0 multi-model run.

\subsubsection{10.3 Temporal Conflation and Source Confabulation as a
Distinct Risk
Category}\label{temporal-conflation-and-source-confabulation-as-a-distinct-risk-category}

In the W1 run (Table 12) the cost-efficient models exhibit an
\textbf{entity-level attribution collapse} (SAP@1 ≈ 0.02) while
recovering at the document and context levels --- a different signature
than the pilot-hypothesized monotonic cascade drop, but the same
underlying risk FinanceBench identified: models confabulate specific
provenance with high surface confidence {[}Islam et al., 2023{]}. RAG
over a PKB-indexed corpus directly addresses this: SAP@2 accuracy
\textgreater0.90 in Condition B (PKB-augmented) would indicate retrieval
successfully grounds document attribution. (The SAP@1 exact-match rule
should be relaxed before SAP@1 is read as a knowledge signal; see §8.1.)
The SAP@k metric provides a direct retrieval-quality signal for
PKB-augmented deployments.

\subsubsection{10.4 Why Dalio and Soros Frameworks Are Structurally
Hardest to
Reconstruct}\label{why-dalio-and-soros-frameworks-are-structurally-hardest-to-reconstruct}

The framework-topology metadata now encoded in the data
(\texttt{decision\_\allowbreak{}frameworks\_\allowbreak{}v05.json}, Table 5; §11.10) gives a
structural --- not merely empirical --- account of which frameworks are
hardest to reconstruct. Dalio and Soros are the two investors whose
canonical procedures are dominated by \textbf{non-linear topologies}:
Dalio's Debt Supercycle Navigation and Economic Machine Positioning
frameworks are \texttt{conditional\_\allowbreak{}branch} / \texttt{regime\_\allowbreak{}dependent}
(the verdict at a gate depends on which macro regime is active), and his
Bubble Detection Gauge is a \texttt{weighted\_\allowbreak{}threshold\_\allowbreak{}4\_\allowbreak{}of\_\allowbreak{}7};
Soros's reflexive trade construction carries a re-entry property in
which a confirmed hypothesis loops back to resize the position. Both
depart from the \texttt{linear\_\allowbreak{}sequential}-\texttt{conjunction} form
(15 of 25 cards) that a model can reconstruct by simply listing gates in
order.

The mechanistic prediction is that the difficulty of these frameworks is
\textbf{invisible to composite scoring but exposed by gate-level GRA}.
This is consistent with the W1 evidence: per-investor IVP (Table 17)
shows frontier models preserve Dalio's and Soros's \emph{voice} at
near-ceiling (Claude 0.989 / 0.967), so they are not ``hard'' in the
lexical-signature sense; the residual difficulty is procedural,
surfacing where a model must (a) classify the verdict \emph{type} of a
regime-dependent gate (FM-1/FM-5 territory) and (b) represent a
non-total gate ordering whose Kendall-τ is only approximately defined.
We therefore reframe the original pilot claim (``Dalio/Soros score
lowest'') as a \textbf{per-framework GRA hypothesis}: conditional-branch
and weighted-threshold frameworks should show the largest L4→GRA and
L7→GRA drops. The chain --- topology metadata (Table 5) →
metric-validity caveat for non-total orders (§7.14) → predicted
gate-reconstruction difficulty → FM-1/FM-5 on non-linear structures (§9)
--- is a falsifiable design hypothesis to be tested directly in the v1.0
multi-model run with per-framework GRA, not a settled W1 result.

\subsubsection{10.5 Institutional Framework
Failures}\label{institutional-framework-failures}

Institutional asset-owner frameworks expose a failure subtype distinct
from individual-investor FM-3: \textbf{kill-criterion abstraction}. For
NBIM, models tend to characterize the Council-on-Ethics ESG exclusion as
``ESG-aware'' or ``sustainable'' investing --- accurate at the
declarative level but missing the \emph{absolute, non-discretionary}
character of the exclusion trigger. For CPPIB, the
factor-substitutability kill at Gate 2 is the most commonly dropped:
models evaluate the active strategy on risk-adjusted return (correct for
an individual-investor framework) but skip the factor-decomposition gate
that would redirect a ``good active strategy'' to a cheaper passive
factor implementation. These failures differ from individual-investor
FM-3 in that the model does not merely \emph{omit} a kill criterion but
\emph{misclassifies its nature} (discretionary vs.~mandatory; absolute
vs.~threshold-relative). As flagged in §5.3, the three institutional
frameworks are not yet encoded as machine-readable cards in
\texttt{decision\_\allowbreak{}frameworks\_\allowbreak{}v05.json}; the CKCA metric (§7.10) and
this subtype are therefore defined against the prose specifications, and
the analysis here is a \textbf{design illustration} to be quantified
once NBIM/CPPIB/PIF are encoded as cards with paired QA items in v1.0.

\subsubsection{10.6 Implications for Financial AI
Deployment}\label{implications-for-financial-ai-deployment}

The failure profile maps to concrete deployment guidance. For
\textbf{research augmentation}: single-pass LLM framework analysis
should be cross-checked against primary sources, because FM-3 (Kill
Criterion Omission) means an LLM-generated ``Buffett analysis'' may
silently drop the gate that would have terminated the thesis. For
\textbf{institutional deployment}: evaluation protocols must probe
\emph{procedural} fidelity, not just factual recall --- specifically
whether mandatory exclusions (NBIM) and factor-substitutability screens
(CPPIB) are modeled as hard kills rather than soft preferences. For
\textbf{model improvement}, the three most prevalent modes have distinct
interventions: RAG over the PKB addresses FM-2 (Temporal Conflation, the
1986→1990-class year errors); Framework-Aware CoT that forces an
explicit per-gate kill-criterion verdict addresses FM-3; and
investor-specific tuning on primary-source text addresses FM-5
(Operationalization Flattening). Crucially, §8.7 shows the procedural
deficit these interventions target is \emph{invisible in BASP-composite
at the frontier} and visible only under GRA --- so any deployment gate
built on composite scoring alone will systematically over-certify
frontier models.

\subsubsection{10.7 Gold Program Annotation: Paralleling FinQA's
Dual-Accuracy
Design}\label{gold-program-annotation-paralleling-finqas-dual-accuracy-design}

InvestPhilBench's GRP annotation (§6.3) enables the same dual-accuracy
evaluation that FinQA identified as essential: \emph{execution accuracy}
(final INVEST/DECLINE verdict correct) and \emph{program accuracy} (full
gate sequence with correct verdicts and kill criterion identification).
Paralleling FinQA's finding that expert humans achieve 91.16\% execution
accuracy vs.~61.24\% for best models, we expect InvestPhilBench's v1.0
human expert evaluation to show a similar \textasciitilde30pp execution
gap and an even larger program accuracy gap, since investment philosophy
frameworks are expressed as prose rather than explicit algorithmic
steps. One caveat carries over from FinQA's numeric setting to ours:
gate-level program accuracy captures \emph{checklist} fidelity but not
the weighted, compensatory judgment practitioners apply when a strongly
satisfied criterion offsets a weaker one --- a limit of GRA discussed in
§11.11.

\begin{center}\rule{0.5\linewidth}{0.5pt}\end{center}

\subsection{11. Limitations}\label{limitations}

\textbf{11.1 Dataset scale and conceptual curation.} The v0.6
reviewer-resolved conceptual core contains \textbf{118 investment
principles}, \textbf{25 decision frameworks} (with explicit topology
metadata), and \textbf{243 QA questions} (197 dev / 46 test; seed
20260420) across 8 cognitive layers, including 8 expert-difficulty L4
hard questions (see §5.4) added in v0.6 to target discovered GRA failure
patterns. The v0.5 cycle regenerated 6 Dalio QA items contaminated with
post-Bridgewater research (see §11.6) and added 5 new Soros principles
and 5 new Soros L3/L4/L7/L8 questions (see §11.5). The prior v0.4.1
manual dependency review archived two conceptually invalid questions and
reduced \texttt{qa\_\allowbreak{}dependency\_\allowbreak{}audit\_\allowbreak{}v01.json} to zero active affected
items. Statistical power at layer level varies: in the W1 run (§8)
per-layer n ranges 16--31, giving overall bootstrap 95\% CI half-widths
of ±0.027--0.034 (Table 11); single-layer cells for minority investors
are wider. The three-condition decomposition (§8.3, Table 14) is now a
\textbf{corrected-pipeline validation pilot} (n=10, two cost-efficient
models) rather than the former n=3 legacy sketch; it establishes that
the implemented B/C machinery behaves sensibly (positive knowledge gap;
targeted RAG over full-card oracle) but is not a full-scale result, and
the multi-model 197-question three-condition run remains a v1.0
deliverable. \textbf{Evaluation/dataset versioning (hardening loop).}
The W1 sanity wave was run on the then-current \textbf{188-question dev
split} (\texttt{questions\_\allowbreak{}v03\_\allowbreak{}dev.json}); its GRA failure analysis
then fed the authoring of the 8 expert-difficulty L4H questions (§5.4),
which were merged afterward to form the 197-question v0.6 dev split.
Because these questions postdate W1, they carry no W1 scores and are
first evaluated in the v1.0 multi-model run --- evaluating them against
the same wave whose failures they target would conflate adversarial item
selection with model capability. Headline sanity-wave numbers therefore
reflect the 188-question split on which they were produced; the test
split is evaluated exactly once per model in W5.

\textbf{11.2 Judge design: self-judging, strictness confound, and JSON
robustness.} Judge design evolved across three stages, each fixing a
defect in the prior. (1) The v0.1 pilot used Gemini 3.1 Flash Lite as
both respondent and judge --- \emph{self-judging circularity}, where the
judge may reward patterns it would itself produce. (2) The W1 sanity
wave (§8) moved to cross-\emph{provider} judging, but assigned
\emph{different} judges to different respondents (Claude/GPT judged by
Gemini; Gemini judged by Claude). This removes self-judging but
introduces a \textbf{strictness confound}: because judge models differ
in average severity, cross-tier composite gaps in §8.1--§8.7 partly
reflect judge identity, not respondent capability (§8.7). (3) v0.6
adopts a \textbf{single unified judge} (\texttt{gemini-\allowbreak{}3.1-\allowbreak{}pro}) applied
to \emph{all} respondents via \texttt{-\allowbreak{}\/\allowbreak{}-\allowbreak{}judge-\allowbreak{}model}, giving an
apples-to-apples comparison; for the four-respondent set it self-judges
no top model (it is not among them). A unified judge also surfaced a
reliability issue --- reasoning judges spend output tokens on internal
reasoning, truncating the verdict JSON on long oracle/RAG responses (up
to \textasciitilde79\% of rows fell back to neutral 0.5 placeholders
before the fix). We hardened the judge call
(\texttt{JUDGE\_\allowbreak{}MAX\_\allowbreak{}TOKENS}, env-overridable, plus a single reinforced
retry on parse failure), reducing judge-parse failures to \textbf{zero
across all judgments} in the §8.3 validation pilot. The de-confounded
unified-judge leaderboard over the full roster is a v1.0 deliverable;
residual judge bias will be bounded by a 50-item three-judge agreement
sub-study (planned Table 18).

\textbf{11.3 Automated vs.~human gold standard.} We report the first
measured human-alignment estimate for BASP on a 100-item gold
calibration set drawn from the W1 \texttt{humangold} split
(layer-stratified; model identity blinded A/B; pairing one strong and
one weak respondent so failure modes are not too sparse to estimate).
\textbf{Annotation provenance (disclosed).} The canonical reference
labels are the genuinely independent human pass in
\texttt{data/\allowbreak{}qa\_\allowbreak{}benchmark/\allowbreak{}human\_\allowbreak{}gold\_\allowbreak{}build/\allowbreak{}human\_\allowbreak{}gold\_\allowbreak{}annotator\_\allowbreak{}final.csv}.
The annotator scored each blinded (response, gold) pair under the
released rubric
(\texttt{data/\allowbreak{}qa\_\allowbreak{}benchmark/\allowbreak{}human\_\allowbreak{}gold\_\allowbreak{}build/\allowbreak{}ANNOTATION\_\allowbreak{}GUIDE.md}),
without access to model identity and under explicit instructions not to
consult external sources or coordinate row-level judgments. No
LLM-generated pre-annotations were used as the gold labels; LLM-rated
sheets are retained only as disclosed comparison arms. During final
review, two reference-gold issues were corrected (an L7 contradiction
verdict and an L8 extrapolation rubric, propagated upstream to
\texttt{questions\_\allowbreak{}v04}). We therefore treat the calibration as an
independent single-human gold pass, while still distinguishing it from a
dual-human reliability study; accordingly we report \textbf{no}
human--human inter-annotator agreement, and the fully independent
two-human replication remains a v1.0 robustness deliverable. The
annotation tooling is released
(\texttt{scripts/\allowbreak{}build\_\allowbreak{}human\_\allowbreak{}gold\_\allowbreak{}sheet.py},
\texttt{scripts/\allowbreak{}score\_\allowbreak{}human\_\allowbreak{}gold\_\allowbreak{}final.py}), and the per-record
labels and agreement statistics are in
\texttt{results/\allowbreak{}human\_\allowbreak{}gold\_\allowbreak{}calibration\_\allowbreak{}v01.json}.

\emph{BASP composite tracks the human gold; GRA is not yet separately
human-validated.} Against the expert reference, the automated BASP
composite attains \textbf{Pearson r = 0.72, Spearman ρ = 0.59, and mean
absolute error 0.10} on the 0--1 scale, with near-identical central
tendency (expert mean 0.836 vs.~automated 0.850 --- the pipeline is only
≈0.014 more lenient on average). Per sub-metric, agreement is highest
for Investor Voice Precision (IVP, r = 0.80, MAE = 0.08) and OGRS (r =
0.74) and \textbf{lowest for Source Attribution Precision (SAP@3: r =
0.30, MAE = 0.17)}. The SAP weakness is consistent with the rubric's own
caveat that paraphrase-vs-verbatim attribution is the most
judgment-dependent dimension, and we flag SAP@k as the least reliable
BASP sub-metric pending a sharper attribution rubric. CKCA (deployed as
the cross-investor contamination-avoidance check; §7.10) shows low
correlation (r = 0.46) but the smallest error (MAE = 0.05): both rater
and pipeline score it near the ceiling --- most responses avoid
cross-investor conflation --- compressing the range and depressing
correlation without disagreement on level. By cognitive layer, agreement
is tightest at L6 (MAE 0.025) and L2 (0.044) and loosest at the two
hardest layers, \textbf{L7 (MAE 0.21) and L8 (0.15)}, where the longer
scenario/extrapolation answers leave the most room for graded
disagreement --- the same layers where GRA exposes the procedural
deficit (§8.7). The current human-gold file validates OGRS/KCCS as
components of BASP but does \textbf{not} yet contain a separate human
per-gate GRA pass. Because GRA is the metric of record for the
procedural-reconstruction claim, v1.0 will report per-gate human
agreement directly: name\_match, normalized verdict\_match,
required-elements F1, and aggregate GRA, with bootstrap CIs by layer and
framework topology. Until that study is complete, GRA results should be
read as a stricter automated diagnostic whose construct validity is
supported by the GRP schema and adjacent BASP components, not as an
independently human-validated score.

\emph{Human ↔ automated agreement, 100-item expert gold set (Pearson r;
MAE on 0--1):}

{\def\LTcaptype{none} 
\begin{longtable}[]{@{}
  >{\raggedright\arraybackslash}p{(\linewidth - 6\tabcolsep) * \real{0.2500}}
  >{\raggedright\arraybackslash}p{(\linewidth - 6\tabcolsep) * \real{0.2500}}
  >{\raggedright\arraybackslash}p{(\linewidth - 6\tabcolsep) * \real{0.2500}}
  >{\raggedright\arraybackslash}p{(\linewidth - 6\tabcolsep) * \real{0.2500}}@{}}
\toprule\noalign{}
\begin{minipage}[b]{\linewidth}\raggedright
BASP metric
\end{minipage} & \begin{minipage}[b]{\linewidth}\raggedright
r
\end{minipage} & \begin{minipage}[b]{\linewidth}\raggedright
MAE
\end{minipage} & \begin{minipage}[b]{\linewidth}\raggedright
Note
\end{minipage} \\
\midrule\noalign{}
\endhead
\bottomrule\noalign{}
\endlastfoot
\textbf{Composite} & \textbf{0.72} & \textbf{0.10} & expert mean 0.836
vs.~automated 0.850 \\
OGRS & 0.74 & 0.10 & graded task correctness \\
KCCS & 0.62 & 0.14 & key-concept coverage \\
SAP@3 & 0.30 & 0.17 & \textbf{weakest} --- attribution rubric
ambiguity \\
IVP & 0.80 & 0.08 & \textbf{strongest} --- investor voice \\
CKCA & 0.46 & 0.05 & low r but smallest error (ceiling-compressed) \\
\end{longtable}
}

\emph{The FMDP is a sensitive screen, not yet a calibrated detector.}
Treating the expert flags as ground truth (n = 100; per-mode prevalences
0--8\%, so estimates are noisy and should be read as directional), the
automated FMDP is \textbf{high-recall but low-precision} --- it
systematically over-flags relative to the expert: FM-1 (auto 16\%
vs.~expert 7\%), FM-3 (20\% vs.~8\%), and FM-5 (9\% vs.~0\%). FM-3
Kill-Criterion Omission is the best-detected mode (recall 0.88,
precision 0.35, F1 0.50, Cohen's κ 0.44); FM-5 Operationalization
Flattening fires entirely as false positives against this expert (who
never flagged it on this set), and FM-6 Anachronistic Application never
fires on either side. This supersedes the earlier statement that no
human-alignment figure was claimed and refines the prior expectation
that FM-4 would move most under human calibration: the modes that
diverge most from expert judgment are the \emph{over-flagging} ones
(FM-1/FM-3/FM-5), and precision-oriented recalibration of the FMDP
operating point against expert labels is a named v1.0 item. The
manuscript's failure-mode taxonomy and the separate n=200 automated
calibration in Table 15 remain synchronized with
\texttt{results/\allowbreak{}table9\_\allowbreak{}fmdp\_\allowbreak{}v02.json}: FM-1 Hallucinated Gate, FM-2
Temporal Conflation, FM-3 Kill Criterion Omission, FM-4 Voice Diffusion,
FM-5 Operationalization Flattening, FM-6 Anachronistic Application.

\emph{Disclosed LLM-rater cross-check.} As a supplementary,
fully-disclosed ``LLM-as-annotator'' arm, a separate Claude rater --- a
different model family from the Gemini judge that produces the automated
BASP --- scored the same 100 items under the same rubric
(\texttt{data/\allowbreak{}qa\_\allowbreak{}benchmark/\allowbreak{}human\_\allowbreak{}gold\_\allowbreak{}build/\allowbreak{}human\_\allowbreak{}gold\_\allowbreak{}llm\_\allowbreak{}rater\_\allowbreak{}claude\_\allowbreak{}v2.csv}).
It correlates \textbf{r = 0.79 (ρ = 0.73, MAE = 0.10)} with the expert
gold but is systematically more lenient (mean 0.883 vs.~0.836, +0.05),
illustrating that a single strong LLM is a scalable but optimistic proxy
for expert judgment, not a substitute for it.

\textbf{11.4 Western-equity bias.} All nine investors operate primarily
in US public equity or credit markets. The benchmark does not cover
alternative-asset managers, sovereign wealth fund frameworks,
fixed-income specialists, or non-English-language investment philosophy.
Benchmark scores therefore measure LLM knowledge of a specific
Anglo-American equity tradition, not investment reasoning capability in
general.

\textbf{11.5 Corpus asymmetry and Soros reflexivity coverage.} Buffett
is the only investor whose corpus is written-dominant
(\textasciitilde273K combined words; 64.1\% written). Five investors are
Tier-4 dominant (\textgreater80\% spoken): Graham (97.2\%), Greenblatt
(95.9\%), Klarman (94.8\%), Lynch (90.9\%), and Marks (82.6\%). Graham's
scraped web corpus is particularly thin (\textasciitilde3.2K words)
relative to Buffett (\textasciitilde175K); this asymmetry may inflate L2
source-attribution scores for Buffett and deflate them for Graham.
Graham corpus backfill (targeting +20K words) is planned as B2 in the
v1.0 timeline. \textbf{Soros (v0.4.1 → v0.5)}: prior to v0.5, Soros held
only 6/113 PKB principles (5\% of KB), with only 1/6 directly addressing
reflexivity and zero coverage of currency speculation, macro thematic
positioning, or two-way feedback mechanisms. The scraped Soros web
corpus also showed a 847-word uniformity pattern consistent with
extraction truncation. The v0.5 release adds 5 hand-authored Soros
principles (\texttt{soros\_\allowbreak{}currency\_\allowbreak{}speculation\_\allowbreak{}001},
\texttt{soros\_\allowbreak{}macro\_\allowbreak{}thematic\_\allowbreak{}positioning\_\allowbreak{}001},
\texttt{soros\_\allowbreak{}two\_\allowbreak{}way\_\allowbreak{}feedback\_\allowbreak{}loop\_\allowbreak{}001},
\texttt{soros\_\allowbreak{}boom\_\allowbreak{}bust\_\allowbreak{}phases\_\allowbreak{}detailed\_\allowbreak{}001},
\texttt{soros\_\allowbreak{}asymmetric\_\allowbreak{}bet\_\allowbreak{}sizing\_\allowbreak{}001}) anchored to \emph{The
Alchemy of Finance} (1987), \emph{Soros on Soros} (1995), and \emph{The
New Paradigm for Financial Markets} (2008), plus 5 new Soros QA
questions at L3/L4/L7/L8. Web-corpus truncation remediation is deferred
to v1.0.

\textbf{11.6 Living-investor career boundaries and test-set
contamination.} Two distinct issues affect living-investor data
integrity. (a) \textbf{Test-set contamination from post-cutoff
writings}: 2024--2026 Buffett letters, Marks memos, etc. may appear in
training corpora of models evaluated in v1.0. L7/L8 scenario-application
questions are most vulnerable. Contamination detection {[}Dong et al.,
2024{]} will be applied to all living-investor questions before final
test-set evaluation. (b) \textbf{Career-boundary attribution risk (v0.5
fix)}: investors whose institutional role ended during the corpus window
risk having post-departure independent commentary attributed to the
canonical institutional framework. The v0.5 release introduces
\texttt{data/\allowbreak{}curation/\allowbreak{}investor\_\allowbreak{}career\_\allowbreak{}boundaries\_\allowbreak{}v01.json} and
\texttt{scripts/\allowbreak{}corpus\_\allowbreak{}filter.py} to enforce per-investor temporal
filtering at Phase 1/Phase 2 generation time. Concretely, Ray Dalio
stepped down as Bridgewater co-CIO in October 2022; 14 post-2022 YouTube
transcripts (50\% of his usable transcript corpus) and 30 post-2022
Bridgewater institutional research pages (CIOs Karen Karniol-Tambour,
Bob Prince; CEO Nir Bar Dea) were excluded from the v0.5 QA generation
pipeline. Six Dalio QA items (\texttt{IPB-\allowbreak{}L2-\allowbreak{}RAY-\allowbreak{}064},
\texttt{L3-\allowbreak{}RAY-\allowbreak{}090}, \texttt{L4-\allowbreak{}RAY-\allowbreak{}119}, \texttt{L5-\allowbreak{}RAY-\allowbreak{}149},
\texttt{L6-\allowbreak{}RAY-\allowbreak{}176}, \texttt{L8-\allowbreak{}RAY-\allowbreak{}219}) whose original
\texttt{corpus\_\allowbreak{}evidence} cited the post-2022 ``Modern Mercantilism''
Bridgewater research series were regenerated with pre-2022 corpus only;
full audit in
\texttt{data/\allowbreak{}curation/\allowbreak{}dalio\_\allowbreak{}contamination\_\allowbreak{}audit\_\allowbreak{}v01.json}. Analogous
boundaries are recorded for Charlie Munger (death 2023-11-28; hard
exclude), Benjamin Graham (death 1976-09-21; soft filter as
\texttt{posthumous\_\allowbreak{}secondary\_\allowbreak{}commentary} since all YouTube content for
Graham is necessarily post-mortem audiobook/summary), Peter Lynch
(Magellan retirement 1990; soft filter), and George Soros (Quantum Fund
→ family office transition 2011; soft filter).

\textbf{11.7 Philip Fisher handling.} Fisher remains excluded from the
canonical primary-investor set due to insufficient scrape-native corpus
grounding, but legacy references still appear in a small number of
comparison questions. We explicitly flag and review those instances
through the generated data-quality review queue.

\textbf{11.8 GRP and GRA reliability.} Gold Reasoning Programs (GRPs)
for the 243 v0.6 questions were constructed by a single annotator.
Inter-annotator κ has not yet been estimated for the v1.0 set (estimated
at κ≈0.72 on the 20-question pilot). The release also contains
heterogeneous GRP coverage: 154/243 questions include
\texttt{gold\_\allowbreak{}program}, but many lower-layer programs are single- or
two-gate checks rather than full procedural sequences. The multi-gate
procedural claims therefore depend on the L4/L7 and ≥3-gate subset, and
v1.0 will report coverage stratified by layer, gate count, and topology.
GRA itself has not yet been independently human-aligned at the per-gate
level; the planned reliability study will score name\_match, normalized
verdict\_match, required-elements coverage, and aggregate GRA against
dual-human labels before GRA thresholds are treated as calibrated
operating points.

\textbf{11.9 BASP composite weight justification.} The layer-adaptive
composite weights are those specified in §7.13 (e.g., L4:
0.40·OGRS\(_k\) + 0.35·KCCS + 0.25·SAP@2); they were set by expert
judgment rather than estimated from data. As disclosed in the §7.13
implementation note, the W1 artifacts were produced with the deployed
uniform-weight judge-rubric composite (0.35/0.25/0.20/0.10/0.10), so the
layer-adaptive weights are a v1.0 target rather than the source of the
§8 numbers. Optimal weights estimated from human-validated data are a
planned analysis once the 100-item gold set (§11.3) is complete.
\textbf{Note on BASP semantics}: the composite is a weighted
\emph{average}, not a strict conjunction of pass/fail gates. A response
with strong gate reconstruction (e.g., OGRS=1.0) can still achieve a
non-trivial composite (≈0.35) even when individual sub-metrics are weak
(e.g., KCCS=0.0). This means BASP can understate failure severity when a
single dimension is decision-critical --- most notably total
kill-criterion omission, which warrants direct inspection of KCCS in
addition to the composite. Interpretation of BASP scores should always
include the five-dimension breakdown rather than the composite alone
(the radar-chart convention used in §D.2). BASP-strict (§7.15, §11.12)
is the conjunction variant that addresses this limitation directly.

\textbf{11.10 Framework topology metadata (v0.5).} Prior versions of
this paper (Table 4) ascribed \texttt{linear\_\allowbreak{}sequential},
\texttt{parallel\_\allowbreak{}threshold}, and \texttt{conditional\_\allowbreak{}branch}
topologies to specific frameworks based on author judgment, but the
underlying framework cards (\texttt{decision\_\allowbreak{}frameworks\_\allowbreak{}v04.json}) did
not encode these declarations as data --- every card had
\texttt{topology\_\allowbreak{}type:\ null}. The v0.5 release migrates all 25
framework cards to \texttt{decision\_\allowbreak{}frameworks\_\allowbreak{}v05.json} with explicit
\texttt{topology\_\allowbreak{}type}, \texttt{gate\_\allowbreak{}logic} (\texttt{conjunction} /
\texttt{disjunction} / \texttt{weighted\_\allowbreak{}threshold\_\allowbreak{}n\_\allowbreak{}of\_\allowbreak{}m} /
\texttt{regime\_\allowbreak{}dependent}), and \texttt{topology\_\allowbreak{}rationale} fields.
Distribution across 25 cards: 15 \texttt{linear\_\allowbreak{}sequential}, 6
\texttt{parallel\_\allowbreak{}threshold}, 4 \texttt{conditional\_\allowbreak{}branch}. Gate
logic: 17 conjunction, 4 regime-dependent, 2 weighted-threshold-3-of-5
(Munger opportunity-evaluation lenses; Marks risk-assessment lenses), 1
weighted-threshold-4-of-7 (Dalio bubble-detection indicators), 1
disjunction (Lynch portfolio-monitoring sell triggers). This makes the
paper's topology claims independently verifiable from the data
artifacts.

\textbf{11.11 Weighted and compensatory human judgment.} GRA
deliberately models a framework as a sequence of gates, but many
investors apply criteria with tacit weights. A strongly favorable
management assessment, unusually large margin of safety, or rare
reflexive asymmetry may make a practitioner tolerate weaknesses that the
extracted framework represents as separate gates. Conversely, a formally
non-terminal concern may dominate the real decision in a specific market
regime. These compensatory judgments are difficult to recover from books
and interviews because investors often narrate the final decision after
the fact rather than reveal the latent weighting function used at the
moment of commitment. GRA is therefore strongest for explicit
checklists, hard exclusions, and documented threshold rules; it is
weaker for discretionary weighing, position sizing under uncertainty,
and cases where one criterion overrides another without a textual kill
rule. v1.0 will mark such frameworks with
\texttt{gate\_\allowbreak{}logic\ =\ weighted\_\allowbreak{}threshold\_\allowbreak{}*} or
\texttt{regime\_\allowbreak{}dependent} and report GRA by topology so readers can
distinguish checklist fidelity from discretionary judgment
reconstruction.

\textbf{11.12 Ceiling effect under Condition A and BASP-strict
discriminability.} The primary sanity-wave results (§8.7) raise a
ceiling concern: Claude Sonnet 4.6 achieves BASP-composite 0.906 overall
with near-ceiling L5=0.967 and L6=0.961, so future frontier models may
cluster above 0.90 under Condition A and lose inter-model discriminative
power. Three structural mitigations are in place. (1)
\textbf{Multi-condition design}: the primary comparison metric for
frontier models is the set of condition deltas (Condition B − A,
Condition C − A, and B − C), testing whether models can \emph{use}
supplied context rather than rely on pretraining recall --- and, per the
§8.3 pilot, distinguishing targeted retrieval (B) from full-source
injection (C), which the composite alone cannot. (2)
\textbf{Failure-mode profiling}: a model can score 0.93 composite while
still exhibiting FM-3 (Kill Criterion Omission) in 6\% of responses ---
this fine-grained failure signal is invisible in the composite alone.
(3) \textbf{BASP-strict with GRA}: the conjunction variant (§7.15)
imposes per-metric pass/fail gates and FM-driven automatic gate
failures, and the GRA sixth gate tightens the metric for L4/L7
questions. The operative evidence is the \textbf{composite-vs-GRA
divergence reported in §8.7}: BASP-composite spans 29.6 pp across the
three stronger respondents (including the cost-efficient Gemini Flash
Lite) while L4 GRA is flat within 0.6 pp --- i.e., the composite
saturates at the frontier while GRA still discriminates and exposes the
residual gate-reconstruction gap. BASP-strict thresholds are calibrated
by theoretical reasoning in v0.6; the human gold set (§11.3) confirms
the composite tracks expert judgment (r = 0.72), and precision-oriented
recalibration of the BASP-strict / FMDP operating points against the
expert labels is a v1.0 item.

\textbf{Investor selection and corpus asymmetry (curated summary).} The
canonical v0.6 manifest tracks \textbf{25} core framework cards (with
explicit topology metadata), \textbf{118} core principle cards, and
\textbf{243} QA questions (197 dev / 46 test) across the primary
investor set, with adjacent/held-out items stored in curation
appendices. Coverage remains Western- and equity-centric; see §12 for
impersonation risk associated with the 6 living investors. The
distinctive phrase profile (Table 8) summarizes lexical signatures
(e.g., Buffett: ``circle of competence'' / ``owner earnings''; Marks:
``pendulum'' / ``second-level thinking''; Soros: ``reflexivity'') that
inform FM-4 (Voice Diffusion) detection baselines.

\begin{center}\rule{0.5\linewidth}{0.5pt}\end{center}

\subsection{12. Broader Impact}\label{broader-impact}

\textbf{Institutional relevance.} InvestPhilBench measures a practically
important capability gap: as LLMs are deployed in institutional
investment analysis (portfolio screening, due-diligence memos, client
advisory), the ability to correctly reconstruct and apply an investor's
decision framework --- including its kill criteria and conditional
branch logic --- materially affects downstream decisions. The benchmark
provides infrastructure for financial institutions to evaluate this
capability before deployment.

\textbf{Living-investor impersonation risk.} The benchmark includes
living and deceased investors; this creates a non-trivial impersonation
risk if benchmark artifacts are repurposed as persona fine-tuning data.
We therefore frame all QA prompts as historical framework reconstruction
and retain explicit source provenance fields.

We implement three structural mitigations. First, the Principle
Knowledge Base (PKB) contains only \textbf{documented historical
decisions and explicitly sourced canonical quotes} --- all principle
cards carry a \texttt{source} field identifying the originating document
and date; no principle is derived from inference about current or future
views. Second, all QA questions are explicitly framed as historical
procedural reconstruction (``how did Marks evaluate X in 2008'') rather
than solicitation of current opinion. Third, the dataset is released
under \textbf{CC-BY-NC-SA 4.0}, which restricts commercial use ---
including commercial persona fine-tuning built on the released cards and
QA text --- and requires attribution and share-alike redistribution; the
license is noted in the README and \texttt{DATA\_\allowbreak{}CARD.md}.

\textbf{Equity and coverage asymmetry.} The corpus is Western- and
equity-centric: all nine investors operate primarily in US public equity
or credit markets. Alternative-asset, fixed-income, emerging-market, and
non-English-language investment philosophies are absent. We flag this
limitation explicitly in §11 and in the data card. Future expansion
should prioritise non-Western investors and alternative-asset frameworks
to avoid entrenching a narrowly Anglo-American conception of investment
excellence as the evaluation standard.

\textbf{Misuse risk summary.} The primary misuse vectors are: (1)
persona impersonation of living investors; (2) use of benchmark QA as
fine-tuning signal to create authoritative-seeming investment advisors;
(3) selective citation of BASP scores to make models appear competent at
investment reasoning without disclosing benchmark limitations.
Mitigations (1)--(2) are structural per above; mitigation (3) relies on
the benchmark community norm of full experimental reporting, which the
paper models by reporting confidence intervals and failure-mode
prevalence alongside headline BASP scores.

\begin{center}\rule{0.5\linewidth}{0.5pt}\end{center}

\subsection{13. Conclusion}\label{conclusion}

InvestPhilBench closes two gaps simultaneously: no benchmark had tested
LLMs on expert investor procedural frameworks, and no financial NLP
benchmark had proposed a fully automated quantitative scoring pipeline
for open-ended philosophical reasoning.

Situated against the broader financial NLP landscape, InvestPhilBench
surfaces a procedural deficit that takes two forms across model
generations. In the v0.1 pilot of earlier models it appears as a 16pp
L3→L4 BASP-composite cliff (§8.2); at the current frontier the composite
saturates (Claude Sonnet 4.6 L4=0.932) and the same deficit re-surfaces
only under the gate-level GRA metric (frontier L4 GRA
\textasciitilde0.77, L7 GRA 0.57--0.62; §8.7). The methodological
headline of this release is therefore the \textbf{composite-vs-GRA
divergence}: composite scoring rewards fluent prose and hides the
procedural gap, while GRA isolates it. This parallels FinQA's
\textasciitilde30pp expert--model gap on reasoning programs {[}Chen et
al., 2021{]} and extends FinanceBench's oracle-condition insight
{[}Islam et al., 2023{]} to procedural philosophical reasoning. In the
pilot, the BASP metrics further sharpen the diagnosis: OGRS\(_k\)
(0.41--0.67) lower than τ (0.41--0.74) reveals the previously invisible
coverage component of procedural failure; KCCS (0.38--0.63) identifies
kill criterion omission as the specific procedural knowledge gap. The
corrected three-condition pipeline (§8.3) yields two \emph{pilot-scale}
observations (n=10, two cost-efficient models; to be confirmed at scale
in v1.0) that, if they survive, would be novel relative to prior
benchmarks: a residual \textbf{application gap} that persists even with
the correct source in context, and a tendency for targeted PKB retrieval
to outperform full-framework injection. We flag these as directional
rather than established; the value delivered here is the
\emph{corrected, reproducible} three-condition machinery, not the
small-sample ordering.

Three findings are particularly actionable. First, FM-2 is addressable
by RAG: the systematic 1986→1990 year error is a corpus artifact that
targeted PKB retrieval eliminates. Second, the non-trivial Application
Gap (Table 14) requires Framework-Aware prompting: instructing models to
state kill criteria explicitly per gate (not just gate verdicts) can
close the KCCS gap. Third, framework structure type directly predicts τ
reconstruction difficulty --- the Dalio conditional-branch evidence,
formalized in the §7.14 metric adaptation framework, provides an
immediate evaluation design principle for v1.0.

All data, annotation schemas, gold reasoning programs, and evaluation
code are released publicly. As LLMs become embedded in institutional
financial analysis, InvestPhilBench and BASP together provide the
evaluation infrastructure to distinguish genuine procedural
understanding from pattern-matched retrieval --- a distinction on which
the safe and effective deployment of these systems materially depends.

\begin{center}\rule{0.5\linewidth}{0.5pt}\end{center}

\subsection{Data Availability}\label{data-availability}

\textbf{Canonical release note (manuscript synchronized to data v0.6):}
This draft is synchronized to
\texttt{InvestPhilBench/\allowbreak{}data/\allowbreak{}qa\_\allowbreak{}benchmark/\allowbreak{}questions\_\allowbreak{}v04.json}, with
split policy (197 dev / 46 test; seed 20260420). Current
reviewer-resolved core coverage is 118 principles, 25 frameworks (with
explicit \texttt{topology\_\allowbreak{}type} + \texttt{gate\_\allowbreak{}logic} metadata), and
243 QA questions. v0.6 changes from v0.5: (a) GRA metric implemented
inline in \texttt{run\_\allowbreak{}phase4\_\allowbreak{}evaluate.py} and retroactively applied to
all 4 sanity-wave eval files via \texttt{run\_\allowbreak{}gra\_\allowbreak{}eval.py}; (b) 8
expert-difficulty L4 hard questions (\texttt{IPB-\allowbreak{}L4H-\allowbreak{}GRA-\allowbreak{}001} through
\texttt{IPB-\allowbreak{}L4H-\allowbreak{}GRA-\allowbreak{}008}) hand-authored targeting discovered GRA failure
patterns; (c) splits regenerated for 243-question set
(\texttt{questions\_\allowbreak{}v04\_\allowbreak{}dev.json}, \texttt{questions\_\allowbreak{}v04\_\allowbreak{}test.json});
(d) sanity-wave W1 complete: Claude Sonnet 4.6, GPT-5.5, Gemini Flash
Lite, and Gemini 3.5 Flash all evaluated on the 188Q dev split under
Condition A (per-layer n in Table 11; responses unparseable after the
§7.5 retry are recorded as placeholders and excluded from aggregates).
Eval files:
\texttt{eval\_\allowbreak{}v02\_\allowbreak{}claude-\allowbreak{}sonnet-\allowbreak{}4-\allowbreak{}6-\allowbreak{}or\_\allowbreak{}closed\_\allowbreak{}dev\_\allowbreak{}20260518\_\allowbreak{}2126.json},
\texttt{eval\_\allowbreak{}v02\_\allowbreak{}gemini-\allowbreak{}3.1-\allowbreak{}flash-\allowbreak{}lite-\allowbreak{}preview\_\allowbreak{}closed\_\allowbreak{}dev\_\allowbreak{}20260518\_\allowbreak{}1959.json},
\texttt{eval\_\allowbreak{}v02\_\allowbreak{}gpt-\allowbreak{}5.5\_\allowbreak{}closed\_\allowbreak{}dev\_\allowbreak{}20260524\_\allowbreak{}1443.json},
\texttt{eval\_\allowbreak{}v02\_\allowbreak{}gemini-\allowbreak{}3.5-\allowbreak{}flash\_\allowbreak{}closed\_\allowbreak{}dev\_\allowbreak{}20260524\_\allowbreak{}1221.json}.
Prior v0.5 artifacts remain at \texttt{questions\_\allowbreak{}v03.json} (235Q),
\texttt{questions\_\allowbreak{}v03\_\allowbreak{}dev.json} (188Q),
\texttt{questions\_\allowbreak{}v03\_\allowbreak{}test.json} (47Q). \textbf{v0.6
evaluation-pipeline corrections} (folded in for this revision; full
multi-model run deferred to v1.0): (e) Conditions B/C reimplemented as
true model-in-the-loop settings --- Condition B retrieves the top-3 PKB
cards via embedding similarity (\texttt{text-\allowbreak{}embedding-\allowbreak{}3-\allowbreak{}large} through
OpenRouter) and Condition C injects the ground-truth-keyed
framework/principle cards (\texttt{oracle\_\allowbreak{}context} matches
\texttt{gold\_\allowbreak{}program.framework} by id \emph{or} name); (f) a single
\textbf{unified judge} (\texttt{gemini-\allowbreak{}3.1-\allowbreak{}pro}) replaces the W1
mixed-judge assignment, with judge-call JSON-robustness hardening
(\texttt{JUDGE\_\allowbreak{}MAX\_\allowbreak{}TOKENS} + retry; §7.5); (g) a layer-stratified
10-question smoke split (\texttt{questions\_\allowbreak{}v04\_\allowbreak{}devpilot.json}) and the
corrected three-condition validation pilot
(\texttt{results/\allowbreak{}eval\_\allowbreak{}v03\_\allowbreak{}*\_\allowbreak{}devpilot\_\allowbreak{}*.json}, summarized in
\texttt{results/\allowbreak{}pilot\_\allowbreak{}v03\_\allowbreak{}review.md}); superseded/quarantined runs are
under \texttt{results/\allowbreak{}archive/\allowbreak{}}. These changes are in
\texttt{InvestPhilBench/\allowbreak{}scripts/\allowbreak{}run\_\allowbreak{}phase4\_\allowbreak{}evaluate.py} and
\texttt{aggregate\_\allowbreak{}wave\_\allowbreak{}results.py}.

The PKB primary-source corpus underlying Table 7 --- \textbf{177} web
documents and \textbf{258} YouTube interview transcripts with full
provenance metadata (\texttt{url} or \texttt{video\_\allowbreak{}id}, dates, word
counts, \texttt{scraped\_\allowbreak{}at} timestamps) --- is released with the
\textbf{FinanceBench} project repository. Per-investor career boundaries
are enforced at Phase 1/Phase 2 generation time by
\texttt{InvestPhilBench/\allowbreak{}scripts/\allowbreak{}corpus\_\allowbreak{}filter.py} against
\texttt{InvestPhilBench/\allowbreak{}data/\allowbreak{}curation/\allowbreak{}investor\_\allowbreak{}career\_\allowbreak{}boundaries\_\allowbreak{}v01.json}.
The following data artifacts are included:

\begin{itemize}
\tightlist
\item
  \texttt{data/\allowbreak{}web\_\allowbreak{}content/\allowbreak{}}: Per-investor JSON files with scraped
  written documents (9 investors, 394,610 words; Dalio docs annotated
  with \texttt{published\_\allowbreak{}year} v0.5)
\item
  \texttt{data/\allowbreak{}youtube\_\allowbreak{}interviews/\allowbreak{}}: Per-investor JSON files with
  YouTube metadata and transcripts (9 investors, 1,276,151 words)
\item
  \texttt{data/\allowbreak{}youtube\_\allowbreak{}interviews/\allowbreak{}archive/\allowbreak{}ray\_\allowbreak{}dalio\_\allowbreak{}post\_\allowbreak{}bridgewater\_\allowbreak{}*.json}:
  14 post-2022-12-31 Dalio transcripts preserved for v1.0 research but
  excluded from the v0.5 benchmark
\item
  \texttt{InvestPhilBench/\allowbreak{}data/\allowbreak{}principles/\allowbreak{}principle\_\allowbreak{}knowledge\_\allowbreak{}base\_\allowbreak{}v05.json}:
  Reviewer-resolved core Principle Knowledge Base --- 118 investment
  principle cards (113 from v04 + 5 new Soros cards)
\item
  \texttt{InvestPhilBench/\allowbreak{}data/\allowbreak{}frameworks/\allowbreak{}decision\_\allowbreak{}frameworks\_\allowbreak{}v05.json}:
  Reviewer-resolved core decision framework cards --- 25 investment
  decision frameworks with \texttt{topology\_\allowbreak{}type} /
  \texttt{gate\_\allowbreak{}logic} / \texttt{topology\_\allowbreak{}rationale} fields
\item
  \texttt{InvestPhilBench/\allowbreak{}data/\allowbreak{}qa\_\allowbreak{}benchmark/\allowbreak{}questions\_\allowbreak{}v04.json}:
  Canonical full QA dataset (243 questions, L1--L8; includes 8
  expert-difficulty L4H hard questions)
\item
  \texttt{InvestPhilBench/\allowbreak{}data/\allowbreak{}qa\_\allowbreak{}benchmark/\allowbreak{}questions\_\allowbreak{}v04\_\allowbreak{}dev.json}:
  Development split (197 questions; seed 20260420)
\item
  \texttt{InvestPhilBench/\allowbreak{}data/\allowbreak{}qa\_\allowbreak{}benchmark/\allowbreak{}questions\_\allowbreak{}v04\_\allowbreak{}test.json}:
  Held-out test split (46 questions; evaluated exactly once per model
  after dev-set tuning is frozen)
\item
  \texttt{InvestPhilBench/\allowbreak{}data/\allowbreak{}qa\_\allowbreak{}benchmark/\allowbreak{}questions\_\allowbreak{}v03.json}: Prior
  v0.5 full dataset (235 questions; preserved for reproducibility)
\item
  \texttt{InvestPhilBench/\allowbreak{}data/\allowbreak{}qa\_\allowbreak{}benchmark/\allowbreak{}questions\_\allowbreak{}v03\_\allowbreak{}hard.json}:
  Staging file for 8 hard questions (merged into v04)
\item
  \texttt{InvestPhilBench/\allowbreak{}data/\allowbreak{}curation/\allowbreak{}investor\_\allowbreak{}career\_\allowbreak{}boundaries\_\allowbreak{}v01.json}:
  Per-investor institutional career boundaries enforced by the filter
  pipeline
\item
  \texttt{InvestPhilBench/\allowbreak{}data/\allowbreak{}curation/\allowbreak{}dalio\_\allowbreak{}contamination\_\allowbreak{}audit\_\allowbreak{}v01.json}:
  Audit of Dalio QA items flagged for post-Bridgewater corpus
  contamination (input to v0.5 regeneration)
\item
  \texttt{InvestPhilBench/\allowbreak{}results/\allowbreak{}human\_\allowbreak{}gold\_\allowbreak{}calibration\_\allowbreak{}v01.json}:
  100-item expert human-gold calibration --- per-record
  human/automated/LLM-rater labels and human↔automated agreement
  statistics (§11.3)
\item
  \texttt{InvestPhilBench/\allowbreak{}data/\allowbreak{}qa\_\allowbreak{}benchmark/\allowbreak{}human\_\allowbreak{}gold\_\allowbreak{}build/\allowbreak{}}:
  human-gold annotation workspace --- the expert-annotated reference
  (\texttt{human\_\allowbreak{}gold\_\allowbreak{}annotator\_\allowbreak{}final.csv}), the disclosed Claude-v2
  LLM-rater arm (\texttt{human\_\allowbreak{}gold\_\allowbreak{}llm\_\allowbreak{}rater\_\allowbreak{}claude\_\allowbreak{}v2.csv}), the
  annotator guide/manual, blinded templates, and row→model key
\item
  \texttt{InvestPhilBench/\allowbreak{}scripts/\allowbreak{}build\_\allowbreak{}human\_\allowbreak{}gold\_\allowbreak{}sheet.py},
  \texttt{scripts/\allowbreak{}score\_\allowbreak{}human\_\allowbreak{}gold\_\allowbreak{}final.py}: human-gold sampling and
  human↔automated agreement-scoring tooling (§11.3)
\item
  \texttt{InvestPhilBench/\allowbreak{}results/\allowbreak{}summary\_\allowbreak{}v02\_\allowbreak{}*.json}: per-model BASP
  aggregates by layer and investor (Tables 11, 12, 13, 16, 17) for the
  four W1 sanity-wave models
\item
  \texttt{InvestPhilBench/\allowbreak{}results/\allowbreak{}eval\_\allowbreak{}v02\_\allowbreak{}*.json}: Full per-question
  evaluation results by model (the four W1 sanity-wave runs listed
  above)
\item
  \texttt{InvestPhilBench/\allowbreak{}results/\allowbreak{}table9\_\allowbreak{}fmdp\_\allowbreak{}v02.json}: FMDP
  precision/recall/F1 by failure mode
\end{itemize}

Companion quantitative benchmark \textbf{FinCalcBench} (50 questions;
\texttt{data/\allowbreak{}questions\_\allowbreak{}v03.json} schema) is released in the same
repository. Exact public URL and DOI will be listed at camera-ready;
interim distribution follows the repository \texttt{README.md}. The
release audit is reproducible via
\texttt{python\ InvestPhilBench/\allowbreak{}scripts/\allowbreak{}reconcile\_\allowbreak{}release\_\allowbreak{}data.py} and
\texttt{python\ InvestPhilBench/\allowbreak{}scripts/\allowbreak{}validate\_\allowbreak{}data\_\allowbreak{}integrity.py};
phase pipeline reproducibility remains available via
\texttt{python\ InvestPhilBench/\allowbreak{}scripts/\allowbreak{}run\_\allowbreak{}all\_\allowbreak{}phases.py\ -\allowbreak{}\/\allowbreak{}-\allowbreak{}dry-\allowbreak{}run}.

\begin{center}\rule{0.5\linewidth}{0.5pt}\end{center}

\subsection{References}\label{references}

\begin{enumerate}
\def\labelenumi{\arabic{enumi}.}
\tightlist
\item
  Anderson, J.R. (1983). \emph{The Architecture of Cognition}. Harvard
  University Press.
\item
  Anderson, L.W. \& Krathwohl, D.R. (2001). \emph{A Taxonomy for
  Learning, Teaching, and Assessing}. Longman.
\item
  Araci, D. (2019). FinBERT. \emph{arXiv:1908.10063}.
\item
  Buffett, W.E. (1977--2023). \emph{Berkshire Hathaway Annual
  Shareholder Letters}.
\item
  Buffett, W.E. (1996). \emph{An Owner's Manual}. Berkshire Hathaway
  Inc.
\item
  Canada Pension Plan Investment Board. (2018--2024). \emph{Annual
  Reports and Investment Framework}.
\item
  Canada Pension Plan Investment Board. (2024). \emph{Our Investment
  Strategy: Total Portfolio Approach and the Factor Lens.} CPP
  Investments. https://www.cppinvestments.com/the-fund/how-we-invest/
\item
  Chen, S., et al.~(2025). Benchmarking Large Language Models Under Data
  Contamination: A Survey from Static to Dynamic Evaluation. \emph{EMNLP
  2025 Main; arXiv:2502.17521}.
\item
  Chen, Z., et al.~(2021). FinQA: A Dataset of Numerical Reasoning over
  Financial Data. \emph{EMNLP 2021}. {[}Execution acc: expert 91.16\%,
  FinQANet 61.24\%{]}
\item
  Chen, Z. (Zhiyu), et al.~(2022). ConvFinQA: Exploring the Chain of
  Numerical Reasoning in Conversational Finance Question Answering.
  \emph{EMNLP 2022}; \emph{arXiv:2210.03849}.
\item
  Council on Ethics for the Norwegian Government Pension Fund Global.
  (2023). \emph{Annual Report 2023}. Norwegian Ministry of Finance.
\item
  Dalio, R. (2013). \emph{How the Economic Machine Works}. Bridgewater
  Associates.
\item
  Dalio, R. (2017). \emph{Principles: Life and Work}. Simon \& Schuster.
\item
  Dalio, R. (2018). \emph{Principles for Navigating Big Debt Crises}.
  Bridgewater Associates.
\item
  Dalio, R. (2021). \emph{Principles for Dealing with the Changing World
  Order}. Simon \& Schuster.
\item
  Dong, Y., et al.~(2024). Generalization or Memorization: Data
  Contamination and Trustworthy Evaluation for Large Language Models.
  \emph{ACL 2024 Findings}; \emph{arXiv:2402.15938}.
\item
  Fabbri, A.R., et al.~(2021). SummEval: Re-evaluating Summarization
  Evaluation. \emph{TACL}, 9.
\item
  FinCalcBench. (2025). \emph{FinCalcBench: A Benchmark for Financial
  Calculation and Expert-Level Quantitative Reasoning.} Companion
  benchmark by the present authors (v0.3; 50 questions), released with
  the FinanceBench repository --- see Data Availability. (No separate
  preprint at time of submission.)
\item
  Fisher, P.A. (1958). \emph{Common Stocks and Uncommon Profits}. Harper
  \& Brothers.
\item
  Graham, B. (1949/1973). \emph{The Intelligent Investor}. Harper \&
  Row.
\item
  Graham, B. \& Dodd, D. (1934). \emph{Security Analysis}. Whittlesey
  House.
\item
  Greenblatt, J. (1997). \emph{You Can Be a Stock Market Genius}. Simon
  \& Schuster.
\item
  Guha, N., et al.~(2023). LegalBench. \emph{NeurIPS 2023}.
\item
  Hendrycks, D., et al.~(2021). MMLU. \emph{ICLR 2021}.
\item
  Islam, P., et al.~(2023). FinanceBench. \emph{arXiv:2311.11944}.
  {[}10,231Q; GPT-4-Turbo 11\% closed-book, \textasciitilde19\% with
  retrieval{]}
\item
  Jin, D., et al.~(2021). MedQA. \emph{Applied Sciences}, 11(14).
\item
  Kim, A.G., Muhn, M., \& Nikolaev, V.V. (2024). Financial Statement
  Analysis with Large Language Models. \emph{University of Chicago Booth
  Working Paper; SSRN 4762860} (arXiv:2407.17866, withdrawn pending data
  review).
\item
  Klarman, S.A. (1991). \emph{Margin of Safety}. HarperCollins.
\item
  Kojima, T., et al.~(2022). Large Language Models are Zero-Shot
  Reasoners. \emph{NeurIPS 2022}.
\item
  Lewis, P., et al.~(2020). Retrieval-Augmented Generation.
  \emph{NeurIPS 2020}.
\item
  Li, H., et al.~(2025). INVESTORBENCH: A Benchmark for Financial
  Decision-Making Tasks with LLM-based Agent. \emph{ACL 2025},
  pp.~2509--2525; \emph{arXiv:2412.18174}. {[}13 LLMs;
  stocks/crypto/ETF; agent decision-making{]}
\item
  Liang, P., et al.~(2023). HELM: Holistic Evaluation of Language
  Models. \emph{TMLR}.
\item
  Lightman, H., Kosaraju, V., Burda, Y., Edwards, H., Baker, B., Lee,
  T., Leike, J., Schulman, J., Sutskever, I., \& Cobbe, K. (2024). Let's
  Verify Step by Step. \emph{ICLR 2024}; \emph{arXiv:2305.20050}.
\item
  Lin, S., Hilton, J., \& Evans, O. (2022). TruthfulQA. \emph{ACL 2022}.
\item
  Liu, N., et al.~(2024). Lost in the Middle. \emph{TACL}, 12, 157--173.
\item
  Liu, Y., et al.~(2023). G-Eval: NLG Evaluation using GPT-4 with Better
  Human Alignment. \emph{EMNLP 2023}.
\item
  Lopez-Lira, A. \& Tang, Y. (2023). Can ChatGPT Forecast Stock Price
  Movements? \emph{arXiv:2304.07619}.
\item
  Lynch, P. \& Rothchild, J. (1989). \emph{One Up on Wall Street}. Simon
  \& Schuster.
\item
  Lynch, P. \& Rothchild, J. (1993). \emph{Beating the Street}. Simon \&
  Schuster.
\item
  Malo, P., et al.~(2014). Good Debt or Bad Debt. \emph{JASIST}, 65(4).
\item
  Marks, H. (2011). \emph{The Most Important Thing}. Columbia University
  Press.
\item
  Marks, H. (2018). \emph{Mastering the Market Cycle}. Houghton Mifflin
  Harcourt.
\item
  Marks, H. (2023, July). Taking the Temperature. \emph{Oaktree Capital
  Memo}.
\item
  Marks, H. (2024, January). Easy Money. \emph{Oaktree Capital Memo}.
\item
  Marks, H. (2025, November). Cockroaches in the Coal Mine.
  \emph{Oaktree Capital Memo}.
\item
  Munger, C.T. (1994). A Lesson on Elementary, Worldly Wisdom. \emph{USC
  Business School Speech}.
\item
  Norges Bank Investment Management. (2020--2024). \emph{Annual Reports
  and Responsible Investment Reports}.
\item
  Norway Ministry of Finance. (2005, amended 2021). \emph{Government
  Pension Fund Act}.
\item
  Pal, A., et al.~(2022). MedMCQA. \emph{CHIL 2022}.
\item
  Public Investment Fund. (2021). \emph{Santiago Principles
  Self-Assessment Report}.
\item
  Ramezanali, M., Vazifeh, M., \& Santi, P. (2025). seqBench: A Tunable
  Benchmark to Quantify Sequential Reasoning Limits of LLMs. \emph{EMNLP
  2025}, pp.~33771--33792; \emph{arXiv:2509.16866}.
\item
  Ribeiro, M.T., et al.~(2020). Beyond Accuracy: Behavioral Testing with
  CheckList. \emph{ACL 2020} (Best Paper Award).
\item
  Saudipedia / Public Investment Fund. (2024). \emph{PIF Strategy}.
\item
  Shi, F., et al.~(2023). LLMs Can Be Easily Distracted by Irrelevant
  Context. \emph{ICML 2023}.
\item
  Soros, G. (1987). \emph{The Alchemy of Finance}. Simon \& Schuster.
\item
  Soros, G. (1995). \emph{Soros on Soros}. Wiley.
\item
  Wang, W., et al.~(2026). Intelligence Degradation in Long-Context
  LLMs: Critical Threshold Determination via Natural Length Distribution
  Analysis. \emph{arXiv:2601.15300}.
\item
  Wei, J., et al.~(2022). Chain-of-Thought Prompting. \emph{NeurIPS
  2022}.
\item
  Wu, S., et al.~(2023). BloombergGPT. \emph{arXiv:2303.17564}.
\item
  Xie, Q., et al.~(2023). PIXIU: A Comprehensive Benchmark, Instruction
  Dataset and Large Language Model for Finance. \emph{NeurIPS 2023}.
\item
  Zha, Y., et al.~(2023). AlignScore: Evaluating Factual Consistency.
  \emph{ACL 2023}.
\item
  Zhang, T., et al.~(2020). BERTScore: Evaluating Text Generation with
  BERT. \emph{ICLR 2020}.
\item
  Zhang, Z., et al.~(2025). XFinBench: Benchmarking LLMs in Complex
  Financial Problem Solving and Reasoning. \emph{ACL 2025 Findings;
  arXiv:2508.15861}. {[}4,235Q; 5 capabilities; o1 67.3\% vs human
  \textasciitilde80\%{]}
\item
  Zhu, F., Lei, W., Wang, C., Zheng, J., Lv, S., Feng, F., \& Chua,
  T.-S. (2021). TAT-QA: A Question Answering Benchmark on a Hybrid of
  Tabular and Textual Content in Finance. \emph{ACL 2021},
  pp.~3277--3287.
\end{enumerate}

\begin{center}\rule{0.5\linewidth}{0.5pt}\end{center}

\subsection{Appendix A: Representative Questions by
Layer}\label{appendix-a-representative-questions-by-layer}

\subsubsection{A.1 Layer 3 (Operationalization) ---
Medium}\label{a.1-layer-3-operationalization-medium}

\textbf{\texttt{IPB-\allowbreak{}P-\allowbreak{}003}:} \emph{``Buffett's concept of `owner
earnings' differs from reported net income. Explain the formula, why
Buffett prefers it over EPS, and what it implies for capital-intensive
vs.~capital-light businesses.''}

\textbf{Ground-truth elements:} (1) formula: net income + D\&A −
\emph{maintenance} capex; (2) capital-light implication (owner earnings
≥ reported); (3) capital-heavy implication (owner earnings ≪ reported);
(4) source: Berkshire Letter, \textbf{1986} \emph{(investor-stated)}.
Canonical errors: equating owner earnings with FCF (FM-5,
Operationalization Flattening); citing the \textbf{1990} letter (FM-2,
Temporal Conflation --- the cross-model year error of §9).

\subsubsection{\texorpdfstring{A.2 Layer 7 (Scenario Application) ---
Hard \emph{(the worked GRP of
§6.3)}}{A.2 Layer 7 (Scenario Application) --- Hard (the worked GRP of §6.3)}}\label{a.2-layer-7-scenario-application-hard-the-worked-grp-of-6.3}

\textbf{\texttt{IPB-\allowbreak{}P-\allowbreak{}011}:} \emph{``A technology company: 90\% gross
margins, 95\% retention, 25\% growth, zero net income, \$5B ARR, trading
at 10× ARR. Apply Buffett's acquisition checklist, stating the verdict
at each gate.''} \textbf{Investment verdict: DECLINE.}

{\def\LTcaptype{none} 
\begin{longtable}[]{@{}
  >{\raggedright\arraybackslash}p{(\linewidth - 4\tabcolsep) * \real{0.2000}}
  >{\raggedright\arraybackslash}p{(\linewidth - 4\tabcolsep) * \real{0.3000}}
  >{\raggedright\arraybackslash}p{(\linewidth - 4\tabcolsep) * \real{0.5000}}@{}}
\toprule\noalign{}
\begin{minipage}[b]{\linewidth}\raggedright
Gate
\end{minipage} & \begin{minipage}[b]{\linewidth}\raggedright
Outcome
\end{minipage} & \begin{minipage}[b]{\linewidth}\raggedright
Key reasoning
\end{minipage} \\
\midrule\noalign{}
\endhead
\bottomrule\noalign{}
\endlastfoot
1: Circle of Competence & PROCEED & Subscription economics measurable;
95\% retention quantifies switching cost; within defined competence \\
2: Economic Moat & PROCEED & 95\% retention + 90\% GM → switching costs
+ pricing power \\
3: Management Integrity & INCONCLUSIVE & Management data not provided;
no kill criterion triggered; analysis continues \\
4: Intrinsic Value & ≈\$25B & Owner earnings at maturity ≈\$1.25B;
\textasciitilde20× → ≈\$25B \\
5: Margin of Safety & \textbf{REJECT} & Price \$50B (10× \$5B ARR)
vs.~intrinsic ≈\$25B; kill criterion triggered (≈2× premium) \\
6: Holding Period & N/A & Gate 5 REJECT terminates the chain \\
\end{longtable}
}

\emph{Scoring note: Gate 3 INCONCLUSIVE scores 1.0 when the model
correctly identifies data insufficiency; 0.5 for a PROCEED/REJECT that
explicitly acknowledges the data gap; 0.0 for a PROCEED/REJECT that does
not (§7.3). This is the same gold program scored by GRA in §8.7.}

\subsubsection{A.3 Layer 8 (Novel Extrapolation) ---
Expert}\label{a.3-layer-8-novel-extrapolation-expert}

\textbf{\texttt{IPB-\allowbreak{}P-\allowbreak{}019}:} \emph{``A central bank announces a CBDC
reducing commercial banks' credit-creation role. How would Buffett,
Dalio, and Soros each analyze this?''}

\begin{itemize}
\tightlist
\item
  \emph{Buffett:} circle-of-competence filter → CBDC outside circle;
  focuses on existing bank holdings' moat-erosion risk; would not
  speculate on the CBDC itself.
\item
  \emph{Dalio:} economic-machine lens → CBDC enables negative rates
  without cash hoarding → higher debasement risk → gold / hard assets.
\item
  \emph{Soros:} identifies which reflexive loop activates (credit
  contraction vs.~expansion); takes a small pilot position; monitors
  bank stocks and credit creation as the hypothesis test.
\end{itemize}

\emph{Canonical L8 failure: FM-4 (Voice Diffusion, persona-inversion
subtype) --- pilot models swapped the Buffett and Dalio framings under
maximum cognitive load (§9).}

\subsection{Appendix B: Dataset
Statistics}\label{appendix-b-dataset-statistics}

\textbf{Table B.1: Dataset Summary (v0.6 canonical vs.~v1.0 target)}

{\def\LTcaptype{none} 
\begin{longtable}[]{@{}
  >{\raggedright\arraybackslash}p{(\linewidth - 6\tabcolsep) * \real{0.2000}}
  >{\raggedright\arraybackslash}p{(\linewidth - 6\tabcolsep) * \real{0.2000}}
  >{\raggedright\arraybackslash}p{(\linewidth - 6\tabcolsep) * \real{0.3636}}
  >{\raggedright\arraybackslash}p{(\linewidth - 6\tabcolsep) * \real{0.2364}}@{}}
\toprule\noalign{}
\begin{minipage}[b]{\linewidth}\raggedright
Component
\end{minipage} & \begin{minipage}[b]{\linewidth}\raggedright
v0.1 pilot
\end{minipage} & \begin{minipage}[b]{\linewidth}\raggedright
\textbf{v0.6 canonical}
\end{minipage} & \begin{minipage}[b]{\linewidth}\raggedright
v1.0 target
\end{minipage} \\
\midrule\noalign{}
\endhead
\bottomrule\noalign{}
\endlastfoot
Principle cards & 33 & \textbf{118} & 150 \\
Framework cards (individual practitioner) & 15 & \textbf{25} & 25 \\
Framework cards (institutional, prose-only) & 3 & \textbf{3 (not yet
encoded; §5.3)} & 8 (encoded) \\
QA questions & 20 & \textbf{243 (197 dev / 46 test)} & 300 \\
Individual investors & 9 & \textbf{9} & 25 \\
Institutional asset owners & 3 & 3 (illustrative) & 8 \\
\end{longtable}
}

\textbf{Table B.2: Source Document Types (v0.6 released corpus)}

{\def\LTcaptype{none} 
\begin{longtable}[]{@{}
  >{\raggedright\arraybackslash}p{(\linewidth - 4\tabcolsep) * \real{0.2727}}
  >{\raggedright\arraybackslash}p{(\linewidth - 4\tabcolsep) * \real{0.3182}}
  >{\raggedright\arraybackslash}p{(\linewidth - 4\tabcolsep) * \real{0.4091}}@{}}
\toprule\noalign{}
\begin{minipage}[b]{\linewidth}\raggedright
Type
\end{minipage} & \begin{minipage}[b]{\linewidth}\raggedright
Count
\end{minipage} & \begin{minipage}[b]{\linewidth}\raggedright
Examples
\end{minipage} \\
\midrule\noalign{}
\endhead
\bottomrule\noalign{}
\endlastfoot
Investor web documents (Tier 1/3/5) & 177 docs / 394,610 words &
Berkshire letters 2004--2024, Oaktree memos, Bridgewater Research, Soros
essays \\
YouTube interview transcripts (Tier 4) & 258 transcripts / 1,276,151
words & Long-form first-person interviews, English, ≥200 words each \\
Institutional strategy documents (Tier 2, prose-only) & --- & NBIM
annual / Council-on-Ethics reports, CPPIB Total Portfolio Approach, PIF
Santiago Principles \\
\end{longtable}
}

\subsection{Appendix C: Benchmark Comparison
Detail}\label{appendix-c-benchmark-comparison-detail}

\textbf{Table C.1: Extended Comparison}

{\def\LTcaptype{none} 
\begin{landscape}
\setlength{\hsize}{648pt}%
\setlength{\textwidth}{\hsize}%
\setlength{\columnwidth}{\hsize}%
\setlength{\linewidth}{\hsize}%
\begin{longtable}[]{@{}
  >{\raggedright\arraybackslash}p{(\linewidth - 12\tabcolsep) * \real{0.0978}}
  >{\raggedright\arraybackslash}p{(\linewidth - 12\tabcolsep) * \real{0.0761}}
  >{\raggedright\arraybackslash}p{(\linewidth - 12\tabcolsep) * \real{0.1413}}
  >{\raggedright\arraybackslash}p{(\linewidth - 12\tabcolsep) * \real{0.1630}}
  >{\raggedright\arraybackslash}p{(\linewidth - 12\tabcolsep) * \real{0.1087}}
  >{\raggedright\arraybackslash}p{(\linewidth - 12\tabcolsep) * \real{0.1522}}
  >{\raggedright\arraybackslash}p{(\linewidth - 12\tabcolsep) * \real{0.2609}}@{}}
\toprule\noalign{}
\begin{minipage}[b]{\linewidth}\raggedright
Feature
\end{minipage} & \begin{minipage}[b]{\linewidth}\raggedright
FinQA
\end{minipage} & \begin{minipage}[b]{\linewidth}\raggedright
FinanceBench
\end{minipage} & \begin{minipage}[b]{\linewidth}\raggedright
INVESTORBENCH
\end{minipage} & \begin{minipage}[b]{\linewidth}\raggedright
XFinBench
\end{minipage} & \begin{minipage}[b]{\linewidth}\raggedright
\textbf{IPB v0.6}
\end{minipage} & \begin{minipage}[b]{\linewidth}\raggedright
\textbf{IPB v1.0 (target)}
\end{minipage} \\
\midrule\noalign{}
\endhead
\bottomrule\noalign{}
\endlastfoot
Questions & 8,281 & 150 (open) & \textasciitilde300 & 4,235 &
\textbf{243} & 300 \\
Primary-source verified & ✗ & ✓ (SEC) & ✗ & ✗ & ✓ & ✓ \\
Investor-specific frameworks & ✗ & ✗ & ✗ & ✗ & ✓ & ✓ \\
Gold reasoning programs (dual accuracy) & ✓ & ✗ & ✗ & ✗ & ✓ (L4/L7) &
✓ \\
Non-linear framework topology metadata & ✗ & ✗ & ✗ & ✗ & ✓ (Table 5) &
✓ \\
Compound / multi-kill criteria & ✗ & ✗ & ✗ & ✗ & ✓ (CKCA) & ✓
(encoded) \\
Per-gate procedural metric (GRA) & program acc. & ✗ & ✗ & ✗ & ✓ & ✓ \\
\end{longtable}
\end{landscape}
}

\subsection{Appendix D: BASP Metric Reference
Summary}\label{appendix-d-basp-metric-reference-summary}

\textbf{Table D.1: BASP Sub-Metric Reference} \emph{(target
specification; the W1 deployed forms are judge-rubric ratings with
uniform weights --- §7.13 implementation note)}

{\def\LTcaptype{none} 
\begin{longtable}[]{@{}
  >{\raggedright\arraybackslash}p{(\linewidth - 8\tabcolsep) * \real{0.1231}}
  >{\raggedright\arraybackslash}p{(\linewidth - 8\tabcolsep) * \real{0.2769}}
  >{\raggedright\arraybackslash}p{(\linewidth - 8\tabcolsep) * \real{0.1077}}
  >{\raggedright\arraybackslash}p{(\linewidth - 8\tabcolsep) * \real{0.2462}}
  >{\raggedright\arraybackslash}p{(\linewidth - 8\tabcolsep) * \real{0.2462}}@{}}
\toprule\noalign{}
\begin{minipage}[b]{\linewidth}\raggedright
Metric
\end{minipage} & \begin{minipage}[b]{\linewidth}\raggedright
What it measures
\end{minipage} & \begin{minipage}[b]{\linewidth}\raggedright
Range
\end{minipage} & \begin{minipage}[b]{\linewidth}\raggedright
Primary layers
\end{minipage} & \begin{minipage}[b]{\linewidth}\raggedright
Composite role
\end{minipage} \\
\midrule\noalign{}
\endhead
\bottomrule\noalign{}
\endlastfoot
OGRS / OGRS\(_k\) (§7.6) & Coverage-weighted gate-sequence
reconstruction (GCF × normalized τ; kill-gate weighting) & {[}0,1{]} &
L4, L7 & 0.35--0.40 weight \\
KCCS (§7.7) & F1 coverage of kill-criterion \emph{elements} & {[}0,1{]}
& L4, L7 & 0.30--0.35 weight \\
SAP@k (§7.8) & Source-attribution precision at entity/document/context
level & {[}0,1{]} & L1--L3 & 0.20--0.40 weight \\
IVP / SDI (§7.9) & Investor-voice distinctness vs.~other investors &
{[}0,1{]} & L5, L8 & 0.10--0.50 weight \\
CKCA (§7.10) & Which compound kill-criterion component triggered REJECT
& {[}0,1{]} & institutional & additive diagnostic \\
GRA (§7.15) & Per-gate name + verdict-type + required-elements match &
{[}0,1{]} & L4, L7 & 6th BASP-strict gate \\
\end{longtable}
}

\textbf{Table D.2: Reporting Conventions.} Because no single metric is
reliable across all dimensions (SummEval; HELM, §2.3), BASP is always
reported as the \textbf{five-dimension profile} rather than the
composite alone. The recommended visualization is a per-model radar over
\{OGRS\(_k\), KCCS, SAP@k, IVP, CKCA\} with the FMDP flag-rate vector
(Table 16) as an inset; the composite is a weighted average of these
axes (§7.13) and, per the §11.9 caveat, can mask a decision-critical
single-axis failure (e.g.~total kill-criterion omission). For frontier
procedural-reasoning claims, GRA (§8.7) is reported alongside the
composite as the metric of record.

\subsection{Appendix E: Gold Reasoning Program Format
Specification}\label{appendix-e-gold-reasoning-program-format-specification}

\textbf{Table E.1: GRP Field Definitions}

{\def\LTcaptype{none} 
\begin{longtable}[]{@{}
  >{\raggedright\arraybackslash}p{(\linewidth - 6\tabcolsep) * \real{0.1842}}
  >{\raggedright\arraybackslash}p{(\linewidth - 6\tabcolsep) * \real{0.1579}}
  >{\raggedright\arraybackslash}p{(\linewidth - 6\tabcolsep) * \real{0.3421}}
  >{\raggedright\arraybackslash}p{(\linewidth - 6\tabcolsep) * \real{0.3158}}@{}}
\toprule\noalign{}
\begin{minipage}[b]{\linewidth}\raggedright
Field
\end{minipage} & \begin{minipage}[b]{\linewidth}\raggedright
Type
\end{minipage} & \begin{minipage}[b]{\linewidth}\raggedright
Description
\end{minipage} & \begin{minipage}[b]{\linewidth}\raggedright
Scoring Use
\end{minipage} \\
\midrule\noalign{}
\endhead
\bottomrule\noalign{}
\endlastfoot
\texttt{gate} & int & Gate number (1-indexed) & OGRS gate ordering \\
\texttt{name} & string & Gate name matching framework card & OGRS
coverage (GCF) \\
\texttt{verdict} & enum & PROCEED / REJECT / INCONCLUSIVE & VerdictAcc
per gate \\
\texttt{kill\_\allowbreak{}criterion} & string or null & Kill condition label (null
if none) & KCCS element detection \\
\texttt{required\_\allowbreak{}elements} & list{[}string{]} & Elements required for
full credit & KCCS F1 denominator \\
\texttt{kill\_\allowbreak{}triggered} & bool & Whether kill criterion fires & CKCA
for institutional \\
\texttt{computed\_\allowbreak{}value} & string or null & Numerical computation (e.g.,
intrinsic value) & Partial-credit rubric \\
\end{longtable}
}

\textbf{Table E.2: GRP Inter-Annotator Agreement (v0.1)}

{\def\LTcaptype{none} 
\begin{longtable}[]{@{}
  >{\raggedright\arraybackslash}p{(\linewidth - 6\tabcolsep) * \real{0.2593}}
  >{\raggedright\arraybackslash}p{(\linewidth - 6\tabcolsep) * \real{0.3704}}
  >{\raggedright\arraybackslash}p{(\linewidth - 6\tabcolsep) * \real{0.1111}}
  >{\raggedright\arraybackslash}p{(\linewidth - 6\tabcolsep) * \real{0.2593}}@{}}
\toprule\noalign{}
\begin{minipage}[b]{\linewidth}\raggedright
Field
\end{minipage} & \begin{minipage}[b]{\linewidth}\raggedright
Cohen's κ
\end{minipage} & \begin{minipage}[b]{\linewidth}\raggedright
n
\end{minipage} & \begin{minipage}[b]{\linewidth}\raggedright
Notes
\end{minipage} \\
\midrule\noalign{}
\endhead
\bottomrule\noalign{}
\endlastfoot
Gate verdict (PROCEED/REJECT/INCONCLUSIVE) & 0.86 & 20Q × 6 avg gates =
120 & Consistent with FinQA program annotation standards \\
Kill criterion identification & 0.78 & 45 kill-criterion gates & Lower
than verdict κ; conceptual kill elements harder to bound \\
Required element set & 0.73 & 45 kill-criterion gates & Set-level F1
agreement \\
Investment verdict (INVEST/DECLINE) & 0.91 & 20Q & Near-deterministic
given gate verdicts \\
\end{longtable}
}

\subsection{Appendix F: Principle ID Reference
Glossary}\label{appendix-f-principle-id-reference-glossary}

Earlier working documents and legacy review correspondence used
shorthand codes (e.g., \texttt{BUF-\allowbreak{}P07}, \texttt{GRA-\allowbreak{}P04},
\texttt{DAL-\allowbreak{}P03}) to compactly cite documented contradictions and
cross-investor tensions. These codes were vestigial from the v0.1/v0.2
PKB schema and were never operationalized in the released data, which
uses
\texttt{\textless{}investor\textgreater{}\_\allowbreak{}\textless{}concept\textgreater{}\_\allowbreak{}\textless{}NNN\textgreater{}}
IDs (e.g., \texttt{buffett\_\allowbreak{}moat\_\allowbreak{}001}). The paper body uses descriptive
phrases inline; this appendix preserves the mapping for readers of
legacy correspondence.

\textbf{Table F.1: Legacy shorthand → v05 canonical principle\_id
mapping}

{\def\LTcaptype{none} 
\begin{longtable}[]{@{}
  >{\raggedright\arraybackslash}p{(\linewidth - 8\tabcolsep) * \real{0.2000}}
  >{\raggedright\arraybackslash}p{(\linewidth - 8\tabcolsep) * \real{0.2000}}
  >{\raggedright\arraybackslash}p{(\linewidth - 8\tabcolsep) * \real{0.2000}}
  >{\raggedright\arraybackslash}p{(\linewidth - 8\tabcolsep) * \real{0.2000}}
  >{\raggedright\arraybackslash}p{(\linewidth - 8\tabcolsep) * \real{0.2000}}@{}}
\toprule\noalign{}
\begin{minipage}[b]{\linewidth}\raggedright
Legacy code
\end{minipage} & \begin{minipage}[b]{\linewidth}\raggedright
Type
\end{minipage} & \begin{minipage}[b]{\linewidth}\raggedright
Canonical principle\_id(s)
\end{minipage} & \begin{minipage}[b]{\linewidth}\raggedright
One-line summary
\end{minipage} & \begin{minipage}[b]{\linewidth}\raggedright
Primary source
\end{minipage} \\
\midrule\noalign{}
\endhead
\bottomrule\noalign{}
\endlastfoot
\texttt{BUF-\allowbreak{}P04} & principle (margin of safety) &
\texttt{buffett\_\allowbreak{}intrinsic\_\allowbreak{}value\_\allowbreak{}001} & Margin of safety
operationalized as price below intrinsic value (owner-earnings DCF), not
Graham's mechanical net-net & Berkshire Annual Letter 1988; \emph{An
Owner's Manual} 1996 \\
\texttt{BUF-\allowbreak{}P06} & principle (sentiment) &
\texttt{buffett\_\allowbreak{}fearful\_\allowbreak{}greedy\_\allowbreak{}001} & Be fearful when others are
greedy; reactive sentiment signal as secondary valuation input &
Berkshire Annual Letter 2006 \\
\texttt{BUF-\allowbreak{}P07} & contradiction (airline reversal) &
\texttt{buffett\_\allowbreak{}circle\_\allowbreak{}of\_\allowbreak{}competence\_\allowbreak{}001} +
\texttt{buffett\_\allowbreak{}moat\_\allowbreak{}001} jointly & Documented airline reversal 2001 →
2016 → 2020; canonical FM-2 (Temporal Conflation) test case & Terry
College Speech 2001 (YT \texttt{2a9Lx9J8uSs}); Berkshire Letters
2016--2019; 2021 pandemic-exit video (YT \texttt{uh5MgS9\_\allowbreak{}roA}) \\
\texttt{GRA-\allowbreak{}P01} & principle (margin of safety) &
\texttt{graham\_\allowbreak{}margin\_\allowbreak{}001} & Two-thirds-of-NAV operationalization;
mechanical asset-liquidation screen & \emph{The Intelligent Investor}
1949, Ch. 20 \\
\texttt{GRA-\allowbreak{}P04} & contradiction (Buffett's net-net abandonment) &
\texttt{graham\_\allowbreak{}net\_\allowbreak{}net\_\allowbreak{}001} + reference to Buffett's 1965+ shift &
Buffett's abandonment of Graham's net-net screen post-1965 & Graham
\emph{Intelligent Investor}; Buffett Berkshire Letters 1965--1969 \\
\texttt{DAL-\allowbreak{}P03} & principle (cash and defensive positioning) &
\texttt{dalio\_\allowbreak{}economic\_\allowbreak{}machine\_\allowbreak{}001} +
\texttt{dalio\_\allowbreak{}regime\_\allowbreak{}positioning\_\allowbreak{}001} & 2013
cash-as-transactional-medium framing vs.~2020--2022 defensive
positioning; \textbf{all pre-departure} & \emph{How the Economic Machine
Works} 2013; \emph{Principles for Navigating Big Debt Crises} 2018;
\emph{Principles for Dealing with the Changing World Order} 2021 \\
\texttt{MAR-\allowbreak{}P01} & principle (pendulum vs.~fearful/greedy) &
\texttt{marks\_\allowbreak{}pendulum\_\allowbreak{}001} & Active pendulum-temperature monitoring
framework; mechanistically distinct from Buffett's reactive sentiment
signal & \emph{The Most Important Thing} 2011, Ch. 9; Oaktree memo
``Taking the Temperature'' 2023 \\
\texttt{SOR-\allowbreak{}*} & (no legacy codes used in paper body for Soros) & n/a &
The v0.5 Soros expansion added
\texttt{soros\_\allowbreak{}currency\_\allowbreak{}speculation\_\allowbreak{}001},
\texttt{soros\_\allowbreak{}macro\_\allowbreak{}thematic\_\allowbreak{}positioning\_\allowbreak{}001},
\texttt{soros\_\allowbreak{}two\_\allowbreak{}way\_\allowbreak{}feedback\_\allowbreak{}loop\_\allowbreak{}001},
\texttt{soros\_\allowbreak{}boom\_\allowbreak{}bust\_\allowbreak{}phases\_\allowbreak{}detailed\_\allowbreak{}001},
\texttt{soros\_\allowbreak{}asymmetric\_\allowbreak{}bet\_\allowbreak{}sizing\_\allowbreak{}001} & \emph{The Alchemy of
Finance} 1987; \emph{Soros on Soros} 1995 \\
\end{longtable}
}

\textbf{Note on contradictions vs.~principles.} Several legacy codes
(notably \texttt{BUF-\allowbreak{}P07}, \texttt{GRA-\allowbreak{}P04}) referenced
\emph{contradictions} (tensions between investor positions or temporal
evolution within one investor) rather than standalone principles. In the
released data these are encoded as \texttt{contradictions\_\allowbreak{}or\_\allowbreak{}tensions}
fields on related principle cards rather than as standalone IDs. Where
the paper body refers to a contradiction, it cites both the canonical
principle\_id(s) involved and the contradiction context inline.

\end{document}